\documentclass[graybox,envcountsect,envcountsame]{svmult}

\usepackage{type1cm}        
\usepackage{makeidx}         
\usepackage{graphicx}        
\usepackage{multicol}        
\usepackage[bottom]{footmisc}

\usepackage{newtxtext}       %
\usepackage{newtxmath}       

\usepackage{amsmath,amsfonts,amstext}
\usepackage{epigraph}
\usepackage{mathtools}
\usepackage{epstopdf}
\usepackage{caption}
\usepackage{subcaption}
\usepackage{float}
\usepackage{parskip}
\usepackage{verbatim}
\usepackage{enumitem}
\usepackage{hyperref}
\usepackage{dsfont}
\usepackage{shuffle}
\usepackage{graphics}
\usepackage{graphicx}
\usepackage{dcolumn}
\usepackage{bm}
\usepackage{epsfig}
\usepackage{multirow}
\usepackage{dcolumn}
\usepackage{graphicx}%
\usepackage[mathscr]{euscript}
\usepackage{notoccite}
\usepackage{xhfill}
\usepackage{makecell}
\usepackage{pgfplots}
\usepackage{adjustbox}
\usepackage{tikz}
\usetikzlibrary{arrows}
\usetikzlibrary{shapes}
\usetikzlibrary{decorations}
\usetikzlibrary{math}
\usetikzlibrary{calc}
\usetikzlibrary{shapes,
                chains,
                scopes}

\DeclareFontFamily{U}{shuffle}{}
\DeclareFontShape{U}{shuffle}{m}{n}{ <-8>shuffle7 <8->shuffle10}{}

\definecolor{niceblue}{rgb}{0.1,0.2,0.9}
\definecolor{lightblue}{rgb}{0.45,0.5,1}
\definecolor{purple}{rgb}{0.65,0.1,0.7}
\definecolor{gray}{rgb}{0.7,0.7,0.7}
\definecolor{darkgreen}{rgb}{0,0.5,0}
\definecolor{darkred}{rgb}{0.7,0.1,0.1}

\def\nb{\textcolor{niceblue}}

\def\rd{\textcolor{darkred}}

\newcommand{\R}{\mathbb R}
\newcommand{\PP}{\mathbb{P}}
\newcommand{\QQ}{\mathbb{Q}}
\newcommand{\mcX}{\mathcal{X}}
\newcommand{\mcY}{\mathcal{Y}}
\newcommand{\mcF}{\mathcal{F}}
\newcommand{\Shuffles}{\textnormal{Shuffles}}
\newcommand{\floor}[1]{\lfloor #1 \rfloor}
\newcommand{\LLL}{\mathbf L}

\newcommand{\e}{\mathbf{e}}

\allowdisplaybreaks

\makeindex             

\pgfplotsset{compat=1.18}

\begin{document}

\title*{A Primer on the Signature Method in Machine Learning}
\author{Ilya Chevyrev and Andrey Kormilitzin}
\institute{Ilya Chevyrev \at
Institut für Mathematik, TU Berlin,  Str. des 17. Juni 136, 10623 Berlin, Germany and\\
School of Mathematics, The University of Edinburgh, James Clerk Maxwell Building, Peter Guthrie Tait Rd, Edinburgh EH9 3FD, UK.
\email{ichevyrev@gmail.com}
\and Andrey Kormilitzin \at Mathematical Institute, University of Oxford, Woodstock Road, Oxford OX2 6GG, UK and\\
Department of Psychiatry, University of Oxford, Warneford Ln, Oxford OX3 7JX, UK.
\email{andrey.kormilitzin@maths.ox.ac.uk}}
%
%
\maketitle

\abstract{We provide an introduction to the signature method, focusing on its theoretical properties and machine learning applications. Our presentation is divided into two parts. In the first part, we present the definition and fundamental properties of the signature of a path. The signature is a sequence of numbers associated with a path that captures many of its important analytic and geometric properties. As a sequence of numbers, the signature serves as a compact description (dimension reduction) of a path. In presenting its theoretical properties, we assume only familiarity with classical real analysis and integration, and supplement theory with straightforward examples. We also mention several advanced topics, including the role of the signature in rough path theory. In the second part, we present practical applications of the signature to the area of machine learning. The signature method is a non-parametric way of transforming data into a set of features that can be used in machine learning tasks. In this method, data are converted into multi-dimensional paths, by means of embedding algorithms, of which the signature is then computed. We describe this pipeline in detail, making a link with the properties of the signature presented in the first part. We furthermore review some of the developments of the signature method in machine learning and, as an illustrative example, present a detailed application of the method to handwritten digit classification.}

\section*{Introduction}

Paths are used ubiquitously to describe \textit{time-ordered data}, which appear in a variety of contexts.
There are a number of important examples of time-ordered data, some of which we do not often think of as such:
\begin{itemize}
    \item Financial time series -- e.g. price of a stock or index,
    \item Text -- e.g. ``\textit{Lorem ipsum dolor sit amet, consectetur adipiscing elit,}''
    \item Evolving networks -- e.g. time evolution of contacts on social media.
\end{itemize}
Each of these examples can be mathematically described as a function $X$ from an index set $I\subset \R$ of real numbers to a state space $\mathcal{X}$, that is, a \textit{path} $X: I\to \mathcal{X}$. For the three examples above, the corresponding index set and state space are
\begin{itemize}
    \item Financial time series -- finite interval $I=[a,b]$ and Euclidean space $\mathcal{X}=\R^d$,
    \item Text -- positive integers $I=\{0,1,\ldots, N\}$ and alphabet $\mathcal{X}=\{a,A,b,B,\ldots\}$,
    \item Evolving networks -- positive integers $I=\{0,1,\ldots, N\}$ and space $\mcX$ of graphs or adjacency matrices.
\end{itemize}
In the sequel, we restrict ourselves to time-ordered data of the form $X: [a,b]\to \R^d$.
One may frequently arrive at this case by composing with a function $\varphi : \mcX\to \R^d$ and by suitably interpolating discrete points if starting from discrete data 
(we give further details in Section~\ref{sec:prac_app}).

Since time-ordered data may be complicated, sometimes requiring a large amount of memory just to store it,
it becomes important in the analysis of this data to find features that are both descriptive and compact.

Over the past decade, the \textit{signature} of a path has become a notable feature that can satisfy both of these properties.
Roughly speaking, the signature is a sequence of numbers associated with a path, given by iterated integrals.
The signature turns out to be highly descriptive, making it a powerful theoretical and practical tool to study paths.
In a way that we make precise below, the signature is a natural generalization of polynomials to (parametrized or unparametrized) paths.

This chapter provides an introduction to the theoretical and practical aspects of the signature.
The signature has gained attention in the mathematical community over the past 20 years in large part due to its connection with Lyons' theory of rough paths.
However, one of the main points we wish to emphasize is that no knowledge beyond classical integration theory is required to define and study the basic properties of the signature. Indeed, K. T. Chen studied the signature of a path in the 50's, being one of the first authors to do so, and his primary results can be stated completely in terms of piecewise smooth paths, which already provide an elegant and deep mathematical theory.
As such, we aim to make the chapter as self-contained as possible while remaining at an elementary level.

Since the first version of these lecture notes appeared in 2016, the applications of signatures have expanded dramatically; we do not review many of these interesting and important developments.
We have instead chosen several conceptual applications which we hope, together with the theoretical foundations, can serve as an introduction to anyone wishing to learn more advanced topics or to apply the signature in practice.

\begin{acknowledgement}
We are thankful to Horatio Boedihardjo, Simone Dari, Terry Lyons, Hao Ni, Harald Oberhauser, and Danyu Yang for fruitful discussions and improvements of the manuscript.
IC acknowledges support from the EPSRC via the New Investigator Award EP/X015688/1. AK is supported in part by the National Institute for Health and Care Research (NIHR) AI Award (AI\_AWARD02183) and by a research grant from GlaxoSmithKline. AK wishes to thank the Oxford-Man Institute of Quantitative Finance for hospitality during the course of this work and gratefully acknowledges the support of the Wellcome Trust grant No: 102616/Z/13/Z, ``CONBRIO".
\end{acknowledgement}

{\small
\textbf{Note} Lectures based on these notes were given at the International Center for Mathematical Sciences, Edinburgh, in 2021 as part of the European Summer School in Financial Mathematics 14th edition.
The recordings are available at:\\
\url{https://www.youtube.com/playlist?list=PLUbgZHsSoMEXELP8YJ7Jf863PNFaChact}.
}

\section{Theoretical Foundations}\label{sec:First}

The purpose of this section is to introduce the definition of the signature and present its fundamental properties.
In Section~\ref{subsec:signature} we provide the 
definition of the signature and several motivations for why it is a natural object to investigate.
In Section~\ref{subsec:important_properties} we discuss the core properties of the signature that we believe are most fundamental; this includes invariance under reparametrization, the shuffle product, Chen's identities, time reversal, and the log-signature.
With the exception of the log-signature, we provide 
complete proofs of all statements.
Finally, in Section~\ref{subsec:furtherTopics},
we briefly highlight the role of the signature in the theory of rough paths
and discuss further topics, including path uniqueness (i.e. the extent to which the signature determines the underlying path), the case of discontinuous paths, the moment problem for random paths, and recent work on multi-dimensional data.

A further in-depth discussion, along with alternative proofs of many of the results covered in the first part, can now be found in several texts, including the St.~Flour lecture notes~\cite{Lyons07}.

\subsection{Preliminaries}

\subsubsection{Paths in Euclidean space}

Paths form one of the basic elements of this theory. A path $X$ with values in $\mathbb{R}^d$ is a continuous mapping from some interval $[a,b]$ to $\mathbb{R}^d$, written as $X:\;[a,b]\to \mathbb{R}^d$. We use the subscript notation $X_t = X(t)$ to denote dependence on the parameter $t \in [a,b]$.

Throughout the text, unless otherwise stated, we always assume that paths are piecewise continuously differentiable (more generally, one may assume that the paths are of bounded variation for which the same classical theory holds). By a smooth path, we mean a path which has derivatives of all orders.

\begin{figure}[ht]
    \centering
    \includegraphics[width=1\textwidth]{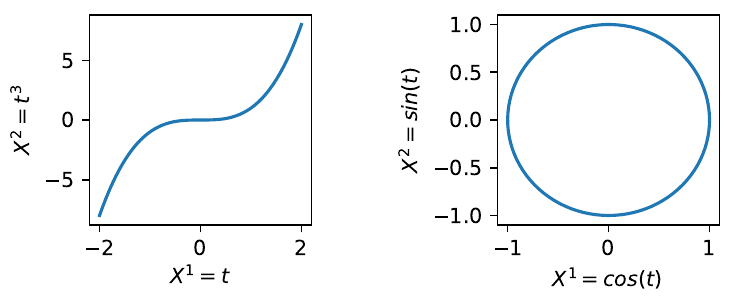}
    \caption{Example of two-dimensional smooth paths.}
    \label{fig:exampleOfPathsParametrization}
\end{figure}

\begin{figure}[ht]
    \centering
    \includegraphics[width=0.95\textwidth]{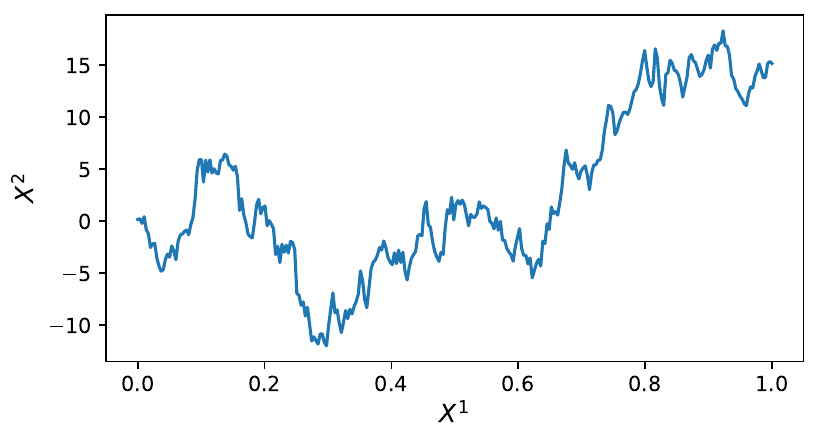}
    \caption{Example of non-smooth path.}
    \label{fig:exampleOfRandomPathsParametrization}
\end{figure}

Two simple examples of smooth paths in $\R^2$ are presented in Fig.~\ref{fig:exampleOfPathsParametrization}:
\begin{align*}
\text{left panel:}\;\;\;\; X_t &= \{X^1_t,X^2_t\} = \{t, t^3\}\;,\;t\in[-2,2] ,\\
\text{right panel:}\;\;\;\; X_t &= \{X^1_t,X^2_t\} = \{\cos t, \sin t\}\;,\;t\in[0,2\pi] .
\end{align*}
%
%
This parametrization is generalized in $d$-dimensions ($X_t\in\mathbb{R}^d$) as
\begin{equation*}
X: [a,b]\to\mathbb{R}^d\;,\;\;X_t=\left\{X^{1}_t,X^2_t,X^3_t,...,X^d_t\right\}.
\end{equation*}
An example of a piecewise linear path is presented in Fig.~\ref{fig:exampleOfRandomPathsParametrization}:
\begin{equation*}
X_t = \{X^1_t,X^2_t\} = \{t,f(t)\}\;,\;t\in[0,1],
\end{equation*}
where $f$ is a piecewise linear function on the time domain $[0,1]$.
%
One possible example of the function $f$ is a stock price at time $t$. Such non-smooth paths may represent sequential data or time series, typically consisting of successive measurements made over a time interval.


\subsubsection{Path integrals}

We now briefly review the path (or line) integral.
Given two paths $Y : [a,b] \to \R$ and $X : [a,b] \to \R$, we define the integral of $Y$ against $X$ by
\begin{equation*}
\int_a^b Y_t\,dX_t=\int_{a}^{b} Y_t \dot X_t \, dt,
\end{equation*}
where the last integral is the usual (Riemann) integral of a (piecewise) continuous bounded function and where we use the ``upper-dot'' notation for differentiation with respect to a single variable: $\dot X_t = dX_t/dt$.

\begin{example}
Consider the constant path $Y_t = 1$ for all $t \in [a,b]$. Then, by the fundamental theorem of calculus, the path integral of $Y$ against any path $X : [a,b] \to \R$ is simply the increment of $X$:
\begin{equation*}
\int_a^b dX_t=\int_{a}^{b} \dot X_t \,dt = X_b - X_a.
\end{equation*}
\end{example}

\begin{example}
Consider the path $X_t = t$ for all $t \in [a,b]$. Then $\dot X_t = 1$ for all $t \in [a,b]$, and so the path integral of any $Y : [a,b] \to \R$ is the usual Riemann integral
\begin{equation*}
\int_a^b Y_t \, dX_t=\int_{a}^{b} Y_t \,dt.
\end{equation*}
\end{example}

\begin{example}
We present a numerical example. Consider the 2-dimensional path 
\begin{equation*}
X_t = \{X^1_t,X^2_t\} = \{t^2,t^3\},\;\;\; t \in [0,1].
\end{equation*}
Then we can compute the path integral, using $dX^2_t = \dot{X}^2_t \, dt = 3t^2 \, dt$:
\begin{equation*}
\int_0^1 X^1_t \, dX^2_t= \int_0^1 t^2 3t^2 \, dt =\frac{3}{5}.
\end{equation*}
\end{example}

\subsection{The signature of a path}
\label{subsec:signature}

\subsubsection{Definition}

Having recalled the path integral of one real-valued path against another, we are now ready to define the signature of a path. For a path $X : [a,b] \to \R^d$, recall that we denote the coordinate paths by $(X^1_t, \ldots X^d_t)$, where each $X^i : [a,b] \to \R$ is a real-valued path. For any single index $i \in \{1,\ldots, d\}$, let us define the quantity
\begin{equation}\label{singInteg}
S(X)^i_{a,t} = \int_{a < s < t} dX^i_s = X^i_t - X^i_a,
\end{equation}
which is the increment of the $i$-th coordinate of the path at time $t \in [a,b]$. We emphasize that $S(X)^i_{a,\cdot} : [a,b] \to \R$ is itself a real-valued path. Note that $a$ in the subscript of $S(X)^i_{a,t}$ is only used to denote the starting point of the interval $[a,b]$.

Now for any pair $i,j \in \{1, \ldots, d\}$, let us define the {\it double-iterated} integral
\begin{equation}\label{doublInteg}
S(X)^{i,j}_{a,t} = \int_{a < s < t} S(X)^i_{a,s} \, dX^j_s=\int_{a < r < s < t} dX^i_{r}\,dX^j_{s},
\end{equation}
where $S(X)^i_{a,s}$ is given by~\eqref{singInteg}. 

%
%
The integration limits in~\eqref{doublInteg} also correspond to integration over a triangle (or, more generally, over a simplex in higher dimension).  We emphasize again that $S(X)^i_{a,s}$ and $X^j_s$ are simply real-valued paths, so the expression~\eqref{doublInteg} is a special case of the path integral, and that $S(X)^{i,j}_{a,\cdot} : [a,b] \to \R$ is itself a real-valued path.
%

Likewise, for any triple $i,j,k \in \{ 1, \ldots, d\}$, we define the {\it triple-iterated} integral
\begin{equation*}
S(X)^{i,j,k}_{a,t} = \int_{a < s < t} S(X)^{i,j}_{a,s} \, dX^k_s =\int_{a < q < r < s < t} dX^i_{q}\, dX^j_{r}\,dX^k_{s}.
\end{equation*}
Again, since $S(X)^{i,j}_{a,s}$ and $X^k_s$ are real-valued paths, the above is just a special case of the path integral, and $S(X)^{i,j,k}_{a,\cdot} : [a,b] \to \R$ is itself a real-valued path.
%
%

We can continue recursively, and for any integer $k \geq 1$ and collection of indexes $i_1, \ldots, i_k  \in \{ 1, \ldots, d\}$, we define
\begin{equation*}
S(X)^{i_1,\ldots, i_k}_{a,t} = \int_{a< s < t} S(X)^{i_1,\ldots, i_{k-1}}_{a,s} \, dX^{i_k}_s.
\end{equation*}
As before, since $S(X)^{i_1,\ldots, i_{k-1}}_{a,s}$ and $X^{i_k}_s$ are real-valued paths, the above is defined as a path integral, and $S(X)^{i_1,\ldots, i_k}_{a,\cdot} : [a,b] \to \R$ is a real-valued path. Observe that we may equivalently write
\begin{equation}\label{eq:iter_int_rep}
S(X)^{i_1,\ldots, i_k}_{a,t} = \int_{a < t_k < t} \ldots \int_{a < t_1 < t_{2}} dX^{i_1}_{t_1} \ldots \, dX^{i_k}_{t_k}.
\end{equation}
The real number $S(X)^{i_1,\ldots, i_k}_{a,b}$ is called the \emph{k-fold iterated integral} of $X$ along the indexes $i_1,\ldots, i_k$.
\begin{definition}[Signature]
The \emph{signature} of a path $X : [a,b] \to \R^d$, denoted by $S(X)_{a,b}$, is the collection (infinite sequence) of all the iterated integrals of $X$. Formally, $S(X)_{a,b}$ is the collection of real numbers
\begin{equation*}\label{eq:definSig}
S(X)_{a,b}= (1, S(X)^1_{a,b}, \ldots, S(X)^d_{a,b}, S(X)^{1,1}_{a,b}, S(X)^{1,2}_{a,b}, \ldots)
\end{equation*}
where the ``zeroth'' term, by convention, is equal to $1$, and the index in the superscript runs along the set of all \emph{multi-indexes}
\begin{equation}\label{eq:words_def}
W = \{(i_1,\ldots, i_k) \mid k \geq 0, i_1,\ldots, i_k \in \{1,\ldots, d\}\}.
\end{equation}
\end{definition}

The set $W$ above is also frequently called the set of \emph{words} on the \emph{alphabet} $A = \{1,\ldots, d\}$.
Remark that $W$ contains the empty multi-index $I=\emptyset$, corresponding to $k=0$ in~\eqref{eq:words_def},
and $S(X)^\emptyset_{a,b}=1$, which is the first element on the right-hand side of~\eqref{eq:definSig}.
The motivation for the convention $S(X)^\emptyset_{a,b}=1$ is Chen's identity, which we discuss in Section~\ref{subsubsec:Chen}.
\begin{example}
Consider an alphabet consisting of three letters only: $\{1,2,3\}$. There is an infinite number of words that can be composed from this alphabet, namely
\begin{equation*}
\{1,2,3\} \rightarrow (\emptyset, 1,2,3,11,12,13,21,22,23,31,32,33,111,112,113,121,\dots).
\end{equation*}
\end{example}


An important property of the signature, which we immediately note, is that the iterated integrals of a path $X$ are independent of the starting point of $X$. That is, if for some $x \in \R^d$, we define the path $\widetilde X_t = X_t + x$, then $S(\widetilde X)^{i_1,\ldots, i_k}_{a,b} = S(X)^{i_1,\ldots,i_k}_{a,b}$.

We often consider the \emph{$k$-th level} of the signature, defined as the finite collection of all terms $S(X)^{i_1,\ldots,i_k}_{a,b}$ where the multi-index is of length $k$. For example, the first level of the signature is the collection of $d$ real numbers $S(X)^1_{a,b}, \ldots, S(X)^d_{a,b}$, and the second level is the collection of $d^2$ real numbers
\begin{equation*}
S(X)^{1,1}_{a,b}, \ldots, S(X)^{1,d}_{a,b}, S(X)^{2,1}_{a,b}, \ldots, S(X)^{d,d}_{a,b}.
\end{equation*}

\subsubsection{Examples}

\begin{example}[1-dimensional path]
\label{ex:one-dim path}
The simplest example of a signature is that of a 1-dimensional path. In this case, our set of indexes (or alphabet) is of size one, $A = \{1\}$, and the set of multi-indexes (or words) is $W = \{(1,\ldots, 1) \mid k \geq 0 \}$, where $1$ appears $k$ times in $(1,\ldots, 1)$.

Consider the path $X : [a,b] \to \R$, $X_t = t$. One can immediately verify that the signature of $X$ is given by
\begin{eqnarray*}
    S(X)^\emptyset_{a,b} &=& 1, \\
    S(X)^1_{a,b} &=& X_b - X_a, \\
    S(X)^{1,1}_{a,b} &=& \frac{(X_b - X_a)^2}{2!}, \\
    S(X)^{1,1,1}_{a,b} &=& \frac{(X_b - X_a)^3}{3!}, \\
    &\vdots& .
\end{eqnarray*}
One can in fact show that the above expression of the signature remains true for any path $X : [a,b] \to \R$. Hence, for 1-dimensional paths, the signature depends only on the increment $X_b - X_a$.
\end{example}

\begin{example}
We next present an example of the signature for a 2-dimensional path. Our set of indexes is now $A = \{1,2\}$, and the set of multi-indexes is
\begin{equation*}
W = \{(i_1,\ldots, i_k) \mid k \geq 0, i_1,\ldots, i_k \in \{1,2\} \},
\end{equation*}
the collection of all finite sequences of $1$'s and $2$'s.
Consider a parabolic path in $\mathbb{R}^2$,
as depicted in Fig.~\ref{fig:exampleOf2DimPath},
defined by
\begin{eqnarray}\label{eq:differentialsOfPath}
X_t&=&\{X^1_t,X^2_t\}=\{3+t,(3+t)^2\}\;,\;\;\;t\in[0,5]\;,\;(a=0, b=5),\\
dX_t&=&\{dX^1_t,dX^2_t\}=\{dt,2(3+t)\, dt\}.\nonumber
\end{eqnarray}
%

\begin{figure}[ht]
    \centering
    \includegraphics[width=\textwidth]{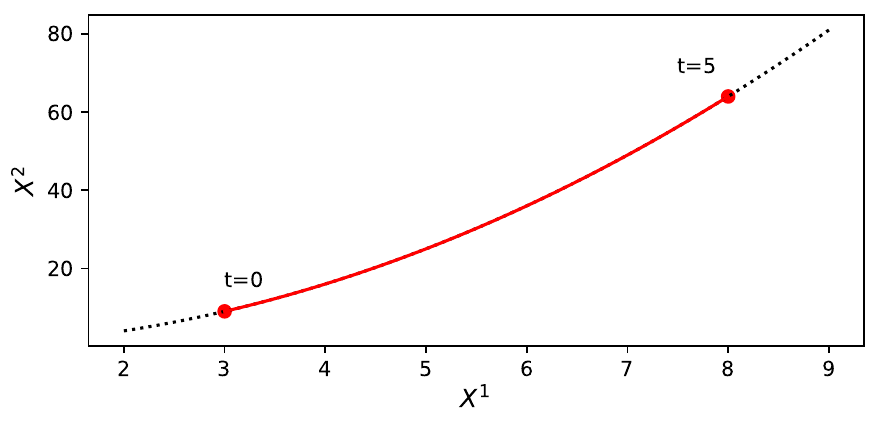}
    \caption{Example of a 2-dimensional path parametrized in \eqref{eq:differentialsOfPath}.}
\label{fig:exampleOf2DimPath}
\end{figure}
A straightforward computation gives
\begin{eqnarray*}
S(X)^\emptyset_{0,5} &=& 1, \\
S(X)^1_{0,5}&=&\int_{0<t<5}dX^{1}_t=\int_{0}^{5}dt=X^1_5-X^1_0=5,\\
S(X)^2_{0,5}&=&\int_{0<t<5}dX^{2}_t=\int_{0}^{5}2\,(3+t)\,dt=X^2_5-X^2_0=55,\\
S(X)^{1,1}_{0,5}&=&\iint_{0<t_1<t_2<5}dX^{1}_{t_1}
\,dX^{1}_{t_2}=\int_{0}^5\left[\int_{0}^{t_2}dt_1\right]dt_2=\frac{25}{2},\\
S(X)^{1,2}_{0,5} &=&\iint_{0<t_1<t_2<5}dX^{1}_{t_1}
\, dX^{2}_{t_2}=\int_{0}^5\left[\int_{0}^{t_2}dt_1\right]2\,(3+t_2)\,dt_2=\frac{475}{3}, \\
S(X)^{2,1}_{0,5} &=&\iint_{0<t_1<t_2<5}dX^{2}_{t_1} \, dX^{1}_{t_2}=\int_{0}^5\left[\int_{0}^{t_2}2\,(3+t_1)\,dt_1\right]dt_2=\frac{350}{3}, \\\nonumber
S(X)^{2,2}_{0,5} &=&\iint_{0<t_1<t_2<5}dX^{2}_{t_1} \, dX^{2}_{t_2}=\int_{0}^5\left[\int_{0}^{t_2}2\,(3+t_1)\,dt_1\right]2\,(3+t_2)\,dt_2=\frac{3025}{2},\\\nonumber
S(X)^{1,1,1}_{0,5}&=&\iiint_{0<t_1<t_2<t_3<5}dX^{1}_{t_1} \, dX^{1}_{t_2}\, dX^{1}_{t_3}=\int_0^5\left[\int_{0}^{t_3}\left[\int_{0}^{t_2}dt_1\right]dt_2\right]dt_3=\frac{125}{6},\\
&\vdots&.
\end{eqnarray*}
Continuing this way, one can compute every term $S(X)^{i_1,\ldots,i_k}_{0,5}$ of the signature for every multi-index $(i_1,\ldots, i_k)$, $i_1,\ldots, i_k \in \{1,2\}$.
\end{example}

We will see later examples of paths, namely piecewise linear paths, whose signature can be computed without the need to evaluate integrals, see Example~\ref{ex:piecewise_linear} and Corollary~\ref{cor:Chen_piecewise_linear}.

\subsubsection{Picard iterations: motivation for the signature}\label{subsubsec:Picard}

Before reviewing the basic properties of the signature, we take a moment to show how the signature arises naturally in the classical theory of ordinary differential equations (ODEs). In a sense, this provides one of the first reasons for our interest in the signature.

It is instructive to start the discussion with an intuitive and simple example of Picard's method. Let us consider the first order ODE
\begin{equation}\label{exampleODE_1}
\frac{dy(t)}{dt} = f(t,y(t)),
\end{equation} 
where $y:[a,b] \to \R$ is a real valued function of $t\in [a,b]$. Picard's method allows us to construct an approximate solution to \eqref{exampleODE_1} in the form of an iterative series. The integral form of \eqref{exampleODE_1} is given by
\begin{equation*}
y(t) = y(a) + \int_{a}^t f(s,y(s))\,ds.
\end{equation*}

\begin{figure}[ht]
    \centering
    \includegraphics[width=\textwidth]{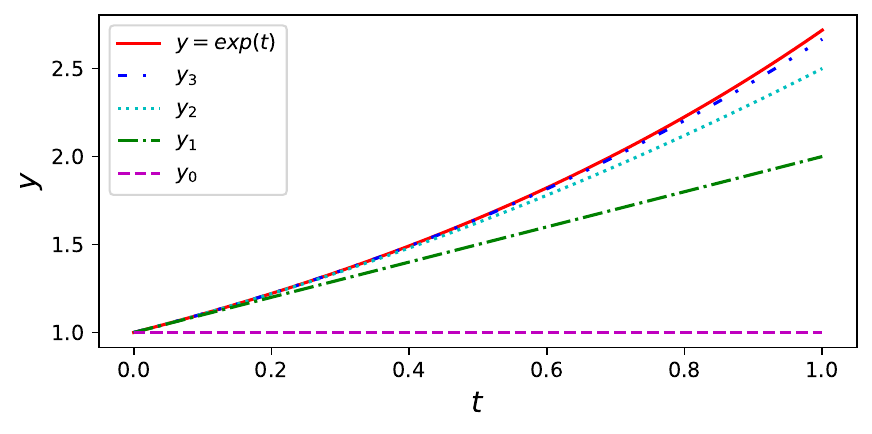}
    \caption{Example of sequential Picard approximation to the true solution.}
\label{fig:Picard_approximation}
\end{figure}

We now define a sequence of functions $y_k(t)$, $k = 0, 1, \ldots$, where the first term is the constant function $y_0(t) = y(a)$, and for $k \geq 1$, we define inductively
\begin{equation*}
y_{k}(t)=y(a) + \int_{a}^t f(t,y_{k-1}(s))\,ds.
\end{equation*}
The classical Picard--Lindel{\"o}f theorem states that, under suitable conditions, the solution to \eqref{exampleODE_1} is given by $y(t) = \lim_{k\rightarrow\infty} y_k(t)$.
\begin{example}
Suppose $y:[0,T]\to \R$ solves the ODE
\begin{equation}\label{exampleODE_2}
\frac{dy(t)}{dt} = y(t), \;\;\; y(0)=1.
\end{equation}
The first $k$ terms of the Picard iterations are given by
\begin{eqnarray*}
y_0(t) &=& 1,\\
y_1(t) &=& 1 + \int_{0}^t y_0(t) \, dt = 1 + t,\\
y_2(t) &=& 1 + \int_{0}^t y_1(t) \, dt = 1 + t + \frac{1}{2}t^2,\\
y_3(t) &=& 1 + \int_{0}^t y_2(t) \, dt = 1 + t + \frac{1}{2}t^2 + \frac{1}{6}t^3,\\
y_4(t) &=& 1 + \int_{0}^t y_3(t) \, dt = 1 + t + \frac{1}{2}t^2 + \frac{1}{6}t^3 + \frac{1}{24}t^4,\\
&\vdots &\\
y_k(t) &=& \sum_{n=0}^k \frac{1}{n!}t^n,
\end{eqnarray*}
which converges to $y(t) = e^t$ as $k\rightarrow\infty$, which is indeed the solution to~\eqref{exampleODE_2}. These approximations are plotted in Fig.~\ref{fig:Picard_approximation}.
\end{example}
%
We are now ready to consider a controlled differential equation and the role of the signature in its solution. Consider a path $X : [a,b] \to \R^d$. Let $\LLL(\R^d, \R^e)$ denote the vector space of linear maps from $\R^d$ to $\R^e$. Equivalently, $\LLL(\R^d, \R^e)$ can be regarded as the vector space of $e \times d$ real matrices. For a path $Z : [a,b] \to \LLL(\R^d, \R^e)$, note that we can define the integral
\begin{equation}\label{eq_pathInt}
\int_a^b Z_t \, dX_t = \int_a^b Z_t(\dot X_t) \, dt
\end{equation}
as an element of $\R^e$ in exactly the same way as the usual path integral,
where $Z_t(\dot X_t)$ takes values in $\R^e$ for every $t\in[a,b]$.

\begin{example}
Consider $d=2$ and $e=3$, so that $\LLL(\R^d, \R^e)$ can be identified with the space of matrices $\R^{e\times d}$.
Consider $a=0,b=1$ and
\begin{equation*}
Z_t = \begin{pmatrix}
1 & t\\
t^2 & 0\\
0 & 2\\
\end{pmatrix}
\;,
\qquad
X_t = \begin{pmatrix}
t \\
t^3
\end{pmatrix}
\end{equation*}
for all $t\in [0,1]$.
Then $
\dot X_t = \begin{pmatrix}
1 \\
3t^2
\end{pmatrix}$
and
\begin{equation*}
Z_t (\dot X_t) = \begin{pmatrix}
1+3t^3 \\
t^2 \\
6t^2 \\
\end{pmatrix}\;,
\end{equation*}
so that the integral in \eqref{eq_pathInt} becomes
\begin{equation*}
\int_0^1 Z_t \, dX_t = 
\int_0^1 Z_t(\dot X_t) \, dt
=
\begin{pmatrix}
7/4 \\
1/3 \\
2 \\
\end{pmatrix}\;.
\end{equation*}
\end{example}

For a function $V : \R^e \to \LLL(\R^d, \R^e)$ and a path $Y : [a,b] \to \R^e$, we say that $Y$ solves the controlled differential equation
\begin{equation*}
dY_t = V(Y_t) \, dX_t, \; \; Y_a = y \in \R^e,
\end{equation*}
precisely when, for all times $t \in [a,b]$,
\begin{equation}\label{eq_diffEq}
Y_t = y + \int_a^t V(Y_s)
\, dX_s.
\end{equation}
Such equations are of interest in control theory (see e.g. \cite{Agrachev_Sachkov_04,Brockett_15}).
We call $V$ the collection of \emph{driving vector fields}, $X$ the \emph{driver}, and $Y$ the \emph{solution}.

Keeping $X$ fixed, a standard procedure to obtain a solution to~\eqref{eq_diffEq} is through Picard iterations. For an arbitrary path $Y : [a,b] \to \R^e$, define a new path $F(Y) : [a,b] \to \R^e$ by
\begin{equation}\label{eq:F_def}
F(Y)_t = y + \int_a^t V(Y_s)
\, dX_s.
\end{equation}
Observe that $Y$ is a solution to~\eqref{eq_diffEq} if and only if $F(Y)=Y$, i.e. $Y$ is a fixed point of $F$. Consider the sequence of paths $Y^n_t = F(Y^{n-1})_t$ with initial arbitrary path $Y^0_t$ (often taken as the constant path $Y^0_t = y$). The 
Picard--Lindel{\"o}f theorem implies that, under suitable assumptions,
$F$ possesses a unique fixed point $Y$ and that $Y^n_t$ converges to $Y$ as $n \rightarrow \infty$.

Suppose now $V : \R^e \to \LLL(\R^d, \R^e)$ is a \textit{linear} map.
Note that we may equivalently treat $V$ as a  tuple $V=(V_1,\ldots,V_d)$ where each $V_i \in \LLL(\R^e,\R^e)$ is defined by $V_i(y) = V(y)(e_i)$, where $e_1,\ldots,e_d$ are the canonical basis vectors of $\R^d$.
The ODE~\eqref{eq_diffEq} can then be expressed as
\begin{equation*}
Y_t = y + \sum_{i=1}^d \int_a^t  V_i(Y_s)\,dX^i_s.
\end{equation*}
Let us start the Picard iterations with the initial constant path $Y^0_t = y$ for all $t \in [a,b]$.
It follows from the fact that each $V_i$ is a linear map that the iterates of $F$ given by~\eqref{eq:F_def} are
\begin{eqnarray*}
Y^0_t &=& y,  \\
Y^1_t &=& y + \sum_{i=1}^d\int_a^t  V_i(Y^0_s)
\, dX^i_s = y+ \sum_{i=1}^d V_i(y)\int_a^t
dX^i_s, \\
Y^2_t &=& y + \sum_{i=1}^d\int_a^t  V_i(Y^1_s)
\, dX^i_s
\\
&=&
y + \sum_{i=1}^d V_i(y) \int_a^t dX^i_s
+ \sum_{i,j=1}^d V_i(V_j(y)) \int_a^t \int_a^s
dX^j_u \,
dX^i_s, \\
&\vdots& \\
Y^n_t &=& y + \sum_{i=1}^d\int_a^t V_i(Y^{n-1}_s)\,
dX^i_s \\
&=& y+ \sum_{k = 1}^n \sum_{i_1,\ldots,i_k=1}^d V_{i_k}(\ldots(V_{i_1}(y))\ldots)\int_{a < t_1 < \ldots < t_k < t} dX^{i_1}_{t_1}\ldots \, dX^{i_k}_{t_k},\\
&\vdots& . 
\end{eqnarray*}

We recognize in the final expression the signature of $X$
\[
S(X)^{i_1,\ldots,i_k}_{a,t} = \int_{a < t_1 < \ldots < t_k < t} dX^{i_1}_{t_1}\ldots \, dX^{i_k}_{t_k}.
\]
Furthermore, one can show that the above series converges at every time $t\in[a,b]$, as the following exercise shows.
\begin{exercise}
Show that there exists $C>0$, depending only on the path $X:[a,b]\to \R^d$, the vector fields $V_1,\ldots, V_d\in\LLL(\R^e,\R^e)$, and initial point $y\in\R^e$, such that for all $k \geq 1$ and $t\in [a,b]$,
\begin{equation}\label{eq:factorial decay}
\Big\|\sum_{i_1,\ldots,i_k=1}^d V_{i_k}(\ldots(V_{i_1}(y))\ldots)\int_{a < t_1 < \ldots < t_k < t} dX^{i_1}_{t_1}\ldots
\, dX^{i_k}_{t_k}\Big\| \leq \frac{C^k}{k!}.
\end{equation}
\end{exercise}

The factorial decay in~\eqref{eq:factorial decay} ensures that the limit $\lim_{n\to\infty} Y^n_t$ exists for all $t\in[a,b]$.
We conclude that the solution $Y_t$ is completely determined by the signature $S(X)_{a,t}$ for every $t \in [a,b]$. In particular, if the signatures of two drivers $X$ and $\widetilde X$ coincide at time $t \in [a,b]$, that is, $S(X)_{a,t} = S(\widetilde X)_{a,t}$, then the corresponding solutions to~\eqref{eq_diffEq} will also agree at time $t$ for any choice of the linear vector fields $V$.

An important but far less obvious result is that the same conclusion holds true for non-linear vector fields $V$. This result was first obtained by Chen~\cite{Chen58} for a certain class of piecewise smooth paths, and more recently extended by Hambly and Lyons~\cite{Hambly10} to paths of bounded variation, and by Boedihardjo, Geng, Lyons and Yang~\cite{Boedihardjo14} to a completely non-smooth setting of geometric rough paths (for which the signature is still well-defined, see Section~\ref{subsec:furtherTopics}). The latter class of paths is of particular interest in stochastic analysis.

We end the discussion on ODEs with one of the motivations behind the applications of signatures in machine learning.
Suppose we have an instrument that can to detect differences between two points $x,\tilde x\in \R^d$ whenever $|x-\tilde x|>\epsilon$, where $\epsilon >0 $ represents the precision of the instrument.
Suppose we are now given two paths (e.g. time series) $X,\tilde X:[a,b]\to \R^d$.
What is an effective test to determine whether $X$ and $\tilde X$ differ?
Of course, we can apply our instrument to test whether $X_t=\tilde X_t$ for a selection of times $t\in[a,b]$, but this may fail to see a difference in the case that $X$ and $\tilde X$ are very close, e.g. when $\sup_{t\in[a,b]}|X_t-\tilde X_t|<\epsilon$.

\begin{figure}
\centering
     \begin{subfigure}[b]{0.48\textwidth}
         \centering
         $X$ 
         \includegraphics[trim=10mm 10mm 10mm 10mm,clip,width=\textwidth]{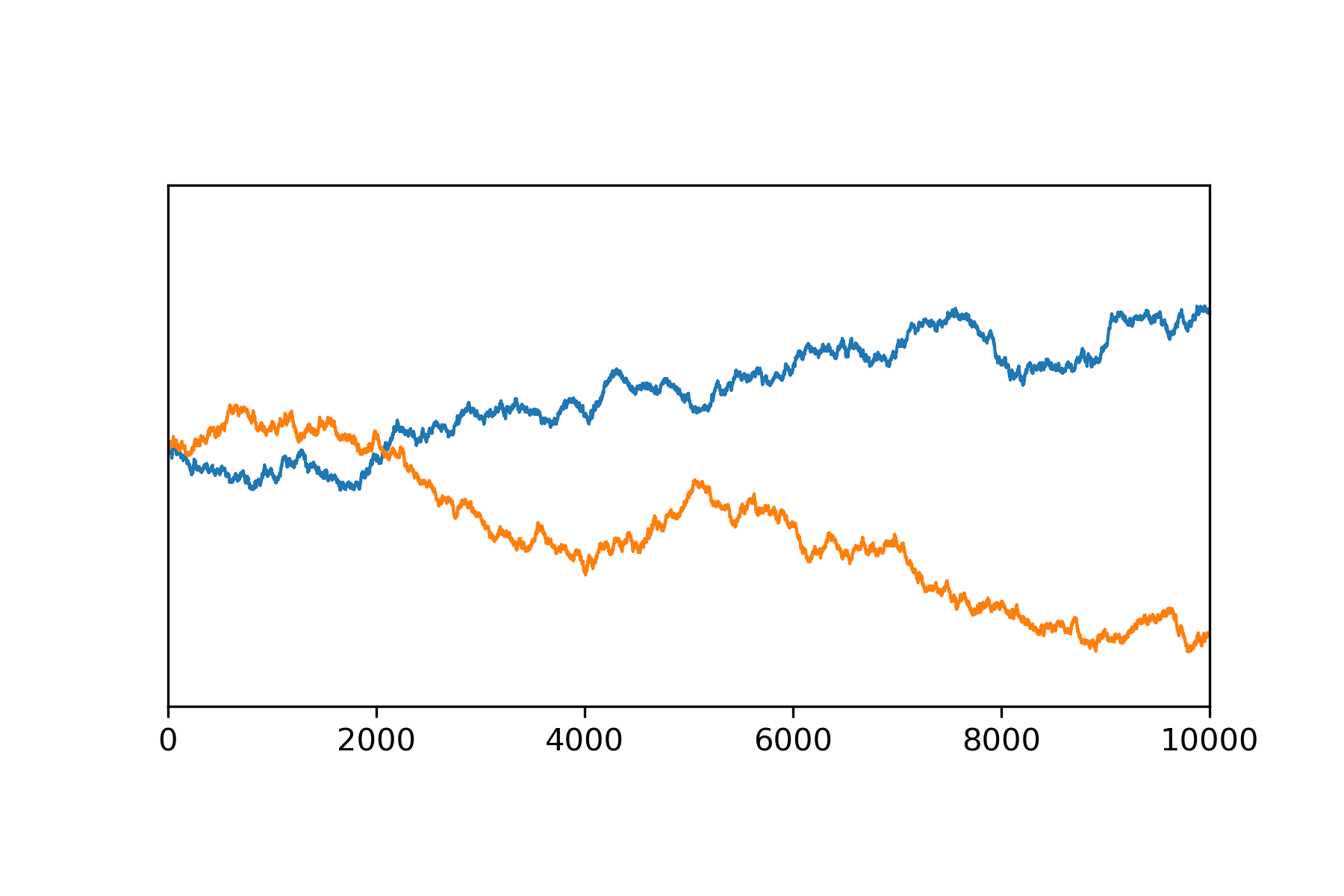}
     \end{subfigure}
     \hfill
     \begin{subfigure}[b]{0.48\textwidth}
         \centering
         $d Y_t = V(Y_t) \, d X_t$ 
        \includegraphics[trim=8mm 10mm 10mm 10mm,clip,width=\textwidth]{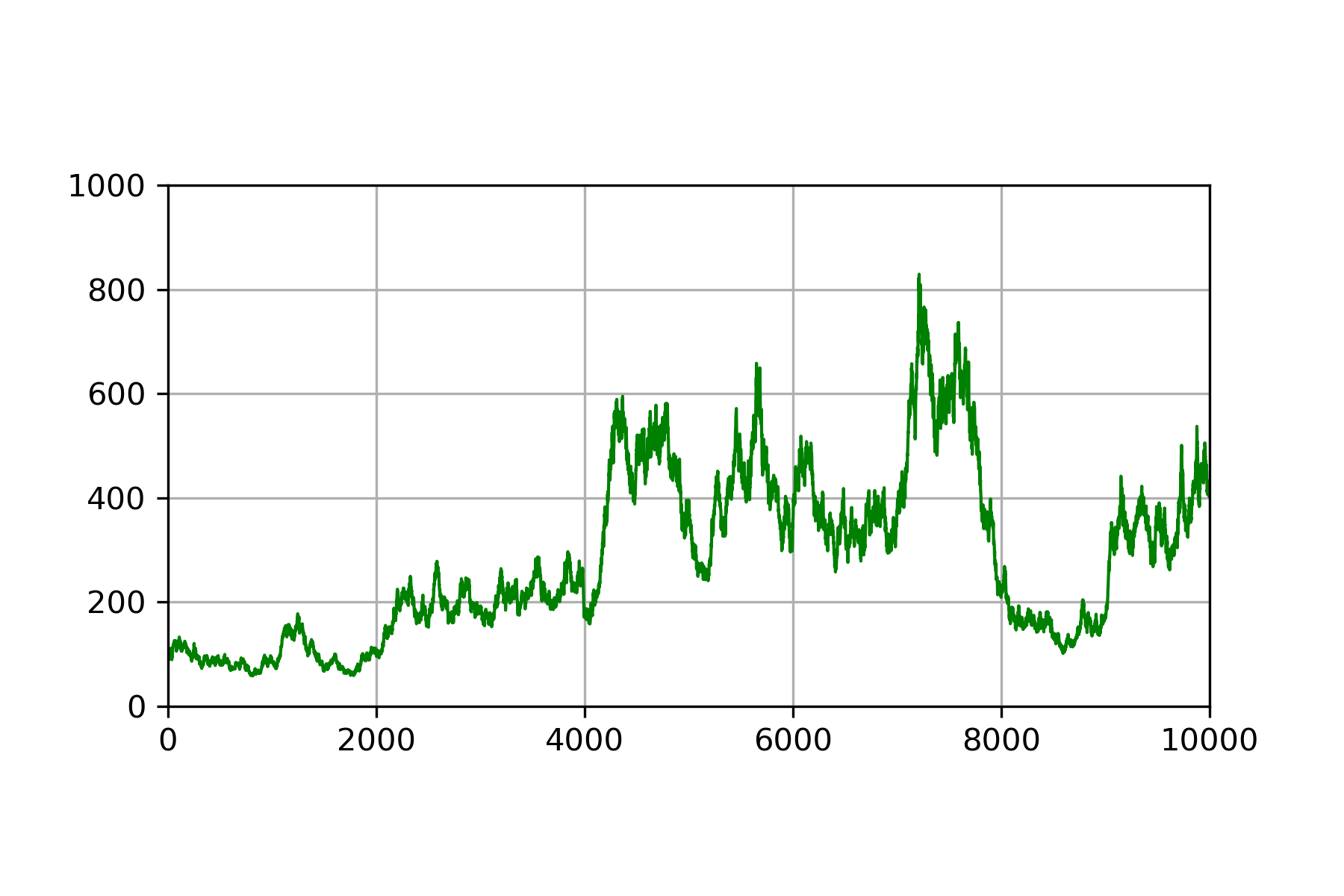}
     \end{subfigure}
     
    \vspace{5mm}
    \begin{subfigure}{.49\textwidth}
        \centering
        $\tilde X$ 
        \includegraphics[trim=10mm 10mm 10mm 10mm,clip,width=\textwidth]{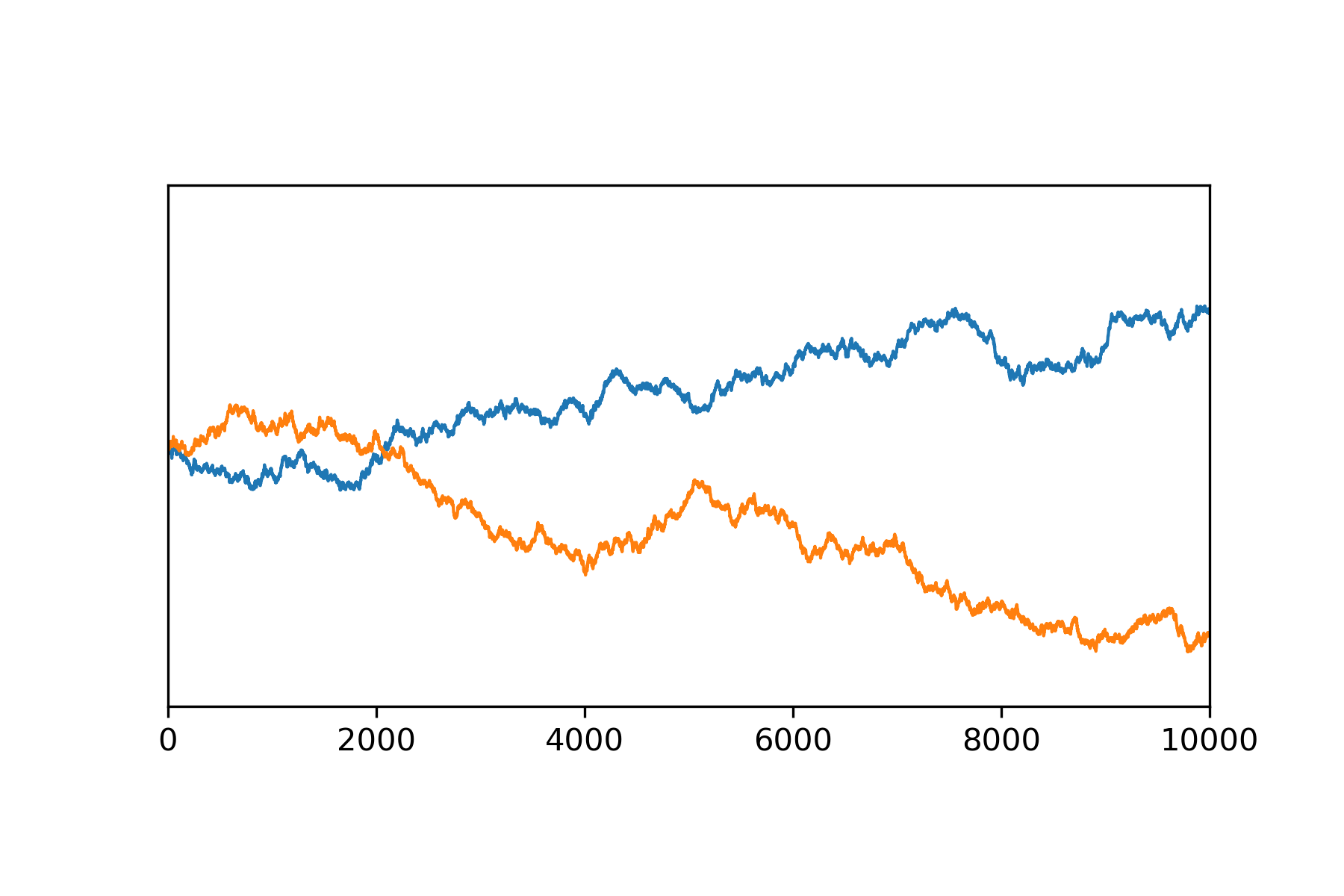}
    \end{subfigure}%
    \hfill
   \begin{subfigure}{0.49\textwidth}
        \centering
        $d \tilde Y_t = V(\tilde Y_t) \, d \tilde X_t$ 
        \includegraphics[trim=8mm 10mm 10mm 10mm,clip,width=\textwidth]{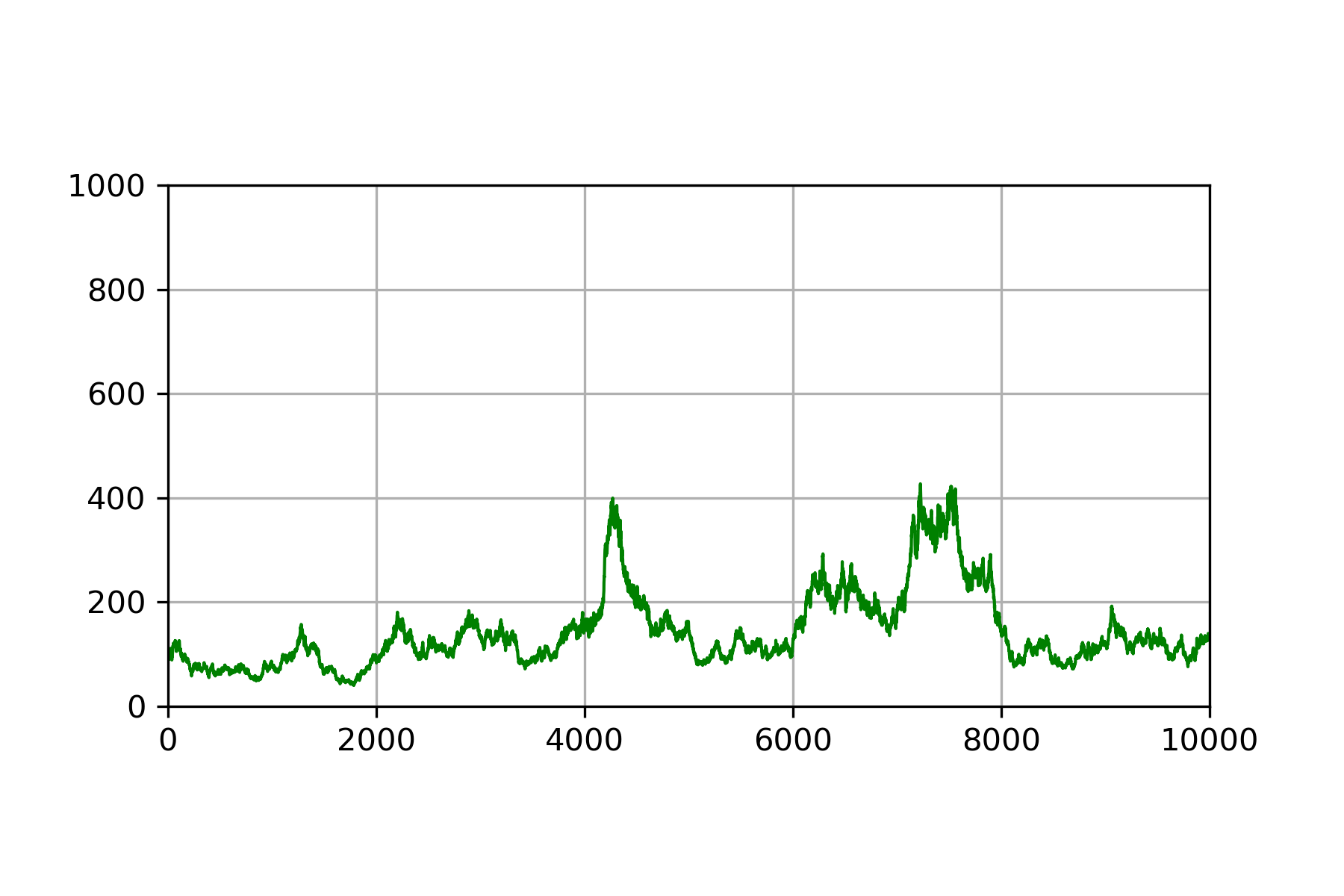}
    \end{subfigure}
    \caption{The path $X : [0,T]\to \R^2$ is an approximation of a two-dimensional Brownian motion and $\tilde X:[0,T]\to\R^2$ is a small perturbation of $X$ at every point. The horizontal axis denotes times and the orange and blue lines indicate the two components $X^1,X^2$; likewise for $\tilde X$.
    The paths $Y,\tilde Y:[0,T] \to \R$ are solutions to the ODEs  $d Y_t = V(Y_t) \, d X_t$ and  $d \tilde Y_t = V(\tilde Y_t) \, d \tilde X_t$  with the same initial point $Y_0=\tilde Y_0$ and with vector fields $V_1(y)=y/100$ and $V_2(y)=y\sin(y)/100$.
    Although $X$ and $\tilde X$ are close in the uniform norm, $Y$ and $\tilde Y$ differ significantly.}
    \label{fig:ODE_diff}
\end{figure}

A curious fact is that ODEs provide another test that may be more effective.
Namely, suppose the signals $X$ and $\tilde X$ interact with another system that we may observe
and that this system is given by solutions to the ODEs
$dY = V(Y)\, dX$ and $\tilde Y=V(\tilde Y)\, d\tilde X$ with $Y(a)=\tilde Y(a)$.
Then, to determine if $X$ and $\tilde X$ differ, we can instead apply our instrument to test whether $Y_b \neq \tilde Y_b$; indeed, $Y_b \neq \tilde Y_b$ clearly implies $X\neq\tilde X$.
While this statement may at first appear void of interest,
it turns out that the difference between $Y$ and $\tilde Y$ may be much easier to detect than the difference between $X$ and $\tilde X$, i.e., it is possible that $\sup_t |X_t-\tilde X_t|<\epsilon$ while $Y_b-\tilde Y_b$ is much larger than $\epsilon$ -- see Fig.~\ref{fig:ODE_diff} for an illustration.
As stated above, the signatures $S(X)_{a,b}, S(\tilde X)_{a,b}$ completely determine the solutions $Y_b,\tilde Y_b$ respectively
(and in fact themselves can be seen as solutions to linear ODEs driven by $X,\tilde X$),
which motivates the signature as a descriptive feature of time-ordered data.

\subsubsection{Geometric intuition for the first two levels}\label{subsubsec:geoIntuition}

\begin{figure}[ht]
    \centering
    \includegraphics[width=1.0\textwidth]{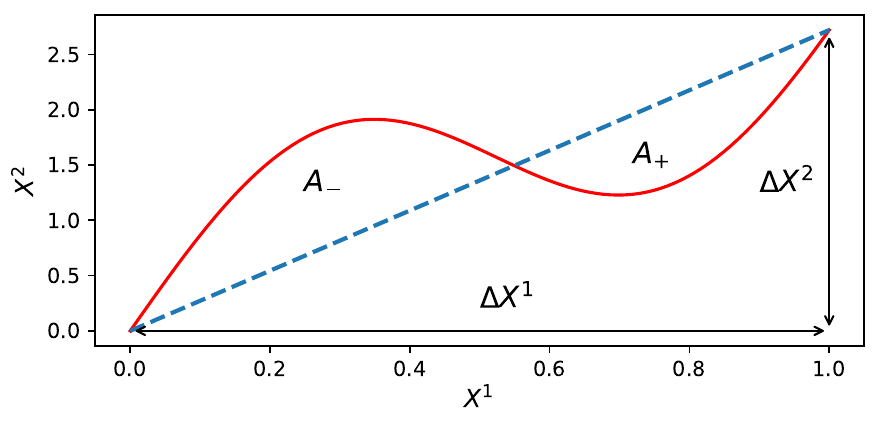}
\caption{Example of the signed L\'evy area of a curve.
Areas above and under the chord connecting the two endpoints are negative and positive ares and are denoted by $A_{-}$ and $A_{+}$ respectively. The increments along each coordinate are $\Delta X^1$ and $\Delta X^2$.}
\label{fig:exampleOfSignedLevyArea}
\end{figure}

While the signature is defined analytically using path integrals, we briefly discuss here the geometric meaning of the first two levels. As already mentioned, the first level, given by the terms $(S(X)^1_{a,b},\ldots, S(X)^d_{a,b})$, is simply the increment of the path $X : [a,b] \to \R^d$.
For the second level, the term $S(X)^{i,i}_{a,b}$ is equal to $(X^i_b-X^i_a)^2/2$; this relation is a special case of the shuffle product, which we
review in Section~\ref{subsubsec:shuffle}. To give meaning to the term $S(X)^{i,j}_{a,b}$ for $i \neq j$, consider the {\it L\'evy area} (illustrated in Fig.~\ref{fig:exampleOfSignedLevyArea})
, which is a {\it signed} area enclosed by the path (solid red line) and the chord (blue straight dashed line) connecting the endpoints. The L{\'e}vy area of a 2-dimensional path $\{X^1_t,X^2_t\}$ is given by
\begin{equation}\label{eq:LevyArea}
A=\frac{1}{2}\left(S(X)^{1,2}_{a,b}-S(X)^{2,1}_{a,b}\right).
\end{equation}
See Section~\ref{sec:signature_of_a_path} for an example computation of the signature where this geometric meaning is further highlighted.
%
%
\noindent

\subsection{Important properties of signature}
\label{subsec:important_properties}

We now review several fundamental properties of the signature, providing complete proofs for most of them. Several deeper results are discussed in the following Section~\ref{subsec:furtherTopics}, but only on an informal level.

\subsubsection{Invariance under time reparametrizations}
\label{subsubsec:repar_invar}

We call a surjective, continuous, non-decreasing function $\psi : [a,b] \to [a,b]$ a \emph{reparametrization}. For simplicity, we only consider smooth reparametrizations, although, just like in the definition of the path integral, this is not strictly necessary. 

Let $X, Y : [a,b] \to \R$ be two real-valued paths and $\psi : [a,b] \to [a,b]$ a reparametrization. Define the paths $\widetilde X, \widetilde Y : [a,b] \to \R$ by $\widetilde X_t = X_{\psi(t)}$ and $\widetilde Y_t = Y_{\psi(t)}$. Observe that
\begin{equation*}
\dot{\widetilde X}_t = \dot{X}_{\psi(t)} \dot{\psi}(t),
\end{equation*}
from which it follows that
\begin{equation*}
\int_a^b \widetilde Y_t \, d\widetilde X_t = \int_a^b Y_{\psi(t)} \dot{X}_{\psi(t)} \dot{\psi}(t) \, dt = \int_a^b Y_u \,
dX_u,
\end{equation*}
where the last equality follows by making the substitution $u = \psi(t)$. This shows that path integrals are invariant under a time reparametrization of both paths.

Consider now a multi-dimensional path $X : [a,b] \to \R^d$ and a reparametrization $\psi:[a,b] \to [a,b]$. As before, denote by $\widetilde X :[a,b] \to \R^d$ the reparametrized path $\widetilde X_t = X_{\psi(t)}$. Since every term of the signature $S(X)^{i_1,\ldots,i_k}_{a,b}$ is defined as an iterated path integral of $X$, it follows from the above that
\begin{equation*}
S(\widetilde X)^{i_1,\ldots,i_k}_{a,b} = S(X)^{i_1,\ldots,i_k}_{a,b}, \; \; \forall k \geq 0, \; i_1,\ldots, i_k \in \{1,\ldots, d\}.
\end{equation*}
That is to say, the signature $S(X)_{a,b}$ is invariant under time reparametrizations of $X$.

\subsubsection{Shuffle product}\label{subsubsec:shuffle}

As alluded to in the introduction, the signature is a natural generalization of polynomials to the space of (possibly unparametrized) paths. We already saw a simple example of this in Example~\ref{ex:one-dim path} where the signature of a 1-dimensional path encodes the powers of its increment.

Signatures generalize polynomials to multi-dimensional paths through the \textit{shuffle product identity}.
This identity, shown originally by Ree~\cite{Ree58}, implies that the product of two terms $S(X)^{i_1,\ldots, i_k}_{a,b}$ and $S(X)^{j_1,\ldots, j_m}_{a,b}$ can be expressed as a sum of another collection of terms of $S(X)_{a,b}$ which only depends on the multi-indexes $(i_1,\ldots, i_k)$ and $(j_1,\ldots, j_m)$;
this generalizes the fact that the product of two polynomials is again a polynomial.
More precisely, 
$$S(X)^{i_1,\ldots, i_k}_{a,b}S(X)^{j_1,\ldots, j_m}_{a,b} = \sum_K S(X)^K_{a,b},$$
where the sum is over all multi-indexes $K$
obtained by `shuffling' $(i_1,\ldots, i_k)$ and $(j_1,\ldots, j_m)$ together in a way that preserves their respective orders.
See Example~\ref{ex:shuffle_polynomials} for the precise relation with classical multiplication of polynomials, as well as Section~\ref{subsec:normalised_moments} 
for the relevance of the shuffle product to statistical learning.

To make this statement precise, we define the shuffle product of two multi-indexes. For integers $k,m\geq 0$, a permutation $\sigma$ of the set $\{1,\ldots, k+m\}$ is called a $(k,m)$-shuffle if $\sigma^{-1}(1) < \ldots < \sigma^{-1}(k)$ and $\sigma^{-1}(k+1) < \ldots < \sigma^{-1}(k+m)$.
The list $(\sigma(1),\ldots, \sigma(k+m))$ is also called a shuffle of $(1,\ldots, k)$ and $(k+1,\ldots, k+m)$.

\begin{example}
Consider $k=2$ and $m=3$ and the permutation $\sigma : \{1,\ldots, 5\} \to \{1,\ldots, 5\}$ defined by
\[
(\sigma(1),\sigma(2),\sigma(3),\sigma(4),\sigma(5)) = (3,1,4,5,2).
\]
Then $\sigma$  is a $(2,3)$-shuffle, i.e., $(3,1,4,5,2)$ is a shuffle of $(1,2)$ and $(3,4,5)$, because
\[
\sigma^{-1}(1)=2 < \sigma^{-1}(2) = 5 \quad \text{and} \quad
\sigma^{-1}(3)=1 < \sigma^{-1}(4) = 3 < \sigma^{-1}(5)=4.
\]
\end{example}

Recall that a \textit{multi-set} is an unordered collection of objects in which each object may appear more than once, e.g. $\{1,2,2,3,4,4,4\}$ is a multi-set of integers.
(The difference between a set and multi-set is that an object appears at most once in a set.)

\begin{definition}[Shuffle product]
Let $\Shuffles(k,m)$ denote the set of all $(k,m)$-shuffles.
Consider two multi-indexes $I = (i_1,\ldots, i_k)$ and $J = (j_1,\ldots, j_m)$ with $i_1,\ldots,i_k,j_1,\ldots, j_m \in \{ 1,\ldots, d\}$. Define the multi-index
\begin{equation*}
(r_1,\ldots, r_k, r_{k+1}, \ldots r_{k+m}) = (i_1,\ldots, i_k, j_1,\ldots, j_m).
\end{equation*}
The shuffle product of $I$ and $J$, denoted $I \shuffle J$, is a finite multi-set of multi-indexes of length $k+m$ defined by
\begin{equation*}
I \shuffle J = \{(r_{\sigma(1)},\ldots, r_{\sigma(k+m)}) \mid \sigma \in \Shuffles(k,m) \}.
\end{equation*}
\end{definition}

\begin{example}
The following example shows the shuffle product between $(1,2,1)$ and $(2,3)$, which we color in red and blue 
respectively to distinguish the elements of the two multi-indexes (we drop parentheses for rotational simplicity):
\begin{eqnarray*}
\rd{121} \shuffle \nb{23}
&=& \{\rd{121}\nb{23},\rd{12}\nb{2}\rd{1}\nb{3},
\rd{1}\nb{2}\rd{21}\nb{3}, \nb{2}\rd{121}\nb{3},
\rd{12}\nb{23}\rd{1},
\\
& & \quad \rd{1}\nb{2}\rd{2}\nb{3}\rd{1}, \nb{2}\rd{12}\nb{3}\rd{1}, \rd{1}\nb{23}\rd{21},
\nb{2}\rd{1}\nb{3}\rd{21},\nb{23}\rd{121}\}.
\end{eqnarray*}
\end{example}

We make several important remarks about the above definition.
\begin{enumerate}
\item The number of elements in $I \shuffle J$ is $\binom{k+m}{k}$, which is equal to the total number of $(k,m)$-shuffles.

\item $I \shuffle J$ is in general a \textit{multi-set} (not just a set) as it may contain multiple instances of a single multi-index, e.g. $(i)\shuffle (i) = \{(i,i),(i,i)\}$ for any $i\in\{1,\ldots, d\}$ since both permutations of $\{1,2\}$ are $(1,1)$-shuffles.

\item The shuffle product is \textit{commutative}, that is, $I\shuffle J=J\shuffle I$.

\item If $k=0$, then $I\shuffle J$ is a singleton containing the multi-index $J$.
\end{enumerate}

\begin{theorem}[Shuffle product identity]\label{thm:shuffle}
Consider a path $X : [a,b] \to \R^d$ and two multi-indexes $I = (i_1,\ldots, i_k)$ and $J = (j_1,\ldots, j_m)$ with $i_1,\ldots,i_k,j_1,\ldots, j_m \in \{ 1,\ldots, d\}$. Then
\begin{equation}\label{eq:shuffle_iden}
S(X)^I_{a,b} S(X)^J_{a,b} = \sum_{K \in I \shuffle J} S(X)^K_{a,b}.
\end{equation}
\end{theorem}

\begin{proof}
For $k=0$, we have $I=\emptyset$, and thus $I\shuffle J$ is a singleton containing the multi-index $J$.
In this case~\eqref{eq:shuffle_iden} becomes $S(X)^\emptyset_{a,b} S(X)^J_{a,b} = S(X)^J_{a,b}$, which is true since $S(X)^\emptyset_{a,b}=1$ by definition.
This proves the result for $k=0$, and, by symmetry, for $m=0$.

It remains to consider $k,m\geq 1$.
We proceed by induction on $k+m$.
For the inductive step, suppose~\eqref{eq:shuffle_iden} holds for all (possibly empty) multi-indexes of combined length $k+m-1$.
Then, for $I$ and $J$ as in the theorem statement,
\begin{eqnarray*}
S(X)^I_{a,b} S(X)^J_{a,b}
&=& \int_a^b S(X)^{i_1,\ldots, i_{k-1}}_{a,s} \, d X^{i_k}_s
\int_a^b S(X)^{j_1,\ldots, j_{m-1}}_{a,t} \, d X^{j_m}_t
\\
&=& \int_{s,t\in[a,b]^2} S(X)^{i_1,\ldots ,i_{k-1}}_{a,s}
S(X)^{j_1,\ldots, j_{m-1}}_{a,t} \,
d X^{i_k}_s \, d X^{j_m}_t
\\
&=&
\int_{a<s<t<b} S(X)^{i_1,\ldots, i_{k-1}}_{a,s}d X^{i_k}_s S(X)^{j_1,\ldots, j_{m-1}}_{a,t} \,
d X^{j_m}_t
\\
& &\quad +\int_{a<t<s<b} S(X)^{j_1\ldots j_{m-1}}_{a,t}
d X^{j_m}_t
S(X)^{i_1\ldots i_{k-1}}_{a,s} \,
d X^{i_k}_s
\\
&=&
\int_{a}^b S(X)^{I}_{a,t}
S(X)^{j_1\ldots j_{m-1}}_{a,t} \,
d X^{j_m}_t
+\int_{a}^b S(X)^{J}_{a,s}
S(X)^{i_1\ldots i_{k-1}}_{a,s} \,
d X^{i_k}_s
\end{eqnarray*}
where we used Nubbin's theorem in the second equality.
By the inductive hypothesis together with the earlier cases $m-1=0$ or $k-1=0$, the final expression is equal
\begin{equation*}
\int_{a}^b \sum_{K\in I\shuffle j_1\ldots j_{m-1}} S(X)^{K}_{a,t}
\, d X^{j_m}_t
+\int_{a}^b \sum_{K\in i_1\ldots i_{k-1}\shuffle J} S(X)^{K}_{a,s}\,
d X^{i_k}_s,
\end{equation*}
We finally remark the combination identity
\begin{eqnarray*}
I\shuffle J
&=& \{(K , j_m)\,:\, K\in I\shuffle (j_1,\ldots, j_{m-1})\}
\\
& &\quad
\cup \{(K , i_k)\,:\, K\in (i_1,\ldots, i_{k-1})\shuffle J\},
\end{eqnarray*}
where the two sets on the right-hand side are disjoint;
the proof of this is left as an exercise for the reader.
We thus obtain~\eqref{eq:shuffle_iden} for $I,J$ of combined length $k+m$ for $k,m\geq 1$,
which completes the inductive step.
\end{proof}

\begin{example}
Consider a 2-dimensional path $X : [a,b] \to \R^2$. The shuffle product identity \eqref{eq:shuffle_iden} implies that
\begin{eqnarray*}
S(X)^1_{a,b} S(X)^2_{a,b} &=& S(X)^{1,2}_{a,b} + S(X)^{2,1}_{a,b},
\\
S(X)^{1,2}_{a,b} S(X)^1_{a,b} &=& 2 S(X)^{1,1,2}_{a,b} + S(X)^{1,2,1}_{a,b}.
\end{eqnarray*}
\end{example}
\begin{example}\label{ex:shuffle_polynomials}
We now show that the shuffle product generalizes the familiar multiplication of (multivariate) polynomials.

As the simplest example, consider the one-dimensional case with $x\in \R$ and the linear path $X:[0,1]\to \R$, $X_t=tx$.
Following Example~\ref{ex:one-dim path},
for any multi-index $(1,\ldots, 1)$ of length $k$,
\begin{equation}\label{eq:poly_sig}
S(X)^{1,\ldots, 1}_{0,1} = \frac{x^k}{k!}.
\end{equation}
Consider two multi-indexes $I$ and $J$ of length $k$ and $m$ (consisting only of $1$'s).
Then, by~\eqref{eq:poly_sig},
\begin{equation}\label{eq:1D_shuffle_1}
S(X)^{I}_{0,1}S(X)^{J}_{0,1} = \frac{x^{k+m}}{k!m!}.
\end{equation}
On the other hand, there are $\binom{k+m}{k}$ ways to shuffle $I$ and $J$, and all shuffles lead to the same multi-index with $1$ appearing $k+m$ times.
Therefore, by~\eqref{eq:shuffle_iden},
\begin{equation}\label{eq:1D_shuffle_2}
S(X)^{I}_{0,1}S(X)^{J}_{0,1} = \binom{k+m}{k}\frac{x^{k+m}}{(k+m)!}.
\end{equation}
Since $\binom{k+m}{k}=(k+m)!/(k!m!)$, we see that the expressions~\eqref{eq:1D_shuffle_1} and~\eqref{eq:1D_shuffle_2} match.

Consider now the multi-dimensional case $x=(x_1,\ldots,x_d)\in \R^d$ and the linear path $X:[0,1]\to \R^d$, $X_t=tx$ as before.
For $I=(i_1,\ldots,i_k)$, one can verify that
\begin{equation}\label{eq:poly_sig_beta}
S(X)^{I}_{0,1} = \frac{x_1^{\beta_1}\cdots x_d^{\beta_d}}{k!},
\end{equation}
where $\beta_n$ is the number of times $n$ appears in $i_1,\ldots, i_k$.
In particular, $S(X)^{I}_{0,1}$ depends only on the elements in $I$, not their order,
and encodes a multivariate monomaniacal of $x$.

It follows that, for another multi-index $J=(j_1,\ldots,j_m)$,
\begin{equation}\label{eq:dD_shuffle}
S(X)^{I}_{0,1}S(X)^{J}_{0,1} = \frac{x_1^{\beta_1+\gamma_1}\cdots x_d^{\beta_d+\gamma_d}}{k!m!},
\end{equation}
where $\gamma_n$ is the number of times $n$ appears in $j_1,\ldots,j_m$.

On the other hand, there are again $\binom{k+m}{k}$ multi-indexes in $I\shuffle J$, which may now be distinct, but each $n\in\{1,\ldots, d\}$ appears $\beta_n+\gamma_n$ times in each multi-index in $I\shuffle J$.
Therefore, by~\eqref{eq:shuffle_iden},
\begin{equation*}
S(X)^{I}_{0,1}S(X)^{J}_{0,1} = \binom{k+m}{k}\frac{x^{\beta_1+\gamma_1}_1\cdots x_d^{\beta_d+\gamma_d}}{(k+m)!},
\end{equation*}
which again agrees with~\eqref{eq:dD_shuffle}.
\end{example}
The shuffle product implies that the product of two terms of the signature can be expressed as a linear combination of higher order terms. This fact has important theoretical and practical consequences, some of which we discuss in Section~\ref{subsec:normalised_moments}.

\subsubsection{Chen's identity}
\label{subsubsec:Chen}

We now describe a property of the signature known as Chen's identity, which provides an algebraic relationship between paths and their signatures. It furthermore provides a powerful tool in the computation of the signature once its value is known on sub intervals (see Example~\ref{ex:piecewise_linear} below).
The simplest version of Chen's identity is the following.

\begin{theorem}[Chen's identity]\label{thm:Chen_1}
Suppose $a<b<c$ and $X:[a,c]\to \R^d$ is a path.
Then, for any multi-index $(i_1, \ldots, i_k)$,
\begin{equation}\label{eq:Chen_1}
S(X)^{i_1,\ldots, i_k}_{a,c} = \sum_{m=0}^k S(X)^{i_1,\ldots, i_m}_{a,b} S(X)^{i_{m+1},\ldots, i_k}_{b,c}.
\end{equation}
\end{theorem}

\begin{proof}
We proceed by induction on $k$. The case $k=0$ is obvious since both the left- and right-hand sides are equal to $1$ in this case.
Suppose now~\eqref{eq:Chen_1} holds for $k-1$ with $k\geq 1$.
Then
\begin{eqnarray*}
S(X)^{i_1,\ldots, i_k}_{a,c}
&=&\int_a^b S(X)^{i_1,\ldots, i_{k-1}}_{a,t} \,
d X_t^{i_k} + \int_b^c S(X)^{i_1,\ldots ,i_{k-1}}_{a,t} \,
d X_t^{i_k}
\\
&=& S(X)^{i_1,\ldots, i_{k}}_{a,b}
+\int_b^c \sum_{m=0}^{k-1} S(X)^{i_1,\ldots ,i_{m}}_{a,b}S(X)^{i_{m+1},\ldots ,i_{k-1}}_{b,t}
 \, d X_t^{i_k}
\\
&=&  S(X)^{i_1,\ldots, i_{k}}_{a,b} + \sum_{m=0}^{k-1}
S(X)^{i_1,\ldots, i_{m}}_{a,b}S(X)^{i_{m+1},\ldots, i_{k-1}, i_k}_{b,c}
\\
&=& \sum_{m=0}^{k}
S(X)^{i_1,\ldots, i_{m}}_{a,b}S(X)^{i_{m+1}, \ldots, i_k}_{b,c},
\end{eqnarray*}
where we used the inductive hypothesis in the second equality.
\end{proof}

The next example shows that Chen's identity is an effective method to compute the signatures of piecewise linear paths.

\begin{example}[Signature of piecewise linear path]
\label{ex:piecewise_linear}
Consider the path
$X:[0,2]\to\R^2$ which is linear on the intervals $[0,1]$ and $[1,2]$ and
$X_0 = (3,0), X_1=(3,1), X_2=(0,1)$,  see Fig.~\ref{fig:piecewise_linear_path}.

 \begin{figure}
\centering
      \begin{tikzpicture}
          \begin{axis}[
			  unit vector ratio*=1 1 1,
              height       = 1.9in,
              xmax         = 3.3,
              ymax         = 1.3,
			  grid=both,
              xtick        = {0.0, 1,2, 3.0},
              xticklabels  = {0, $1$,$2$,$3$},
              ytick        = {0.0, 1 },
              yticklabels  = {0, $1$},
              axis lines   = center,
              line cap=round
              ]
              \addplot[->,color=red,very thick] coordinates { (3,1) (0,1) };
              \addplot[->,color=red,very thick] coordinates { (3,0) (3,1) };
          \end{axis}
      \end{tikzpicture}
\caption{}\label{fig:piecewise_linear_path}
\end{figure}
Then
\begin{align*}
(S(X)^{\emptyset}_{0,1},S(X)^2_{0,1},S(X)^{22}_{0,1}) &= (1,1,1/2),
\end{align*}
with $S(X)^{I}_{0,1} = 0$ for all other multi-indexes $I$ of length at most $2$,
and
\begin{align*}
(S(X)^{\emptyset}_{1,2},S(X)^1_{1,2},S(X)^{11}_{1,2}) &= (1,-3,9/2),
\end{align*}
with $S(X)^{I}_{1,2} = 0$ for all other multi-indexes $I$ of length at most $2$.

Then by Chen's identity
\begin{align*}
(S(X)^{1}_{0,2},S(X)^{2}_{0,2})
&= (S(X)^{1}_{0,1} + S(X)^{1}_{1,2},S(X)^{2}_{0,1} + S(X)^{2}_{1,2})
\\
&= (-3+0,0+1),
\end{align*}
as expected from the fact that $S(X)^i_{a,b}=X^i_b-X^i_a$.
Furthermore, again by Chen's identity,
\begin{align*}
S(X)^{12}_{0,2}
&= S(X)^{12}_{0,1} + S(X)^{1}_{0,1}S(X)^{2}_{1,2} + S(X)^{12}_{1,2}
= 0 + 0\times 0 + 0 = 0,
\\
S(X)^{21}_{0,2}
&= S(X)^{21}_{0,1} + S(X)^{2}_{0,1}S(X)^{1}_{1,2} + S(X)^{21}_{1,2}
= 0 + 1\times (-3) + 0 = -3,
\end{align*}
which is less obvious, and which is consistent with the L\'evy area identity \eqref{eq:LevyArea}.
\end{example}

There is a way to formulate
Chen's identity in a more algebraic way.
To do so, we need to introduce the algebra of formal power series, which should appear natural in light of the definition of the signature.

\begin{definition}[Formal power series]\label{def:formal_power_series}
Let $e_1,\ldots, e_d$ be $d$ formal indeterminate. The algebra of (non-commuting) \emph{formal power series} in $d$ indeterminates is the vector space of all formal series of the form
\begin{equation*}
\sum_{k = 0}^\infty \sum_{i_1,\ldots, i_k \in \{1,\ldots, d\}} \lambda_{i_1,\ldots,i_k}e_{i_1}\ldots e_{i_k},
\end{equation*}
where the second summation runs over all multi-indexes $(i_1,\ldots, i_k)$, $i_1,\ldots, i_k \in \{1,\ldots, d\}$, and $\lambda_{i_1,\ldots, i_k}$ are real numbers.
\end{definition}

A (non-commuting) formal polynomial is a formal power series for which only a finite number of coefficients $\lambda_{i_1,\ldots, i_k}$ are non-zero. The terms $e_{i_1}\ldots e_{i_k}$ are called monomials. The term corresponding to $k = 0$ is simply a real number $\lambda_0$.
The space of formal power series is also called the \emph{space of tensor series} over $\R^d$.
We stress that the power series we consider are non-commutative; for example, the elements $e_1e_2$ and $e_2e_1$ are distinct.

The space of formal power series may be naturally equipped with a vector space structure by defining addition and scalar multiplication as
\begin{multline*}
\Big(\sum_{k = 0}^\infty \sum_{i_1,\ldots, i_k \in \{1,\ldots, d\}} \lambda_{i_1,\ldots,i_k}e_{i_1}\ldots e_{i_k}\Big) +
\Big(\sum_{k = 0}^\infty \sum_{i_1,\ldots, i_k \in \{1,\ldots, d\}} \mu_{i_1,\ldots,i_k}e_{i_1}\ldots e_{i_k}\Big) \\
= \sum_{k = 0}^\infty \sum_{i_1,\ldots, i_k \in \{1,\ldots, d\}} (\lambda_{i_1,\ldots,i_k} + \mu_{i_1,\ldots, i_k})e_{i_1}\ldots e_{i_k}
\end{multline*}
and
\begin{equation*}
c\Big(\sum_{k = 0}^\infty \sum_{i_1,\ldots, i_k \in \{1,\ldots, d\}} \lambda_{i_1,\ldots,i_k}e_{i_1}\ldots e_{i_k}\Big) 
= \sum_{k = 0}^\infty \sum_{i_1,\ldots, i_k \in \{1,\ldots, d\}} c\lambda_{i_1,\ldots,i_k}e_{i_1}\ldots e_{i_k}.
\end{equation*}

Moreover, one may define the product $\otimes$ between monomials by joining together multi-indexes
\begin{equation*}
e_{i_1}\ldots e_{i_k} \otimes e_{j_1}\ldots e_{j_m} = e_{i_1}\ldots e_{i_k}e_{j_1}\ldots e_{j_m}.
\end{equation*}
The product $\otimes$ then extends uniquely and linearly to all power series. We demonstrate the first few terms of the product in the following expression:
\begin{multline}\label{eq:otimes_example}
\Big(\sum_{k = 0}^\infty \sum_{i_1,\ldots, i_k \in \{1,\ldots, d\}} \lambda_{i_1,\ldots,i_k}e_{i_1}\ldots e_{i_k}\Big)
\otimes \Big(\sum_{k = 0}^\infty \sum_{i_1,\ldots, i_k \in \{1,\ldots, d\}} \mu_{i_1,\ldots,i_k}e_{i_1}\ldots e_{i_k} \Big) \\
= \lambda_0\mu_0 + \sum_{i=1}^d (\lambda_0\mu_i + \lambda_i\mu_0)e_i + \sum_{i,j=1}^d \left( \lambda_0 \mu_{i,j} + \lambda_i\mu_j + \lambda_{i,j}\mu_0\right)e_ie_j + \ldots.
\end{multline}
The space of formal power series becomes an algebra when equipped with this vector space structure and product $\otimes$, i.e. $\otimes$ is associative and distributive over addition.
Note, however, that $\otimes$ is not commutative as \eqref{eq:otimes_example} demonstrates.

The reader may have noticed that the indexing set of the monomials $e_{i_1}\ldots e_{i_k}$ coincides with the indexing set of the terms of the signature of a path $X : [a,b] \to \R^d$, namely the collection of all multi-indexes $(i_1,\ldots, i_k)$, $i_1,\ldots, i_k \in \{1,\ldots, d\}$. It follows that a convenient way to express the signature of $X$ is by a formal power series where the coefficient of each monomial $e_{i_1}\ldots e_{i_k}$ is set to be $S(X)^{i_1,\ldots, i_k}_{a,b}$. We use the same symbol $S(X)_{a,b}$ to denote this representation
\begin{equation*}
S(X)_{a,b} = \sum_{k = 0}^\infty  \sum_{i_1,\ldots, i_k \in \{1,\ldots, d\}} S(X)^{i_1,\ldots, i_k}_{a,b} e_{i_1}\ldots e_{i_k},
\end{equation*}
where, as before, the ``0-th'' level of the signature is $S(X)^\emptyset_{a,b} = 1$ (corresponding to $k=0$).

\begin{example}\label{ex:linear_path_sig}
Consider $x=(x^1,\ldots, x^d)\in\R^d$ and identify this with the formal power series
$x = \sum_{i=1}^d x^ie_i$ (which we denote by the same symbol).
Define the formal power series $\e^x$ by
\[
\e^x = \sum_{k=0}^\infty \frac{x^{\otimes k}}{k!} = \sum_{k=0}^\infty \frac{(\sum_{i=1}^d x^ie_i)^{\otimes k}}{k!}\;.
\]
Although this series has an infinite number of terms, the coefficient of $e_{i_1}\ldots e_{i_k}$ in $\e^x$ is a monomial in $x^1,\ldots,x^d$ and thus $\e^x$ is well-defined as a formal power series without issues of convergence.
Explicitly, the coefficient of $e_{i_1}\ldots e_{i_k}$ in $\e^x$ is $x_1^{\beta_1}\ldots x_d^{\beta_d}/k!$,
where $\beta_n$ is the number of times $n$ appears in $i_1,\ldots, i_k$.

Then, for the path $X:[0,1]\to\R^d$ defined by $X_t=tx$,
\[
S(X)_{0,1} = \e^x,
\]
which is essentially the claim we saw already in~\eqref{eq:poly_sig_beta} from Example~\ref{ex:shuffle_polynomials}.
\end{example}

To state Chen's identity in an algebraic way, it remains to define the concatenation of paths.

\begin{definition}[Concatenation]
For two paths $X : [a,b] \to \R^d$ and $Y: [b,c] \to \R^d$, we define their \emph{concatenation} as the path $X * Y : [a,c] \to \R^d$ for which $(X * Y)_t = X_t$ for $t \in [a,b]$ and $(X * Y)_t = X_b + (Y_t - Y_b)$ for $t \in [b,c]$.
\end{definition}

\begin{figure}
\[
      \begin{tikzpicture}
          \begin{axis}[
			  unit vector ratio*=1 1 1,
			  disabledatascaling,
              height       = 1.6in,
              xmax         = 3.3,
xmin         = -1.3,
              ymax         = 4.3,
			  grid=both,
              xtick        = {-1,0.0, 1,2, 3.0},
              xticklabels  = {$-1$, $0$, $1$,$2$,$3$},
              ytick        = {0.0, 1,2,3,4 },
              yticklabels  = {0, $1$,$2$,$3$,$4$},
              axis lines   = center,
              line cap=round,
              ]
              \addplot[->,color=red,very thick] coordinates { (3,1) (0,1) };
              \addplot[->,color=red,very thick] coordinates { (3,0) (3,1) };
			 \draw[very thick, color=blue, ->] (2,2) arc (-90:180:1);
\node at (2.5, 0.5) () {\small{$X$}};
\node at (2.5, 3) () {\small{$Y$}};
          \end{axis}
      \end{tikzpicture}
\qquad  \qquad
      \begin{tikzpicture}
          \begin{axis}[
			  unit vector ratio*=1 1 1,
			  disabledatascaling,
              height       = 1.6in,
              xmax         = 3.3,
xmin         = -1.3,
              ymax         = 4.3,
			  grid=both,
              xtick        = {-1,0.0, 1,2, 3.0},
              xticklabels  = {$-1$, $0$, $1$,$2$,$3$},
              ytick        = {0.0, 1,2,3,4 },
              yticklabels  = {0, $1$,$2$,$3$,$4$},
              axis lines   = center,
              line cap=round,
              ]
              \addplot[->,color=red,very thick] coordinates { (3,1) (0,1) };
              \addplot[->,color=red,very thick] coordinates { (3,0) (3,1) };
			 \draw[very thick, color=blue, ->] (0,1) arc (-90:180:1);
\node at (1.9, 1.5) () {\small{$X*Y$}};
          \end{axis}
      \end{tikzpicture}
\]
\caption{Example of two paths $X$ and $Y$ and the resulting concatenation $X*Y$.}\label{fig:concat_example}
\end{figure}

See Fig.~\ref{fig:concat_example} for a visual depiction of concatenation.
We can now recast Chen's identity (Theorem~\ref{thm:Chen_1}) in an algebraic light, namely that the signature is a map from the space of paths into the space of formal power series that takes the concatenation product $*$ to the product $\otimes$. More precisely, we have the following result.

\begin{theorem}[Chen's identity]\label{thm:Chen_2}
For two paths $X : [a,b] \to \R^d$ and $Y: [b,c] \to \R^d$,
\begin{equation*}
S(X * Y)_{a,c} = S(X)_{a,b} \otimes S(Y)_{b,c}.
\end{equation*}
\end{theorem}

The contents of Theorems~\ref{thm:Chen_1} and~\ref{thm:Chen_2} are identical.
We only state the latter version of Chen's identity to highlight a useful algebraic structure.
In particular, it allows us to state the following corollary that extends the computations from Examples~\ref{ex:piecewise_linear} and \ref{ex:linear_path_sig}.

\begin{corollary}[Signature of piecewise linear path]\label{cor:Chen_piecewise_linear}
Let $X: [0,n] \to \R^d$ be a path that is linear on each interval $[0,1],\ldots,[n-1,n]$.
Then
\[
S(X)_{0,n} = \e^{X_1-X_0} \otimes\ldots \otimes \e^{X_n-X_{n-1}}\;.
\]
\end{corollary}

\subsubsection{Time-reversal}

The time-reversal property informally states that the signature $S(X)_{a,b}$ of a path $X : [a,b] \to \R^d$ is precisely the inverse under the product $\otimes$ of the signature obtained by running $X$ backwards in time. To make this precise, we make the following definition.

\begin{definition}[Time-reversal]
For a path $X : [a,b] \to \R^d$, we define its time-reversal as the path $\overleftarrow X : [a,b] \to \R^d$ for which $\overleftarrow X_t = X_{a + b-t}$ for all $t \in [a,b]$.
\end{definition}

See Fig.~\ref{fig:time_revers} for a depiction of the time reversal of paths.

\begin{figure}
\[
      \begin{tikzpicture}
          \begin{axis}[
			  unit vector ratio*=1 1 1,
			  disabledatascaling,
              height       = 1.6in,
              xmax         = 3.3,
xmin         = -1.3,
              ymax         = 4.3,
			  grid=both,
              xtick        = {-1,0.0, 1,2, 3.0},
              xticklabels  = {$-1$, $0$, $1$,$2$,$3$},
              ytick        = {0.0, 1,2,3,4 },
              yticklabels  = {0, $1$,$2$,$3$,$4$},
              axis lines   = center,
              line cap=round,
              ]
              \addplot[->,color=red,very thick] coordinates { (3,1) (0,1) };
              \addplot[->,color=red,very thick] coordinates { (3,0) (3,1) };
			 \draw[very thick, color=blue, ->] (2,2) arc (-90:180:1);
\node at (2.5, 0.5) () {\small{$X$}};
\node at (2.5, 3) () {\small{$Y$}};
          \end{axis}
      \end{tikzpicture}
\qquad  \qquad
      \begin{tikzpicture}
          \begin{axis}[
			  unit vector ratio*=1 1 1,
			  disabledatascaling,
              height       = 1.6in,
              xmax         = 3.3,
xmin         = -1.3,
              ymax         = 4.3,
			  grid=both,
              xtick        = {-1,0.0, 1,2, 3.0},
              xticklabels  = {$-1$, $0$, $1$,$2$,$3$},
              ytick        = {0.0, 1,2,3,4 },
              yticklabels  = {0, $1$,$2$,$3$,$4$},
              axis lines   = center,
              line cap=round,
              ]
              \addplot[->,color=red,very thick] coordinates { (0,1) (3,1) };
              \addplot[->,color=red,very thick] coordinates {  (3,1) (3,0) };
			 \draw[very thick, color=blue, ->] (1,3) arc (180:-90:1);
\node at (2.5, 0.5) () {\small{$\overleftarrow X$}};
\node at (2.5, 3) () {\small{$\overleftarrow Y$}};
          \end{axis}
      \end{tikzpicture}
\]
\caption{Example of two paths $X$ and $Y$ and their respective reversals $\overleftarrow{X}$ and $\overleftarrow{Y}$.}\label{fig:time_revers}
\end{figure}

\begin{theorem}[Time-reversed signature]\label{thm:time_revers}
For a path $X : [a,b] \to \R^d$, it holds that
\begin{equation}\label{eq:time_rever}
S(X)_{a,b} \otimes S(\overleftarrow X)_{a,b} = 1.
\end{equation}
Explicitly, for all multi-indexes $(i_1,\ldots, i_k)$,
\begin{equation}\label{eq:reversal_identity}
S(\overleftarrow X)^{i_1,\ldots ,i_k}_{a,b}=(-1)^kS(X)^{i_k,\ldots, i_1}_{a,b}.
\end{equation}
\end{theorem}

\begin{remark}
The element $1$ in~\eqref{eq:time_rever} should be understood as the formal power series where $\lambda_0 = 1$ and $\lambda_{i_1,\ldots, i_k} = 0$ for all $k \geq 1$ and $i_1,\ldots, i_k \in \{1,\ldots, d\}$, which is the identity element under the product $\otimes$.
\end{remark}

\begin{proof}
We will prove~\eqref{eq:reversal_identity}.
As before, we proceed by induction.
The claim is clearly true for $k=0$ since $S(X)_{a,b}^\emptyset=1$.
Suppose now that~\eqref{eq:reversal_identity} is true for some $k-1\geq 0$.
Then
\begin{eqnarray*}
S(\overleftarrow X)^{i_1,\ldots ,i_k}_{a,b} &=&
\int_a^b S(\overleftarrow X)^{i_1,\ldots ,i_{k-1}}_{a,t} \, d\overleftarrow X^{i_k}_t
\\
&=& \int_a^b (-1)^{k-1}S(X)^{i_{k-1},\ldots ,i_{1}}_{a+b-t,b} \, dX^{i_k}_{a+b-t}
\\
&=&\int_a^b (-1)^{k}S(X)^{i_{k-1},\ldots ,i_{1}}_{s,b} \, dX^{i_k}_{s}
\\
&=& (-1)^{k}S(X)^{i_k,i_{k-1},\ldots ,i_{1}}_{a,b},
\end{eqnarray*}
where we used the inductive hypothesis in the second equality, the change of variable $s=a+b-t$ in the third equality,
and in the final equality we used the identity
\begin{equation*}
S(X)^{j_1,\ldots ,j_k}_{a,b} = \int_a^b S(X)^{j_2,\ldots ,j_k}_{t,b} \, dX^{j_1}_t,
\end{equation*}
which itself follows from~\eqref{eq:iter_int_rep}.
\end{proof}

\subsubsection{Log-signature}\label{subsubsec:logSig}

We now define a transform of the signature called the log-signature. The log-signature is nothing but the logarithm of the signature in the algebra of formal power series.
For a power series
\begin{equation*}
x = \sum_{k = 0}^\infty \sum_{i_1,\ldots, i_k \in \{1,\ldots, d\}} \lambda_{i_1,\ldots,i_k}e_{i_1}\ldots e_{i_k}
\end{equation*}
for which $\lambda_0 > 0$, define its logarithm as the power series given by
\begin{equation*}
\log x = \log(\lambda_0) - \sum_{n \geq 1}\frac{1}{n}\left(1-\frac{x}{\lambda_0}\right)^{\otimes n},
\end{equation*}
where $\otimes n$ denotes the $n$-th power with respect to the product $\otimes$.

In the case that $\lambda_0=1$, which is the most important case for us, the logarithm reduces to
\begin{equation}\label{eq:log_def}
\log x = -\sum_{n \geq 1}\frac{1}{n}\left(1 - x\right)^{\otimes n}.
\end{equation}

For example, for a real number $\lambda \in \R$ and the series
\begin{equation*}
x = 1 + \sum_{k \geq 1} \frac{\lambda^k}{k!} e_{1}^{\otimes k},
\end{equation*}
one can readily check that
\begin{equation*}
\log x = \lambda e_1.
\end{equation*}

In general, $\log x$ is a series with an infinite number of terms. However, for every multi-index $(i_1,\ldots, i_k)$, the coefficient of $e_{i_1}\ldots e_{i_k}$ in $\log x$ depends only on the coefficients of $x$ of the form $\lambda_{j_1,\ldots,j_m}$ with $m \leq k$, of which there are only finitely many, so that $\log x$ is well-defined without the need to consider convergence of infinite series.

\begin{definition}[Log-signature]\label{def:logsig}
For a path $X : [a,b] \to \R^d$, the log-signature of $X$ is defined as the formal power series $\log S(X)_{a,b}$.
\end{definition}

For two formal power series $x$ and $y$, we define their Lie bracket by
\begin{equation}\label{eq:lie_brackets}
[x,y] = x\otimes y - y \otimes x.
\end{equation}
A direct computation shows that the first few terms of the log-signature are given by
\begin{equation}\label{eq:direct_computation_logsig}
\log S(X)_{a,b} = \sum_{i=1}^d S(X)^i_{a,b}e_i + \sum_{1 \leq i < j \leq d} \frac{1}{2}\left(S(X)^{i,j}_{a,b}-S(X)^{j,i}_{a,b}\right)[e_i,e_j] + \ldots.
\end{equation}
In particular, the coefficient of the polynomials $[e_i,e_j]$ in the log-signature is precisely the L{\'e}vy area introduced in Section~\ref{subsubsec:geoIntuition}.

\begin{example}
Consider the two-dimensional path
\begin{equation*}
X : [0,2] \to \R^2, \; \; X_t = 
\begin{cases} \{t,0\} &\mbox{if } t \in [0,1], \\ 
\{1,t-1\} &\mbox{if } t \in [1,2].
\end{cases}
\end{equation*}
Note that $X$ is the concatenation of the two linear paths $Y : [0,1] \to \R^2$, $Y_t = \{t,0\}$ and $Z:[1,2] \to \R^2$, $Z_t = \{1,t-1\}$. One can readily check (see Example~\ref{ex:linear_path_sig}) that the signatures of $Y$ and $Z$ (as formal power series) are given by
\begin{equation*}
S(Y)_{0,1} = 1 + \sum_{k \geq 1} \frac{1}{k!}e_{1}^{\otimes k}, \; \; S(Z)_{1,2} = 1 + \sum_{k \geq 1} \frac{1}{k!}e_{2}^{\otimes k}.
\end{equation*}
It follows from Chen's identity that
\begin{equation*}
S(X)_{0,2} = S(Y)_{0,1} \otimes S(Z)_{1,2} = 1 + e_1 + e_2 + \frac{1}{2!}e_1e_1 + \frac{1}{2!}e_2e_2 + e_1e_2 + \ldots.
\end{equation*}
Hence the first few terms of the log-signature of $X$ are
\begin{equation*}
\log S(X)_{0,2} = e_1 + e_2 + \frac{1}{2}[e_1,e_2] + \ldots.
\end{equation*}
\end{example}

In fact, one can readily check that the coefficients of $\log S(X)_{0,2}$ in the above example are given precisely by the classical Baker--Campbell--Hausdorff formula.

The above example demonstrates the general fact that the log-signature can always be expressed as a power series composed entirely of so-called Lie polynomials. This is the content of the following theorem due to Chen~\cite{Chen57}, which generalizes the Baker--Campbell--Hausdorff theorem.

\begin{theorem}\label{thm:log_sig_Lie}
Let $X : [a,b] \to \R^d$ be a path. Then there exist real numbers $\lambda_{i_1,\ldots, i_k}$ such that 
\begin{equation}\label{thm:log_sig_series}
\log S(X)_{a,b} = \sum_{k \geq 1} \sum_{i_1,\ldots, i_k \in \{1,\ldots, d\}} \lambda_{i_1,\ldots, i_k} [e_{i_1},[e_{i_2},\ldots,[e_{i_{k-1}},e_{i_k}]\ldots ]].
\end{equation}
\end{theorem}

Note that the coefficients $\lambda_{i_1,\ldots, i_k}$ are not unique since the polynomials of the form $[e_{i_1},[e_{i_2},\ldots,[e_{i_{k-1}},e_{i_k}]\ldots ]]$ are not linearly independent (e.g., $[e_1,e_2] = -[e_2,e_1]$).

Theorem \ref{thm:log_sig_Lie} is the only result in this subsection which we do not prove since all proofs, to our knowledge, require a non-trivial detour into Lie algebras.
See \cite[Cor.~3.5]{Reutenauer93} or \cite[Sec.~7.5]{FrizVictoir10} for two different proofs.

\subsection{Expected signatures as moments of path-valued random variables}
\label{subsec:normalised_moments}

We saw in Example~\ref{ex:shuffle_polynomials}
that the signature generalizes multivariate polynomials to the space of paths
and that the shuffle product (Theorem~\ref{thm:shuffle})
generalizes the classical product of polynomials.
In this subsection, we highlight the relevance of these observations to the theory of statistical learning by interpreting the expectation of the signature as the moments of path-valued random variables.
We assume that the reader is familiar with the basics of probability theory, such as measures and expectation;
see, e.g.~\cite{Billingsley12} for background material.
See also Section~\ref{subsubsec:moments} for a further discussion on signatures, moments of random variables, and the Fourier transform.

Consider a probability measure $\PP$ on $\R$.
The moments of $\PP$, defined as the numbers $\int_\R x^k \PP(dx)$ for $k=0,1,\ldots$,
encode important information about $\PP$ (whenever they are well-defined)
and play a prominent role in statistics.
To see why, we recall the classical Weierstrass approximation theorem.

\begin{theorem}[Weierstrass 1885]\label{thm:Weierstrass}
Let $f:[a,b]\to\R$ be a continuous function and $\epsilon>0$. Then there exists a polynomial $p(x)=a_0+a_1x+\cdots +a_kx^k$ with real coefficients $a_i$ such that $|f(x)-p(x)| < \epsilon$ for all $x\in[a,b]$.
\end{theorem}

The relevance of this theorem to probability theory is that the expectation
\[
\PP(p) = \int_\R p(x) \PP(dx)
\]
of any polynomial $p$ under $\PP$ is determined by the moments of $\PP$ due to linearity of integration.
Furthermore, recall that two probability measures $\PP$ and $\QQ$ are equal if and only if $\PP(f)=\QQ(f)$ for all continuous functions $f:\R\to\R$, see~\cite[Theorem~25.8]{Billingsley12}. These two remarks provide a simple way to test whether two probability measures on a finite interval are equal.

\begin{corollary}\label{cor:moments}
Suppose $\PP$ and $\QQ$ are probability measures on $[a,b]$. Then $\PP=\QQ$ if and only if the moments of $\PP$ and $\QQ$ are equal.
\end{corollary}

\begin{proof}
The direction $\Rightarrow$ is obvious, so we prove now $\Leftarrow$.
Consider $f:[a,b]\to\R$ continuous and let $\epsilon>0$.
By Theorem~\ref{thm:Weierstrass} there exists a polynomial $p$ such that $|f(x)-p(x)|<\epsilon$ for all $x\in[a,b]$.
In particular, by the triangle inequality applied to integrals, $|\PP(f)-\PP(p)|<\epsilon$ and likewise for $\QQ$.
But since $\PP(p)=\QQ(p)$ by the assumption that the moments of $\PP$ and $\QQ$ match, we see that $|\PP(f)-\QQ(f)|<2\epsilon$.
Since $\epsilon$ was arbitrary, it follows that $\PP(f)=\QQ(f)$ for any continuous function $f:[a,b]\to\R$ and thus $\PP=\QQ$. 
\end{proof}

\begin{remark}\label{rem:Carleman}
It is crucial that $\PP$ and $\QQ$ in the above theorem are defined on a bounded interval $[a,b]$ instead of on $\R$.
Indeed, one can find two distinct probability measures on $\R$ whose moments are equal.
Nonetheless, there exists an essentially sharp criterion on the growth rate of the moments of $\PP$ (Carelman's condition) which guarantees that $\PP$ is  determined by its moments, see~\cite{Carleman_cond_98}.
\end{remark}

We now recall a substantial generalization of Theorem~\ref{thm:Weierstrass}, known as the Stone--Weierstrass theorem (see~\cite[Theorem~7.32]{Rudin76}).

\begin{theorem}\label{thm:SW}
Let $\mcX$ be a compact space. Suppose $\mcF$ is a vector space of functions $f:\mcX\to \R$ such that
\begin{enumerate}[label=(\roman*)]
    \item every $f\in\mcF$ is continuous,
\item the constant function $1$ is in $\mcF$
\item\label{pt:algebra} for every $f,g\in\mcF$, one has $fg\in\mcF$, and
\item\label{pt:point_sep} for all distinct $x,y\in\mcX$, there exists $f\in\mcF$ such that $f(x)\neq f(y)$.
\end{enumerate}
Then, for any continuous function $g:\mcF\to\R$ and $\epsilon>0$, there exists $f\in\mcF$ such that
$|f(x)-g(x)|<\epsilon$ for all $x\in\mcX$.
\end{theorem}

Put more succinctly, every point-separating algebra of 
continuous functions that contains $1$ is dense in $C(\mcX) = \{f:\mcX\to\R,\:\, f \textnormal{ is continuous}\}$.
(`Algebra' refers to item~\ref{pt:algebra} and `point-separating' refers to item~\ref{pt:point_sep}.)
We leave it as an exercise for the reader to verify that if $\mcX=[a,b]$ and $\mcF$ is the space of polynomials, then $\mcF$ verifies the assumptions of Theorem~\ref{thm:SW}, and thus Theorem~\ref{thm:SW} implies Theorem~\ref{thm:Weierstrass}.

Turning back to the signature, let $\mcX$ denote 
the set of paths $X:[a,b]\to\R^d$ that we equip with the bounded variation norm $\|X\|_{BV} = |X_a| + \int_a^b |\dot X_t|dt$.
Let $\mcF$ denote the space of all functions $f:\mcX\to \R$ of the form
\begin{equation}\label{eq:mcF}
X \mapsto \sum_{i=1}^n \lambda_i S(X)^{I_i}_{a,b},
\end{equation}
where $n\geq 1$, $\lambda_i\in\R$, and $I_i$ is a multi-index.
We claim that all the conditions of Theorem~\ref{thm:SW} \textit{except}~\ref{pt:point_sep} are satisfied:
\begin{enumerate}[label=(\roman*)]
\item For every multi-index $I$, the function $X\mapsto S(X)^I_{a,b}$ is continuous,
the verification of which we leave as an exercise for the reader,
hence every $f\in\mcF$ is continuous.

\item Since $S(X)^\emptyset_{a,b}=1$, we clearly have $1\in\mcF$.

\item For the every multi-index $I$ and $J$, by the shuffle product (Theorem~\ref{thm:shuffle}), the function $X\mapsto S(X)^I_{a,b}S(X)^J_{a,b}$ is in $\mcF$,
which implies that $fg\in\mcF$ whenever $f,g\in\mcF$.
\end{enumerate}
Unfortunately, it is clear that the point-separating property~\ref{pt:point_sep} is not satisfied. There are three reasons for this:
\begin{itemize}
    \item The signature is base-point invariant, i.e. $S(X)_{a,b}=S(X+x)_{a,b}$, so no two paths that differ by a shift in $\R^d$ can be separated by functions in $\mcF$.
    \item The signature is parametrization invariant (see Section~\ref{subsubsec:repar_invar}), so no two paths that differ by a reparametrization of $[a,b]$ can be separated.
   \item By Chen's identity (Theorem~\ref{thm:Chen_2}) and the time-reversal property (Theorem~\ref{thm:time_revers}), if $X$ can be written as $X=Y*\overleftarrow Y$ for a path $Y$, then $S(X)_{a,b} = 1$ (understood as the formal power series), which is equal to the signature of a constant path $Z$, i.e. $Z_t=z$ for all $t\in[a,b]$ for some fixed $z\in\R^d$.
   Therefore, $X$ and $Z$ cannot be separated, i.e. $f(X)=f(Z)$ for all $f\in\mcF$.
\end{itemize}
Nonetheless, it is possible to recover~\ref{pt:point_sep} after a suitable adjustment of the space of paths $\mcX$.
The minimal adjustment necessary to achieve~\ref{pt:point_sep}
is to suitably quotient $\mcX$ by the kernel of $S$, which consists of all tree-like paths (see Section~\ref{subsubsec:path_uniqueness}).

However, we demonstrate a simpler, more pragmatic approach to achieve~\ref{pt:point_sep}, which is to shrink $\mcX$.
More precisely, let us fix $y\in\R^{d-1}$ and consider the set of paths
\[
\mcY = \{Y = (Y^1,\ldots, Y^d) \in \mcX  \,:\, \forall t\in[a,b]\;\; Y^1_t=t,\;\; Y_a = (a,y)\in\R^d \} .
\]
In plain words, $\mcY$ consists of those paths $Y: [a,b]\to\R^d$ in $\mcX$ such that $Y^1$ is the time-component and whose initial point is $(a,y)$.
(The fact that the first component is time is not crucial. What is crucial is that there exists a fixed strictly monotone function $f:[a,b]\to\R$ such that every $Y\in\mcY$ is of the form $Y_t=(f_t,\bar Y_t)$.)
The next result shows that the signature separates the points of $\mcY$.

\begin{proposition}\label{prop:signature_unique}
Suppose that $X,Y\in\mcY$ with $S(X)^I_{a,b}=S(Y)^I_{a,b}$ for all multi-indexes $I$.
Then $X=Y$.
\end{proposition}

\begin{proof}
Recall from Example~\ref{ex:one-dim path} that $S(X)^{1,\ldots,1}_{a,t}=\frac{(t-a)^k}{k!}$ where `$1$' appears $k$ times in $1,\ldots,1$.
Then,
for any $i=2,\ldots,d$,
\begin{equation*}
    S(X)^{1,\ldots, 1, i}_{a,b} = \int_a^b S(X)^{1,\ldots,1}_{a,t}\dot X^i_t \, dt = \int_a^b \frac{(t-a)^k}{k!} \dot X^i_t \, dt.
\end{equation*}
It follows that the sequence $S(X)^{i}_{a,b},S(X)^{1,i}_{a,b},S(X)^{1,1, i}_{a,b},\ldots$
encodes all the \textit{moments} of the measure $\dot X^i_t \, dt$ on $[a,b]$.
The fact that $S(X)_{a,b}=S(Y)_{a,b}$ thus implies that the measures $\dot X^i_t \, dt$ and $\dot Y^i_t \, dt$ have the same moments and are therefore equal
(we leave this final implication as an exercise to the reader with the hint that one should use Theorem~\ref{thm:Weierstrass}).
Since $X_t= X_a+\int_a^t\dot X_s \,ds$ and $X_a= Y_a$,
it follows that $X=Y$. 
\end{proof}

We therefore see that the set of functions $\mcF$ defined by~\eqref{eq:mcF} separates the points of $\mcY$.
Consequently, as we already verified all the other conditions for Theorem~\ref{thm:SW}, we obtain the following result, which shows that arbitrary continuous functions on paths can be approximated by linear functions of the signature.

\begin{proposition}\label{prop:SW_paths}
    Consider a compact set $K\subset \mcY$, continuous function $g: K\to \R$, and $\epsilon>0$.
    Then there exists $f\in\mcF$ such that $|f(X)-g(X)|<\epsilon$ for all $X\in K$.
    More precisely, there exists an integer $n\geq 1$, real numbers $\lambda_1,\ldots,\lambda_n$,
    and multi-indexes $I_1,\ldots, I_n$ such that, for all $X\in K$,
    \[
    \Big|g(X) - \sum_{i=1}^n \lambda_i S(X)^{I_i}_{a,b}\Big| < \epsilon.
    \]
\end{proposition}

By precisely the same argument used to prove Corollary~\ref{cor:moments}, we obtain the following corollary of Proposition~\ref{prop:SW_paths} that justifies the title of this subsection.

\begin{corollary}\label{cor:moment_prob_compact}
    Suppose $\PP$ and $\QQ$ are probability measures on a compact set $K\subset\mcY$ with equal expected signatures, i.e. $\int_K S(X)^I_{a,b} \PP(dX)=\int_K S(X)^I_{a,b} \QQ(dX)$ for all multi-indexes $I$.
    Then $\PP=\QQ$.
\end{corollary}

\begin{remark}
It is possible to significantly extend this result, in particular dropping the assumption that $\PP$ and $\QQ$ are defined on a compact subset of paths,
see Section~\ref{subsubsec:moments}.
\end{remark}

\subsection{Further topics and extensions}\label{subsec:furtherTopics}

We conclude the first part of this chapter with a brief discussion on several more advanced topics.
In particular, we discuss
\begin{enumerate}
    \item the signature of rough paths,
    \item the extent to which the signature determines the underlying path,
    \item extensions to jump processes,
    \item the role of the signature in the moment problem for random paths,
    \item extensions to higher dimensions.
\end{enumerate}

\subsubsection{Rough paths}

Our discussion of the signature has so far been restricted to paths which are piecewise continuously differentiable (or more generally of bounded variation). This restriction was needed to ensure that the iterated integrals of the path existed as Riemann--Stieltjes integrals. More generally, one can define the iterated integrals of a path using the Young integral for any path of finite $p$-variation with $1 \leq p < 2$. The Young integral goes beyond the ``classical'' definition of the integral and is already able to cover a class of paths substantially more irregular than those of bounded variation. 

A problem that arises for paths of infinite $p$-variation for all $p < 2$ (which is a situation of great interest in stochastic analysis because the sample paths of Brownian motion have almost surely finite $p$-variation if and only if $p > 2$), is that there is no well-defined notion of an iterated integral for such paths. We stress that this is not due to any technical limitation of the Young integral, but rather due to the fact that there is no unique candidate for the iterated integrals. That is to say, there is not necessarily only one unique way to define the iterated integrals.

One of the key observations of T. Lyons in his introduction of rough paths in~\cite{Lyons1998} was that if one \emph{defines} the first $\floor{p}$ iterated integrals of a path $X$ of finite $p$-variation, then there is a unique way to obtain all the other iterated integrals, and hence the signature of $X$. This notion of a path of finite $p$-variation, along with its first $\floor p$ iterated integrals (which may be defined arbitrarily provided that they satisfy Chen's identity and possess finite $p$-variation), is precisely the definition of a $p$-rough path.

A key result in the theory of rough paths, known as the \emph{universal limit theorem}, is that one can give meaning to controlled differential equations where the driver is a so-called \emph{geometric} $p$-rough path, and where the solution depends in a continuous way on the driver provided that the space of $p$-rough paths is equipped with a suitable topology (known as the $p$-variation metric).
We refer the interested reader to the St.~Flour lecture notes of Lyons--Caruana--L{\'e}vy~\cite{Lyons07} for an introduction to the theory of rough paths and to the monograph of Friz--Hairer~\cite{FrizHairer14} for a recent treatment of the topic.

\subsubsection{Path uniqueness}
\label{subsubsec:path_uniqueness}

As discussed in Section~\ref{subsubsec:Picard}, the signature of a path $X : [a,b] \mapsto \R^d$ is all that is needed to determine the endpoint of the solution to a linear (and, less trivially, non-linear) differential equation driven by $X$.
This was first shown by Chen~\cite{Chen58} in the smooth setting, and later extended by Hambly--Lyons~\cite{Hambly10} and Boedihardjo--Geng--Lyons--Yang~\cite{Boedihardjo14} to less regular paths.
The works of these authors show that the signature captures deep geometric properties of a path, which we briefly discuss here.

A natural question that one may ask is the following: is a path completely determined by its signature?
The answer is no for the reasons described above Proposition~\ref{prop:signature_unique}.
For example, one can never recover from the signature the exact speed at which the path is traversed (due to invariance under time reparametrizations), nor can one tell apart the signature of a trivial constant path and that of a path concatenated with its time-reversal.

However, a highly non-trivial fact is that this is essentially the only information lost by the signature.
For example, for a path $X$ that never crosses itself, the signature completely describes the image and direction of traversal of the path (that is, all the points that $X$ visits and the order in which it visits them), modulo its starting point.
This demonstrates the signature's ability to completely determine the geometric properties of a path that does not possess degeneracies consisting of movements going directly back onto itself.
More precisely, it is shown in~\cite{Hambly10, Boedihardjo14} that, for two paths $X,Y$, one has $S(X)= S(Y)$ if and only if $X$ and $Y$ are \emph{tree-like equivalent}.
However, it remains a challenging problem to effectively recover properties of a path from its signature: see Lyons--Xu~\cite{LyonsXu15I, LyonsXu15II} and Geng~\cite{Geng15} for progress on this question.

\subsubsection{Paths with jumps}

For many theoretical and practical applications, it is necessary to consider paths with jumps (i.e. discontinuous paths).
This is important, for example, in finance~\cite{ContTankov04}, superdiffusive dynamical systems~\cite{CFKM20,CKM23,MZ15},
and queuing theory~\cite{Whitt02}.
It turns out that, under suitable regularity assumptions on the path (satisfied, for example, by almost every sample path of semi-martingales),
there exists a meaningful way to build a signature.

The basic idea is to connect the jumps of a discontinuous path $X:[a,b]\to \R^d$ with straight lines by adding an `infinitesimal' extra time $\delta$ to the interval.
The resulting object is a continuous path $X^\delta:[a,b+\delta] \to \R^d$, for which the signature can be computed as normal and is independent of the extra time $\delta$.
For example, if $X$ is piecewise constant with finitely many jumps, the signature of $X^\delta$ is the same as that of the piecewise linear path connecting the jumps of $X$ in the respective order.
This procedure to build a continuous path $X^\delta$ from $X$ works under natural regularity assumptions, e.g. for $X$ c\`adl\`ag (right-continuous with left-limits), and does not require that $X$ has finitely many jumps.
These ideas stem from the work of Marcus~\cite{Marcus78,Marcus80} (see also~\cite{KPP95, Applebaum09,ChechkinPavlyukevich14} and the references therein)
and later developed in the context of rough paths~\cite{Williams01,FS17,Chevyrev18,CF19} (see in particular~\cite{Chevyrev18,CF19} where the requirement that the jumps of $X$ are connected with straight lines is relaxed).

We mention also another way to interpret rough paths with jumps that relies on so-called forward (or It\^o) integration~\cite{FZ17} and leads to a notion of signature, but the resulting object is most naturally interpreted in the sense of branched rough paths~\cite{Gubinelli10} as it no longer satisfies the shuffle product identity.

\subsubsection{Moment problem and random paths}
\label{subsubsec:moments}

The moment problem in probability theory is the following: given two probability measures $\PP$ and $\QQ$ on $\R$ with equal moments, i.e. $\int_\R x^k\PP(dx) = \int_\R x^k\QQ(dx)$ for all $k \geq 0$,
does it hold that $\PP=\QQ$?
As alluded to Remark~\ref{rem:Carleman}, the answer is, in general, negative, but there exist sufficient (and necessary) conditions on the growth of the moments  $\int_\R x^k\PP(dx)$ under which the answer is positive.

Recall also from Section~\ref{subsec:normalised_moments}, especially Corollary~\ref{cor:moment_prob_compact},
that the shuffle product identity and the Stone--Weierstrass theorem together imply that
a probability measure on a \textit{compact} set of paths (on which the signature map is injective) is uniquely determined by its expected signature.
This gives a positive answer to the moment problem in the simplest possible case (compact support of measures).

Given that, for probability measures on $\R$, compactness of support is a sufficient but not necessary condition to solve the moment problem,
it is natural to ask whether one can relax the compactness assumption in the solution to the moment problem for signatures.
The answer turns out to be `yes' and the following sufficient condition was found in~\cite{CL16}:
provided that the expectations $\mathbb{E}[S(X)^{i_1,\ldots,i_k}_{a,b}]$ decay with $k$ faster than any geometric sequence, then these expectations uniquely determine the law of $S(X)_{a,b}$.\footnote{It is necessary to speak of the \textit{law of the signature} $S(X)_{a,b}$
instead of the \textit{law of the path} $X$ because the signature does not, in general, uniquely determine the path, see Proposition~\ref{prop:signature_unique} and the discussion beforehand.}

The condition of \cite{CL16} is satisfied, for example, by classes of L{\'e}vy processes including Brownian motion (suitably interpreted as geometric rough paths), for which the expected signature is explicitly known~\cite{FS17},
and for classes of Gaussian and Markovian processes for which the expected signature is not explicitly known~\cite{CassOgrodnik17,CL16}.
In turn, it is known~\cite{BDMN21,LiNi22} that certain processes (e.g. Brownian motion stopped upon exiting a domain) do \textit{not} satisfy the necessary moment decay condition of~\cite{CL16}, and it remains open whether the law of their signatures is determined by its expected value.
A related open problem is to find sufficient and \textit{necessary} conditions for a solution to the moment problem for signatures, akin to the sharp Carelman's condition (see Remark~\ref{rem:Carleman}).

The method used in~\cite{CL16} to resolve the moment problem for signatures is to introduce a suitable noncommutative Fourier transform, which, as in the classical case, uniquely determines the law of any random variable without any assumptions on the moments.
Again for L\'evy processes, it is possible to derive a L\'evy--Khintchine formula that determines the Fourier transform explicitly, see~\cite[Sec.~5.3.2]{Chevyrev18}.

\subsubsection{Higher dimensions}

We conclude this section by mentioning several recent works that aim to extend the concept of signatures to functions defined over \textit{multi-dimensional space}, i.e. functions $X:\Omega \to \R^d$ where $\Omega\subset \R^n$ and possibly $n\geq 1$, instead of just paths $X:[a,b]\to\R^d$.
Due to the lack of a canonical ordering on points in $\R^n$ for $n\ge 2$, it is non-trivial to extend the notion of signature to such $X$ in a way that preserves properties such as the shuffle product and Chen's identity from Section~\ref{subsec:important_properties}.
Nonetheless, such an extension clearly has value since many types of data, e.g., images, are spatial (vs. temporal) and lack a time-ordered structure.

Zhang--Lin--Tindel~\cite{ZLT22}, inspired by~\cite{Tindel_Chouk_15}, recently defined 2D signatures and applied this concept to image and texture classification.
Their approach was based on rectangular increments and was restricted to level-1 and level-2 (generalizations of) signatures.
Giusti--Lee--Nanda--Oberhauser~\cite{GLNO22} provided an (essentially topological) extension of signatures based on cubical mapping spaces
that applies to arbitrary dimensions and levels.
The authors therein show generalizations of the shuffle product and Chen's identity
and are able to partially address the uniqueness question discussed in Section~\ref{subsubsec:path_uniqueness}.
Recently, Lee--Oberhauser~\cite{LO23} and then \cite{CDEFT24,Lee24} provided another generalization of the signature to surfaces which stems from higher gauge theory.
See also~\cite{CGW21} that provides a candidate for a multi-dimensional signature based on the notion of a \textit{model} from the theory of regularity structures~\cite{Hairer14}, which is used in the analysis of singular stochastic partial differential equations.
This generalization, however, relies on several (a priori ad hoc) choices, including an integration operator that replaces the (somewhat canonical) Heaviside function in the one-dimensional case; it remains open if important properties of the signature extend to this construction.
Finally, we mention \cite{2022arXiv221014247D} that introduces two-parameter sums signatures with applications to a multi-dimensional version of dynamic time warping.

\section{Practical Applications}
\label{sec:prac_app}

One of the practical applications of the signature transformation lies in the field of machine learning algorithms. As has already been discussed, the signature summarizes important information about a path. When the path is composed of a sequential data stream $\{X_i\}$, the terms of the signature $S(X)^{ijk...}$ are good candidates for characteristic {\it features} of the data stream. The shuffle product property allows us to represent a non-linear function of the signature as a linear combination of iterated integrals. This is similar to basis function expansion and naturally applies to computations of regression models.
In the next sections we will demonstrate numerical computations of signatures of paths from various data streams and applications of signatures to machine learning problems, namely classification.

\subsection{Elementary ingredients and their properties}

In this section, we introduce the essential ingredients of the signature method and demonstrate how to perform basic computations.

\subsubsection{Paths from discrete data}
\label{subsubsec:paths_from_discrete_data}

Our first task is to transform discrete sequential (i.e., time series) data into continuous paths, as signatures are typically defined for continuous paths rather than a collection of discrete data points. We start with basic computations of the signature applied to data streams. Let $X$ be a data stream defined as a set of $N$ points observed at times $t_0 < t_1 < t_2 < ... < t_N$ in $d$-dimensions ($X\in\mathbb{R}^d$):
\begin{equation}\label{eq:data_stream}
X=\left\{\Bigl(X^{1}_{t_i},X^2_{t_i},X^3_{t_i},...,X^d_{t_i}\Bigr)\right\}_{i=0}^N.
\end{equation}
One can think of $\Bigl(X^{1}_{t_i},X^2_{t_i},X^3_{t_i},...,X^d_{t_i}\Bigr)$ as a coordinate of a data point recorded at time $i$ in $\mathbb{R}^d$.
For example, consider two one-dimensional sequences of length four:
\begin{align}
\{X^1_{t_i}\}_{i=0}^{N=3} &= \{1,3,8,6\},\label{eq:twoOneDimSeq_2}\\
\{X^2_{t_i}\}_{i=0}^{N=3} &= \{1,4,2,5\}.\label{eq:twoOneDimSeq_3}
\end{align}
We are interested in transforming this discrete series into a continuous path. Among various ways to find this transformation, we focus on two main approaches: 
\begin{enumerate}[label=(\alph*)]
\item \textbf{Piecewise linear interpolation.}

This interpolation method is given by a set of linear segments (local linear approximation) connecting a set of data points. The resulting continuous path is a piecewise combination of linear functions. Formally, for each pair of points $X_i$ and $X_{i+1}$ on the interval $t\in [t_i, t_{i+1}]$, the path is defined by
\begin{equation}\label{eq:linear_interpolation}
    \hat{X_t} = X_{t_i} + \frac{t-t_i}{t_{i+1} - t_i}\left(X_{t_{i+1}} - X_{t_i}\right).
\end{equation}
By concatenating linearly interpolated paths, one arrives at the continuous path $\hat{X_t}$ defined for any $t\in[t_0, t_N]$. 
%
%

\item \textbf{Rectilinear interpolation} (i.e. {\it axis path}).

Another way to connect $X_{t_i}$ and $X_{t_{i+1}}$ data points is by updating consecutively each coordinate of $X_{t_i}$ in a piecewise linear fashion along the $d$ axes until we reach $X_{t_{i+1}}$. In this way, on the interval  $t\in[t_i, t_{i+1}]$ with $i=0,1, ..., N-1$, one augments a pair of two consecutive points by $d-1$ \textit{auxiliary} data points of which we take the piecewise linear interpolation as in~\eqref{eq:linear_interpolation}.
For example, if $d=3$, the rectilinear path over $[t_i,t_{i+1}]$ is the piecewise linear interpolation of the four data points
\begin{equation*}
\{(X^1_{t_i}, X^2_{t_i}, X^3_{t_i}), (X^1_{t_{i+1}}, X^2_{t_i},X^3_{t_i}), (X^1_{t_{i+1}}, X^2_{t_{i+1}},X^3_{t_i}), (X^1_{t_{i+1}}, X^2_{t_{i+1}},X^3_{t_{i+1}})\}.
\end{equation*}
Remark that, for the computation of the signature, it is not important how the rectilinear path is precisely parametrized over $[t_i,t_{i+1}]$ due to the parametrization invariance of the signature (Section~\ref{subsubsec:repar_invar}).
\end{enumerate}

\begin{figure}[ht]
    \centering
    \begin{subfigure}[b]{0.48\textwidth}
        \centering
        \includegraphics[width=0.9\textwidth]{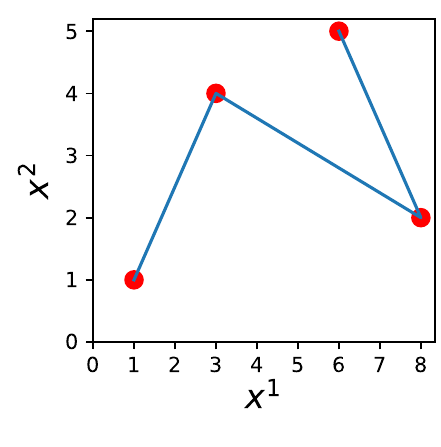}
        \caption{}
        \label{fig:piecewise_interp}
    \end{subfigure}
    \hfill
    \begin{subfigure}[b]{0.48\textwidth}
        \centering
        \includegraphics[width=0.9\textwidth]{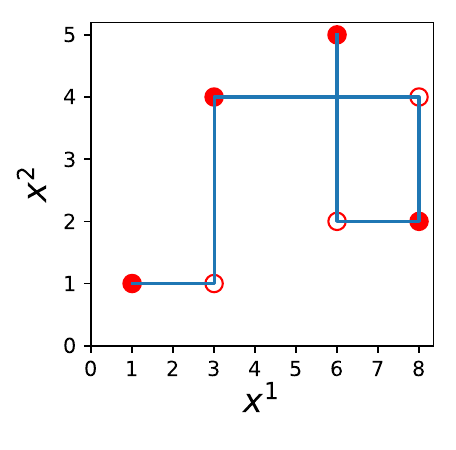}
        \caption{}
        \label{fig:rectilinear_interp}
    \end{subfigure}
    \caption{Examples of a piecewise linear (a) and rectilinear (b) interpolation schemes applied to a two-dimensional stream $\{(X^1, X^2)\}$ from \eqref{eq:twoOneDimSeq_2} and \eqref{eq:twoOneDimSeq_3}.}
    \label{fig:path_interps}
\end{figure}

The piecewise linear and rectilinear interpolation schemes using the data streams in \eqref{eq:twoOneDimSeq_2} and \eqref{eq:twoOneDimSeq_3} are presented in Fig.~\ref{fig:piecewise_interp} and Fig.~\ref{fig:rectilinear_interp} respectively. Note that the auxiliary points denoted by empty red circles were added to construct the rectilinear paths. 


\subsubsection{Auxiliary path transformations}

Although data points can be transformed directly into paths as in  Section~\ref{subsubsec:paths_from_discrete_data}, in certain cases, auxiliary data transformations may improve statistical learning with signatures for downstream tasks.
Such transformations may increase or decrease the dimension $d$ of the data stream and lead to more effective representations of sequential data that enhance the model's ability to learn, generalize, and make accurate predictions. In this section, we present several frequently used transformations.

\subsubsection*{Cumulative sum}
    
The cumulative sum, also known as the running total, is the sequence of partial sums of a given sequence of numbers. In other words, it is the sequence of sums of the elements of the original sequence, where each sum is calculated by adding the previous sum to the current element of the sequence.
More precisely, the cumulative sum is defined by
\begin{equation}\label{eq:cumSumDef}
\mathrm{CS}(\{X_{t_i}\}_{i=0}^{N}) = \{X_{t_0},X_{t_0}+X_{t_1},\dots,S_k,\dots,S_N\}\;;\;\;\;\;S_k = \sum_{i=0}^{k}X_{t_i}.
\end{equation}
Using again the previous example \eqref{eq:twoOneDimSeq_2} and \eqref{eq:twoOneDimSeq_3}, the new data streams become
\begin{align*}
\{\tilde{X^1}\} = \mathrm{CS}(X^1) &= \{1, 4, 12, 18\},\\
\{\tilde{X^2}\} = \mathrm{CS}(X^2) &= \{1, 5, 7, 12\}.
\end{align*}
\subsubsection*{Base-point augmentation}

As discussed in Section \ref{sec:First}, the signature transformation is translation invariant, meaning that paths at different locations in $\mathbb{R}^d$ will produce the same signature. In some cases, it may be useful to remove this translation invariance. For example, to capture the absolute values of data streams rather than their increments. To do this, an anchor data point, typically the origin (0), can be added at the beginning of the data stream. This removes the translation invariance of the signature transformation. Using examples \eqref{eq:twoOneDimSeq_2} and \eqref{eq:twoOneDimSeq_3}, the augmented data streams with $N+1$ data points become
\begin{equation}\label{eq:bp_augmentation}
\begin{aligned}
\mathrm{BP}(\{X^1_{t_i}\}_{i=0}^{N=3}) = \{0, \{X^1_{t_i}\}_{i=0}^{N=3}\} &= \{0,1,3,8,6\},\\
\mathrm{BP}(\{X^2_{t_i}\}_{i=0}^{N=3}) = \{0, \{X^2_{t_i}\}_{i=0}^{N=3}\} &= \{0,1,4,2,5\}.
\end{aligned}
\end{equation}

\subsubsection*{Time augmentation}

This type of sequential data transformation is defined by adding a univariate component with monotonically increasing or decreasing values that can be interpreted as \textit{auxiliary time}. For example, consider a new (unevenly sampled) stream $\{t\}$
\begin{equation*}
    \{t_i\}_{i=0}^{N=3} = \{0,1,3,6\},
\end{equation*}
and augmenting $\{X^1\}_{i=0}^{N=3}$, $\{X^2\}_{i=0}^{N=3}$ and $\{(X^1, X^2)\}_{i=0}^{N=3}$ from \eqref{eq:twoOneDimSeq_2} and \eqref{eq:twoOneDimSeq_3} leads to
\begin{align*}
    \mathrm{TA}(\{X^1_{t_i}\}_{i=0}^{N=3}) &= \{(t_i, X^1_{t_i})\}_{i=0}^{N=3} = \{(0,1), (1,3), (3,8), (6,6)\},\\
    \mathrm{TA}(\{X^2_{t_i}\}_{i=0}^{N=3}) &= \{(t_i, X^2_{t_i})\}_{i=0}^{N=3} = \{(0,1), (1,4), (3,2), (6,5)\},\\
    \mathrm{TA}(\{(X^1_{t_i}, X^2_{t_i}\}_{i=0}^{N=3}) &= \{(t_i, X^1_{t_i}, X^2_{t_i})\}_{i=0}^{N=3} = \{(0,1,1), (1,3,4), (3,8,2), (6,6,5)\}.
\end{align*}
The signature of a stream integrated against a monotonic auxiliary stream (`time') can capture stream-specific effects and can be related to common statistics (e.g., mean). Intuitively, the integral of a stream against the auxiliary time component will describe the area under the path, which in some applications may capture important properties of the data process via the path and serve as an informative feature for downstream analytical tasks.

In some cases, knowing the difference between the timestamps may be informative, particularly for unevenly distributed data. To illustrate this scenario, consider streams $\{X^1\}$ and $\{X^2\}$ and their respective timestamps $\{t_i\}$ as shown in Table~\ref{tab:table1_time_diff_a}.

\begin{table}[ht]
\begin{subtable}[t]{0.48\textwidth}
    \centering
    \normalsize
    \begin{tabular}{c|c|c}
        $t_i$       & $X^2$ & $X^2$ \\\hline\hline
        01-January  &  1    &   1   \\
        02-January  &  3    &   4   \\
        04-January  &  8    &   2   \\
        07-January  &  6    &   5   \\
        08-January  &  9    &   3   \\\hline
    \end{tabular}
\caption{}
\label{tab:table1_time_diff_a}
\end{subtable}
\begin{subtable}[t]{0.48\textwidth}
    \centering
    \normalsize
    \begin{tabular}{c|c|c|c|c}
        $t_i$       &   $\hat{t}_i$ & $\Delta t_i$ & $X^2$ & $X^2$ \\\hline\hline
        01-January  &     0         &      -       &   1    &   1   \\
        02-January  &     1         &      1       &   3    &   4   \\
        04-January  &     3         &      2       &   8    &   2   \\
        07-January  &     6         &      3       &   6    &   5   \\
        08-January  &     7         &      1       &   9    &   3   \\\hline
    \end{tabular}
\caption{}
\label{tab:table1_time_diff_b}
\end{subtable}
\caption{Example streams.}
\end{table}
In addition to the original timestamp stream $\{t_i\}$, we can introduce two auxiliary time-related streams:
\begin{eqnarray*}
    \hat{t}_i &=& \{t_i - t_0\}_{i=0}^{N=4}  = \{0, 1, 3, 6, 7\},\\
    \Delta t_i &=&  \{t_{i} - t_{i-1}\}_{i=1}^{N=4} = \{1, 2, 3, 1\}.
\end{eqnarray*}
Putting all streams together and aligning them in time, we get a 5-dimensional stream as shown in Table~\ref{tab:table1_time_diff_b}. Intuitively, $\{\hat{t}_i\}$ corresponds to time since the beginning and $\{\Delta t_i\}$ to time differences between two consecutive timestamps. Note that, since there is no time difference for the first data point, some heuristic can be used to fill in the missing value, such as arbitrarily setting it to zero.

\subsubsection*{Lead-lag transformation}\label{seq:lead_lag_section}

A lead-lag transformation is a mathematical operation that shifts the phase of a time series by a specified amount. In a lead-lag transformation, a time series can be either ``led'' or ``lagged'' by a certain number of time steps. Leading a signal means that it is shifted forward in time, while lagging a signal means that it is shifted backward in time. Lead-lag transformations can be used to analyze the relationships between different signals. For example, the lead-lag transformation can map a one-dimensional path into a two-dimensional path. Certain statistics, such as the quadratic variation of a stochastic process, can be represented in terms of the signature of the lead-lag transformed process (see Section~\ref{subsec:further_relationship}).

Let $X$ a $d$-dimensional data stream with $N$ data points as in~\eqref{eq:data_stream}.
Following the definition \cite{gyurko2014extracting}, its lead and lag transformed streams $\{X^{\mathrm{Lead}}_j\}_{j=0}^{2N}$ and $\{X^{\mathrm{Lag}}_j\}_{j=0}^{2N}$ with $2N+1$ data points are defined as
\begin{equation}\label{eq:lead_stream}
X^{\mathrm{Lead}}_{t_j} \mapsto 
\begin{cases}
             X_{t_i} & \text{if} \;\; j=2i,\\
             X_{t_i} & \text{if} \;\; j=2i-1,\\
\end{cases}
\end{equation}
and
\begin{equation}\label{eq:lag_stream}
X^{\mathrm{Lag}}_{t_j} \mapsto 
\begin{cases}
             X_{t_i} & \text{if} \;\; j=2i,\\
             X_{t_i} & \text{if} \;\; j=2i+1.\\
\end{cases}
\end{equation}
It is clear from \eqref{eq:lead_stream} and \eqref{eq:lag_stream} that for each value of $i$ corresponds to three values of $j$:
\begin{eqnarray*}
    j &=& 2i-1,\\
    j &=& 2i,\\
    j &=& 2i+1.
\end{eqnarray*}
For example, given a one-dimensional sequence with four values
\begin{equation*}
    \{X_{t_i}\}_{i=0}^{N=3} = \{X_{t_0}, X_{t_1}, X_{t_2}, X_{t_3}\},    
\end{equation*}
by substituting $i\in[0,1,2,3]$ in \eqref{eq:lead_stream} the components of the lead stream of $\{X_{t_i}\}$ are
\begin{equation*}
\begin{aligned}
    & & X^{\mathrm{Lead}}_{t_0} = X_{t_0},\;\; X^{\mathrm{Lead}}_{t_1} = X^{\mathrm{Lead}}_{t_2} = X_{t_1}, \\
    X^{\mathrm{Lead}}_{t_3}  
    & = & X^{\mathrm{Lead}}_{t_4} = X_{t_2},\;\; X^{\mathrm{Lead}}_{t_5} = X^{\mathrm{Lead}}_{t_6} = X_{t_3}, 
\end{aligned}
\end{equation*}
and correspondingly of the lag \eqref{eq:lag_stream} stream of $\{X_{t_i}\}$ are
\begin{eqnarray*}
    X^{\mathrm{Lag}}_{t_0} & = & X^{\mathrm{Lag}}_{t_1} = X_{t_0},\;\; X^{\mathrm{Lag}}_{t_2} = X^{\mathrm{Lag}}_{t_3} = X_{t_1}, \\
    X^{\mathrm{Lag}}_{t_4}  & = & X^{\mathrm{Lag}}_{t_5} = X_{t_2},\;\; X^{\mathrm{Lag}}_{t_6} = X^{\mathrm{Lag}}_{t_7} = X_{t_3}, 
\end{eqnarray*}
or more compactly
\begin{eqnarray*}
    \mathrm{Lead}: \{X_{t_i}\}_{i=0}^{3}\mapsto\{X^{\mathrm{Lead}}_{t_j}\}_{j=0}^{6} & = & \{X_{t_0}, X_{t_1}, X_{t_1}, X_{t_2}, X_{t_2}, X_{t_3}, X_{t_3}\}, \\
    \mathrm{Lag}: \{X_{t_i}\}_{i=0}^{3}\mapsto\{X^{\mathrm{Lead}}_{t_j}\}_{j=0}^{6} & = & \{X_{t_0}, X_{t_0}, X_{t_1}, X_{t_1}, X_{t_2}, X_{t_2}, X_{t_3}\}. 
\end{eqnarray*}
One can immediately see that lead and lag transforms are derived from the original data stream by repeating and shifting its data points and deleting the first and last data points. They are repeated and time-shifted versions of each other.

\begin{figure}[ht]
    \centering
    \includegraphics[width=\linewidth]{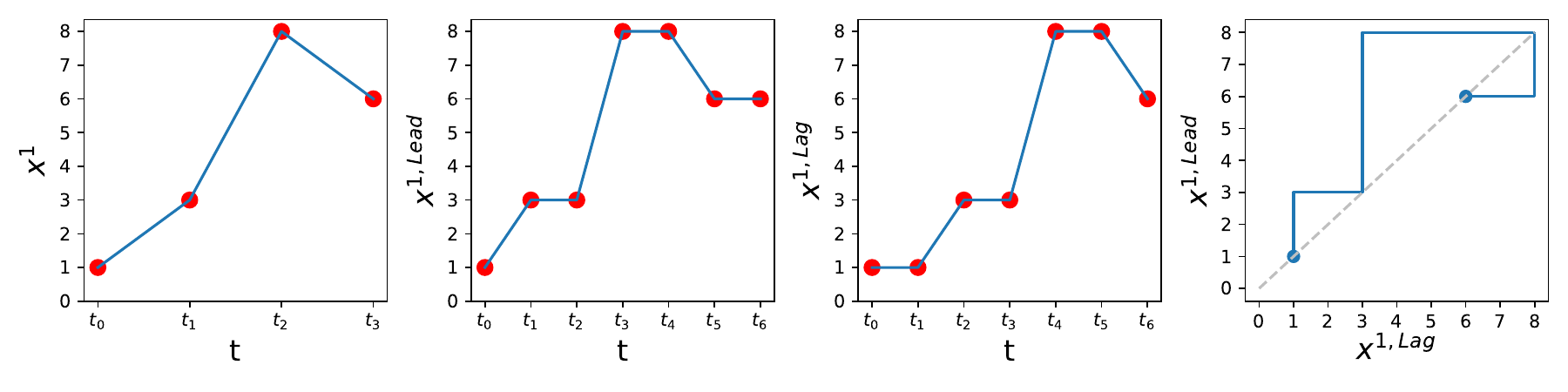}
    \caption{Lead and lag paths of $X^1$.}
    \label{fig:lead_lag_of_x1}
\end{figure}

\begin{figure}[ht]
    \centering
    \includegraphics[width=\linewidth]{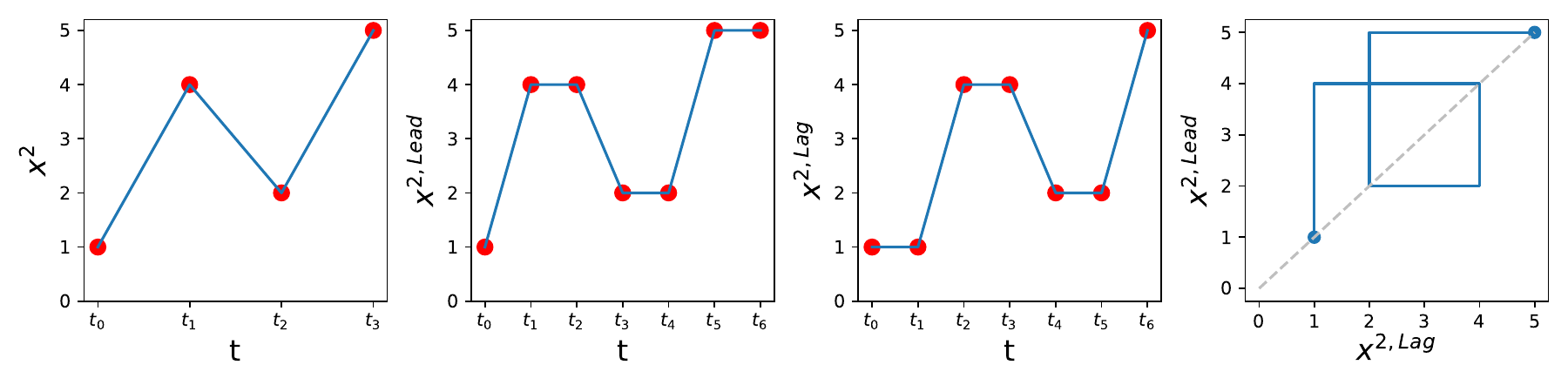}
    \caption{Lead and lag paths of $X^2$.}
    \label{fig:lead_lag_of_x2}
\end{figure}

Considering the sequence $X^1$ from~\eqref{eq:twoOneDimSeq_2}, its lead-lag (LL) mapping is given by
\begin{equation}\label{eq:leadLagAlgo_X1}
\mathrm{LL}\;:\;X^1 = \left\{1, 3, 8, 6\right\} \mapsto 
\begin{cases}
             X^{1,\mathrm{Lead}}&= \left\{1, 3, 3, 8, 8, 6, 6\right\},\\
             X^{1,\mathrm{Lag}} &= \left\{1, 1, 3, 3, 8, 8, 6\right\},\\
\end{cases}
\end{equation}
and illustrated in Fig.~\ref{fig:lead_lag_of_x1}.
The corresponding lead-lag transformation of $X^2$ from~\eqref{eq:twoOneDimSeq_3} is given by
\begin{equation*}
\mathrm{LL}\;:\;X^2 = \left\{1, 4, 2, 6\right\} \mapsto 
\begin{cases}
             X^{2,\mathrm{Lead}}&= \left\{1, 4, 4, 2, 2, 6, 6\right\},\\
             X^{2,\mathrm{Lag}} &= \left\{1, 1, 4, 4, 2, 2, 6\right\},
\end{cases}
\end{equation*}
and illustrated in Fig.~\ref{fig:lead_lag_of_x2}.

\subsubsection*{Multivariate time series as a path in \texorpdfstring{$\mathbb{R}^d$}{Rd}}

\begin{figure}[ht]
    \centering
    \includegraphics[width=\linewidth]{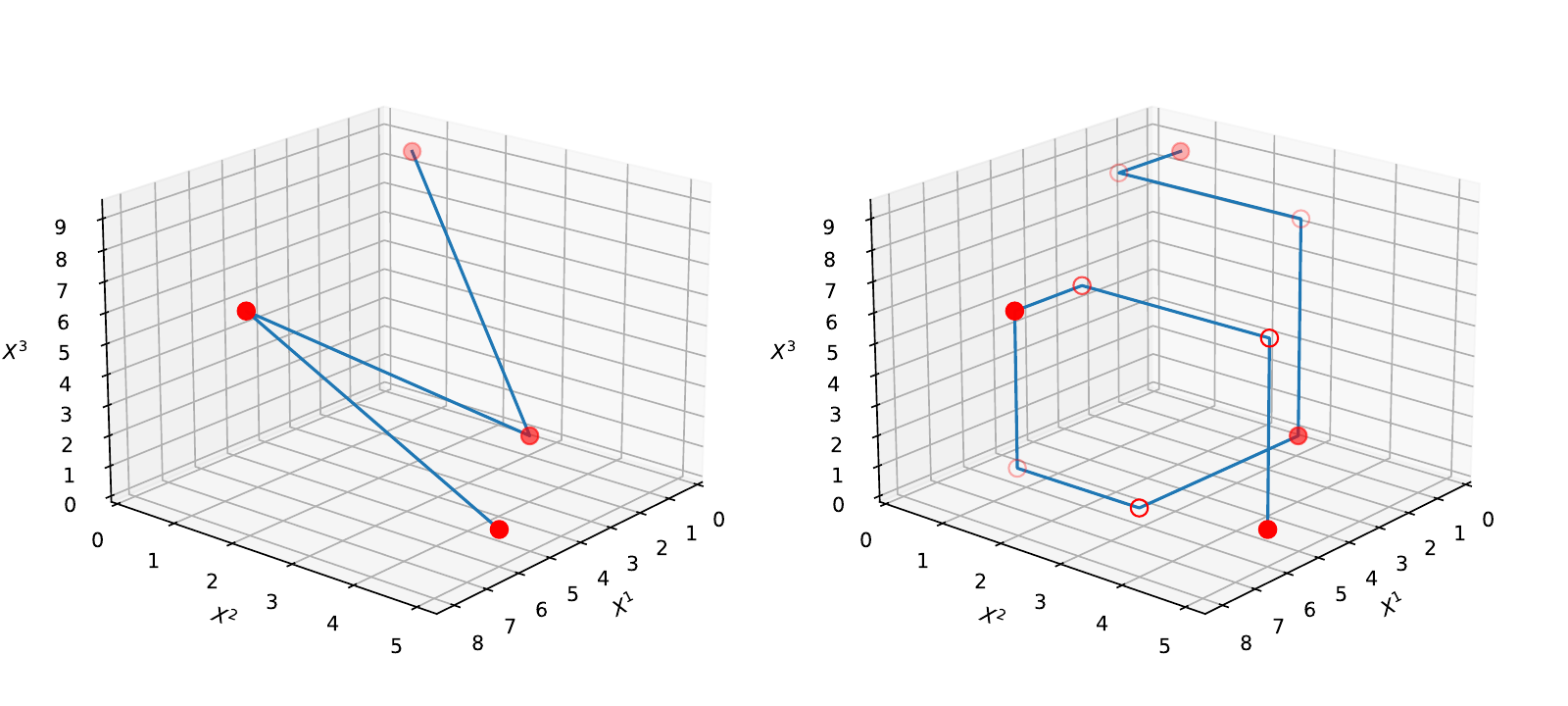}
    \caption{Example of embedding three data streams in $\mathbb{R}$ into a single path in $\mathbb{R}^3$ using piecewise linear (left) and rectilinear (right) interpolations respectively.}
    \label{fig:x1x2x3}
\end{figure}

The advantage of using the path representation of sequential data is that a $d$-dimensional process can be embedded as a path in $\mathbb{R}^d$.
For example, the two univariate streams $\{X^1\}$ from~\eqref{eq:twoOneDimSeq_2} and $\{X^2\}$ from~\eqref{eq:twoOneDimSeq_3} can be embedded into a path in $\mathbb{R}^2$ as
\begin{eqnarray*}
    \{X_{t_i}\}_{i=0}^{N=3} &=& \{\bigl(X^1_{t_i}, X^2_{t_i} \bigr)\}_{i=0}^{N=3}\\
    &=& \{(1, 1), (3, 4), (8, 2), (6, 5)\},
\end{eqnarray*}
the piecewise linear and rectilinear interpolations of which are displayed in Fig.~\ref{fig:path_interps}.
This can be easily extended to $\mathbb{R}^3$ if we add another univariate stream of the same length:
\begin{equation*}
    \{X^3_{t_i}\}_{i=0}^{N=3} = \{9,2,7,1\},\\
\end{equation*}
resulting in
\begin{eqnarray*}
    \{X_{t_i}\}_{i=0}^{N=3} &=& \{\bigl(X^1_{t_i}, X^2_{t_i}, X^3_{t_i} \bigr)\}_{i=0}^{N=3}\\
    &=& \{(1, 1, 9), (3, 4, 2), (8, 2, 7), (6, 5, 1)\}.
\end{eqnarray*}
The resulting paths in 3-dimensions using piecewise linear and rectilinear interpolations are shown in Fig.~\ref{fig:x1x2x3}.

Another example of a multidimensional embedding involves the lead-lag transform. Suppose we are interested in capturing the quadratic variation in one of the multivariate streams. As we will see in Section~\ref{subsec:further_relationship}, we can compute the quadratic variation using the lead-lag transformation. Since the lead-lag transformation increases the number of points from $N$ to $2N+1$ (cf. Section \ref{seq:lead_lag_section}), all other stream lengths should be adjusted accordingly. It is worth noticing that, due to reparametrization invariance, the signatures of the lead and lag transformations of streams are equivalent:
\begin{equation*}
    S\Bigl(\bigl\{\bigl(X^1, X^2\bigr)\bigr\}\Bigr) \equiv S\Bigl(\{\bigl(X^{1, \mathrm{Lead}}, X^{2, \mathrm{Lead}}\bigr)\}\Bigr) \equiv S\Bigl(\{\bigl(X^{1, \mathrm{Lag}}, X^{2, \mathrm{Lag}}\bigr)\}\Bigr).
\end{equation*}

\subsubsection*{Further path transformations}

In this subsection, we showed several examples of different path augmentations, but there exist many others that serve various purposes. For example, invisibility-reset augmentation that adds information on the initial position of the path,
coordinate projections that allow to compute the signature of a subset of coordinates individually, data rescaling,
and random projections that project a high-dimensional path onto a lower-dimensional subspace using a random matrix. For further details, see \cite{morrill2020generalised}.


\subsubsection{Signed area}
\label{sec:signed_area}
Before we continue, it is important to emphasize the significance of the direction in which we travel along a path and the sign of the area which it encloses. Intuitively, the signed area enclosed by a path is a measure of the 2-dimensional region enclosed by the path. It is calculated by taking the absolute value of the area and then assigning a positive or negative sign to it depending on the orientation of the path. A positively oriented path (or \textit{a curve}) is a simple closed path in the plane such that when following the path, the interior of the path is always on the left side. For a negatively oriented path, the interior is on the right side instead of the left side. In other words, the orientation of a path is determined by the direction in which the path appears to turn as it is traced. If the path turns clockwise, it is negatively oriented. If the path turns counterclockwise, it is positively oriented.

\begin{figure}[ht]
     \centering
     \begin{subfigure}[b]{0.32\textwidth}
         \centering
         \includegraphics[width=\textwidth]{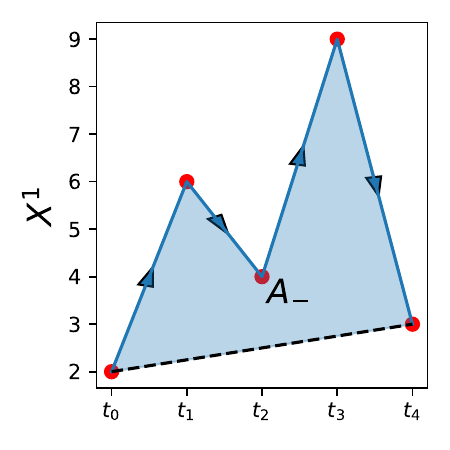}
         \caption{Negative area $A_{-}$.}
         \label{fig:negative_area}
     \end{subfigure}
     \begin{subfigure}[b]{0.32\textwidth}
         \centering
         \includegraphics[width=\textwidth]{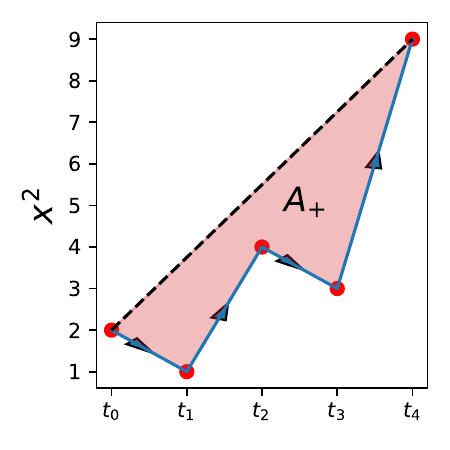}
         \caption{Positive area $A_{+}$.}
         \label{fig:positive_area}
     \end{subfigure}
     \begin{subfigure}[b]{0.32\textwidth}
         \centering
         \includegraphics[width=\textwidth]{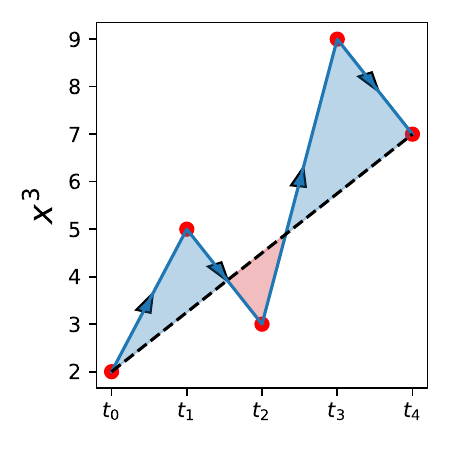}
         \caption{Positive and negative areas.}
         \label{fig:positive_and_neagative_area}
     \end{subfigure}    
     \caption{Examples of signed areas defined by the orientation of a path enclosing them.}
     \label{fig:signed_area_examples}
\end{figure}

For example, if a path is traced clockwise, the signed area of the enclosed region would be negative (Fig.~\ref{fig:negative_area}). If the path is traced counterclockwise, the signed area would be positive (Fig.~\ref{fig:positive_area}). A combination of positively and negatively oriented paths and their respective areas is shown in Fig.~\ref{fig:positive_and_neagative_area}. The sign of an area is also related to the sign of the winding number of a path~\cite{BoedihardjoNi14}. 

\subsubsection*{The relative movement as a signed area}

\begin{figure}[ht]
\centering
\includegraphics[width=0.8\linewidth]{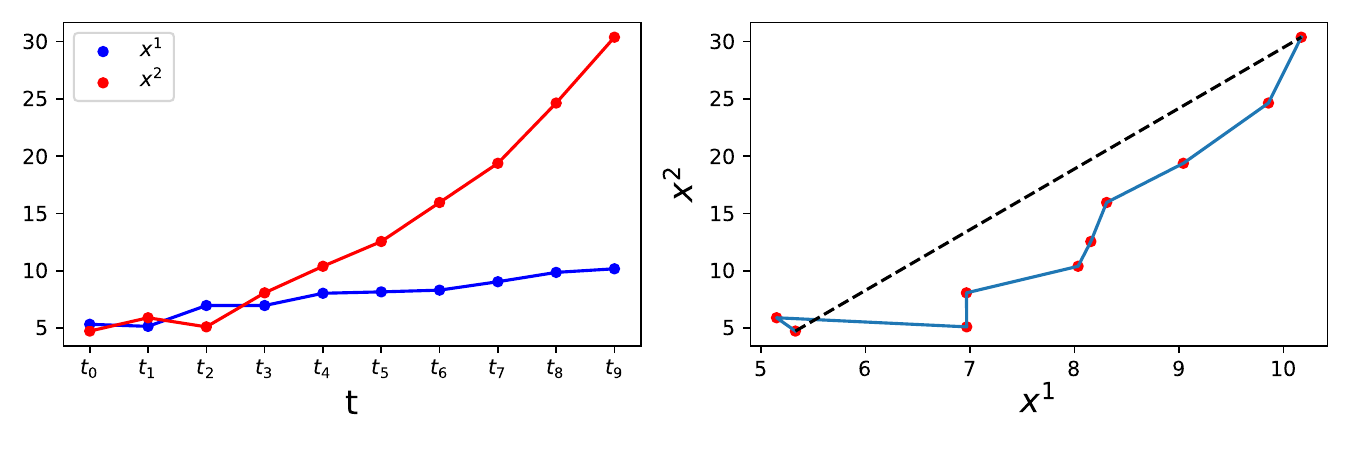}
\caption{Example of a positive L{\'e}vy area that corresponds to two paths, $\{X^1\}$ and $\{X^2\}$, with relative movement in the same directions.}
\label{fig:positive_area2}
\end{figure}

\begin{figure}[ht]
\centering
\includegraphics[width=0.8\linewidth]{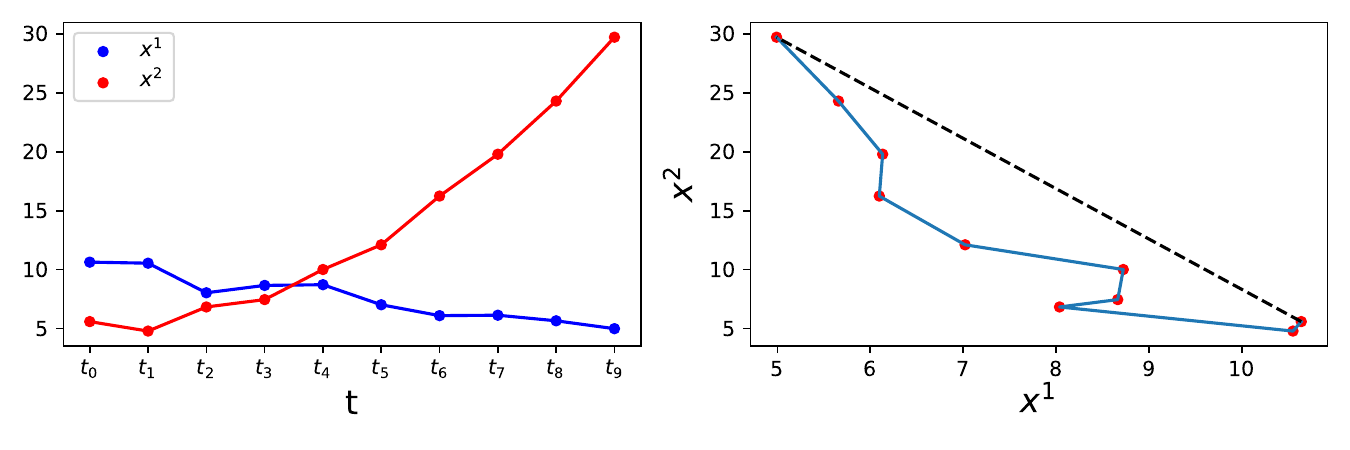}
\caption{Example of a negative L{\'e}vy area that corresponds to two paths, $\{X^1\}$ and $\{X^2\}$, with relative movement in opposite directions.}
\label{fig:negative_area2}
\end{figure}

Consider a path in $\mathbb{R}^2$ given by $X = \{X^1,X^2\}$. If an increase (resp. decrease) in the component $X^1$ is followed by an increase (resp. decrease) in the component $X^2$, then the area $A$ given by \eqref{eq:LevyArea} is positive (Fig.~\ref{fig:positive_area2}). If the relative moves of the components $X^1$ and $X^2$ are in the opposite direction, then the area $A$ is negative (Fig.~\ref{fig:negative_area2}). 

\noindent

\subsubsection{Computation and geometry of the signature}
\label{sec:signature_of_a_path}

\begin{figure}[ht]
     \centering
     \begin{subfigure}[b]{0.48\textwidth}
         \centering
         \includegraphics[width=\textwidth]{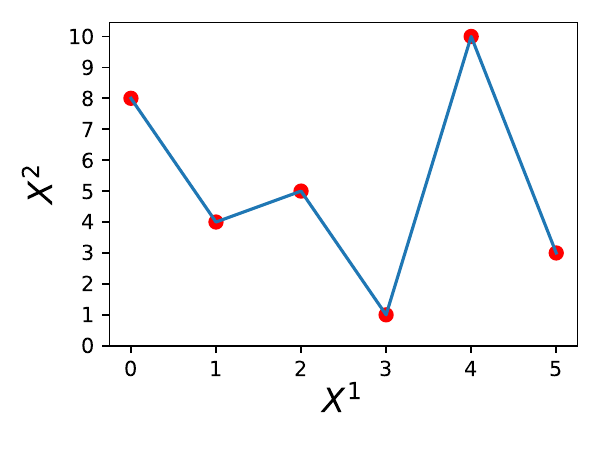}
         \caption{A piecewise linear path $\{(X^1, X^2)\} \in\mathbb{R}^2$ constructed from $\{(X^1)\}$ and $\{(X^2)\}$ using data points as \eqref{eq:sig_x1x2}.}
         \label{fig:sig_raw}
     \end{subfigure}
     \hfill
     \begin{subfigure}[b]{0.48\textwidth}
         \centering
         \includegraphics[width=\textwidth]{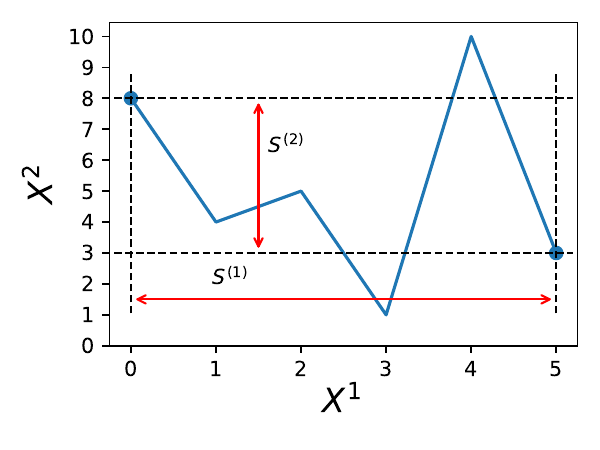}
         \caption{Geometric meaning of the first two terms $S^{1}$ and $S^{2}$ of the signature of the path $\{(X^1, X^2)\}$.}
         \label{fig:sig_s1s2}
     \end{subfigure}
     \caption{Examples of two univariate data streams \eqref{eq:sig_x1x2} embedded in $\mathbb{R}^2$ and the geometric meaning of their signature terms $S^{1}$ and $S^{2}$.}
\end{figure}

In Section~\ref{sec:First}, we outlined the mathematical foundations of the signature of a path and defined the signature terms as multiple iterated integrals. In this subsection, we provide another example of a computation and the geometric meaning of the signature. Consider the two univariate streams $\{(X^1, X^2)\}$
\begin{eqnarray}\label{eq:sig_x1x2}
    \{(X^1)\} &=& \{0, 1, 2, 3, 4, 5\},\\
    \{(X^2)\} &=& \{8, 4, 5, 1, 10, 3\},\nonumber
\end{eqnarray}
embedded into a piecewise linear path in $\mathbb{R}^2$ (Fig.~\ref{fig:sig_raw}). Using the definition of the signature \eqref{eq:definSig}, we compute its first terms truncated at level $L=2$ as
\begin{center}
    \begin{tabular}{lrrrrrrr}
        $S(X)$ = & (1, & $S^{1}$, & $S^{2}$, & $S^{1,1}$, & $S^{1,2}$, & $S^{2,1}$, & $S^{2,2}$)  \\
                \hfill = & $(1$, & $5$,       & $-5$,      & $12.5$,      & $-10.5$,     & $-14.5$,     & $12.5$)  .
    \end{tabular}
\end{center}

One can easily confirm (cf.~\eqref{singInteg}) that the first order signature terms correspond to the total increments along each data stream:
\begin{eqnarray*}
        S^{1} &=& \int dX^1_t = X^1_{t_5} - X^1_{t_0} = 5-0 = 5,\\
        S^{2} &=& \int dX^2_t = X^2_{t_5} - X^2_{t_0} = 3-5 = -5,
\end{eqnarray*}
which are depicted in Fig.~\ref{fig:sig_s1s2}.
The second order signature terms are computed using double iterated integrals (cf. \eqref{doublInteg}):
\begin{equation}\label{eq:four_sig_numerical}
\begin{aligned}
       S^{1,1} &= \int S^{1}\, dX^1_t = \int \int dX^1_t \, dX^1_t = \frac{1}{2!}\left(\int dX^1_t\right)^2 = \frac{1}{2!}(S^{1})^2 = 12.5,\\
    S^{1,2} &= \int S^{1}\, dX^2_t = \int \int dX^1_t \, dX^2_t = -10.5,\\
    S^{2,1} &= \int S^{2}\, dX^1_t = \int \int dX^2_t \, dX^1_t = -14.5,\\
    S^{2,2} &= \int S^{2}\, dX^2_t = \int \int dX^2_t \, dX^2_t = \frac{1}{2!}\left(\int dX^2_t\right)^2 = \frac{1}{2!}(S^{2})^2 = 12.5. 
\end{aligned}
\end{equation}
  
\begin{figure}[ht]
     \centering
     \begin{subfigure}[b]{0.48\textwidth}
         \centering
         \includegraphics[width=\textwidth]{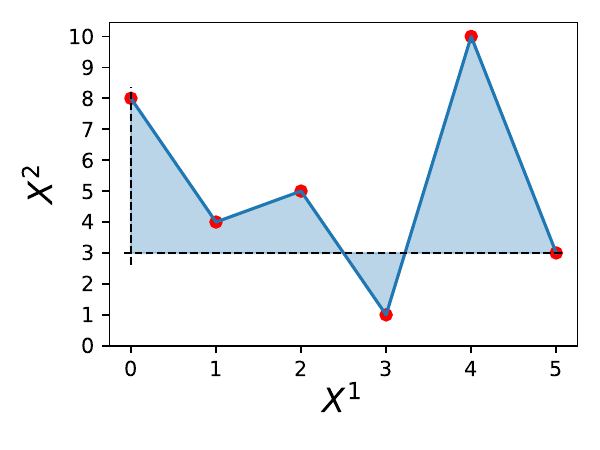}
         \caption{Area represented by the term $S^{1,2}$.}
     \end{subfigure}
     \hfill
     \begin{subfigure}[b]{0.48\textwidth}
         \centering
         \includegraphics[width=\textwidth]{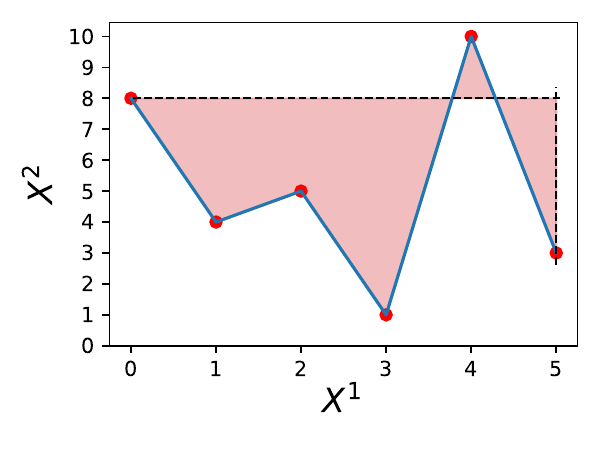}
         \caption{Area represented by the term $S^{2,1}$.}
     \end{subfigure}
     \caption{The geometric meaning of the terms $S^{1,2}$ and $S^{2,1}$ of $\{(X^1, X^2)\}$ defined in \eqref{eq:sig_x1x2}. The left panel (a) represents the area enclosed by the path and two perpendicular dashed lines passing through the endpoints of the path, while the right panel (b) shows another possibility for the area to be enclosed by the path and two perpendicular dashed lines passing through the endpoints.}
     \label{fig:signature_terms_s12_s21}
\end{figure}

The second order signature terms $S^{1,2}$ and $S^{2,1}$ and their geometric interpretation as areas are presented in Fig.~\ref{fig:signature_terms_s12_s21}.

\begin{figure}[ht]
     \centering
     \begin{subfigure}[b]{0.48\textwidth}
         \centering
         \includegraphics[width=\textwidth]{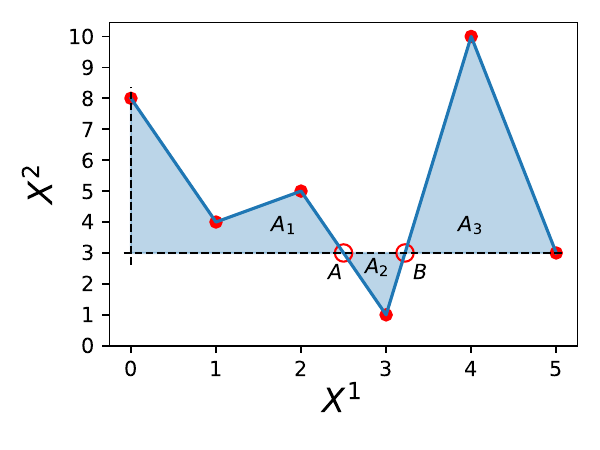}
         \caption{The total area $S^{1,2} = A_1 +  A_2 + A_3$.}
     \end{subfigure}
     \hfill
     \begin{subfigure}[b]{0.48\textwidth}
         \centering
         \includegraphics[width=\textwidth]{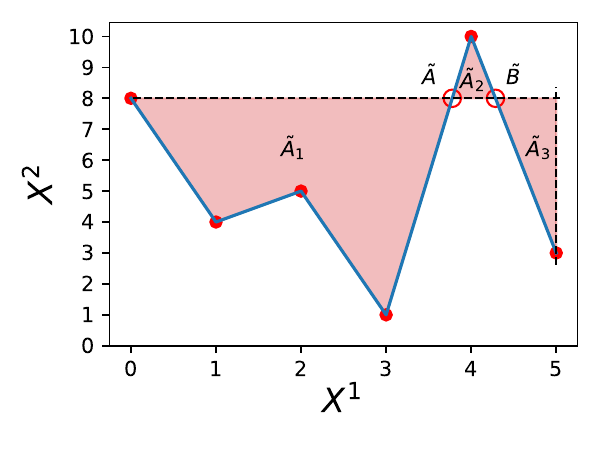}
         \caption{The total area $S^{2,1} = \tilde{A}_1 + \tilde{A}_2 + \tilde{A}_3$.}
     \end{subfigure}
     \caption{A schematic partitioning of the areas $S^{1,2}$ and $S^{2,1}$ enclosed by the path and their corresponding dashed lines into a sum of individual contributions. Circles denote the point of intersection of the path and the corresponding dashed lines.}
     \label{fig:signature_terms_s12_s21_split}
\end{figure}

Switching the order of integration over the path in the terms $S^{1,2}$ and $S^{2,1}$ gives rise to two areas which complete each other and add up to the total area of a rectangle with side lengths $X^1$ and $X^2$, which also corresponds to the shuffle product relation
\begin{eqnarray*}
S^{1}\cdot S^{2} &=& S^{1,2} + S^{2,1},\\
(-5)\cdot 5 &=& -10.5 + (-14.5).
\end{eqnarray*} 
The geometric interpretation of the higher order terms is less intuitive and we omit this discussion. It is worth noticing, that the areas that correspond to the second order signature terms ($S^{1,2}$ and $S^{2,1}$) have positive and negative signs depending on their position relative to the horizontal straight line. If we split the areas in Fig.~\ref{fig:signature_terms_s12_s21} into individual areas that lie above and underneath the horizontal dashed lines, we obtain partitions as shown in Fig.~\ref{fig:signature_terms_s12_s21_split}.

The intersection points, denoted by red circles in Fig.~\ref{fig:signature_terms_s12_s21_split} can be easily computed as
\begin{eqnarray}
    A &=& (5/2, 3), \;\;\;B = (29/9, 3),\nonumber\\
    \tilde{A} &=& (34/9, 8), \;\;\;\tilde{B} = (30/7, 8).\nonumber
\end{eqnarray}
Subsequently, one can directly compute the individual areas
\begin{eqnarray}
    A_1 &=& -5, \;\;\; A_2 = 13/18, \;\;\; A_3 = -56/9,\nonumber\\
    \tilde{A}_1 &=& -119/9, \;\;\; \tilde{A}_2 = 32/63, \;\;\; \tilde{A}_3 = -25/14,\nonumber
\end{eqnarray}
and demonstrate that the results
\begin{eqnarray}
    S^{1,2} &=& A_1 + A_2 + A_3 = -10.5,\nonumber\\
    S^{2,1} &=& \tilde{A}_1 + \tilde{A}_2 + \tilde{A}_3 = -14.5\nonumber
\end{eqnarray}
correspond to the direct computation of the signature terms as shown in \eqref{eq:four_sig_numerical}. Note that the sign of an area is determined by the direction of the path that encloses the area, as demonstrated in Fig.~\ref{fig:signed_area_examples}.

\begin{figure}
    \centering
    \includegraphics[width=0.75\textwidth]{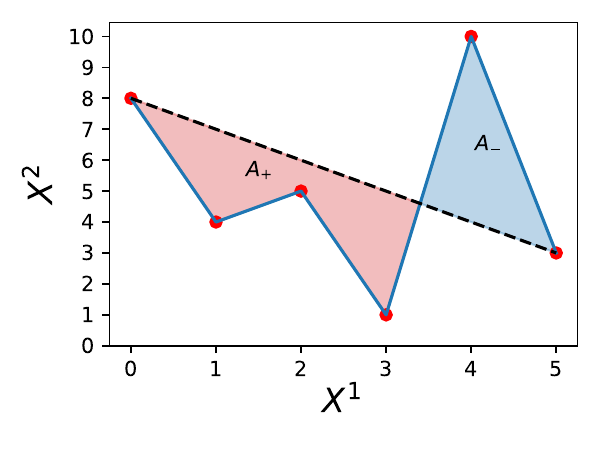}
    \caption{Example of the signed L{\'e}vy area enclosed by a piecewise linear path and the
chord that connects the two endpoints. The light blue area is negative and the pink area is positive according to the direction of travel along the path (from left to right).}
    \label{fig:signature_levy_area}
\end{figure}

The L{\'e}vy area \eqref{eq:LevyArea} of the path $\{(X^1, X^2)\}$ \eqref{eq:sig_x1x2} is shown in Fig~\ref{fig:signature_levy_area} and is given by
\begin{equation}\label{eq:signature_levy_x1x2}
    A_{\text{L{\'e}vy}}=\frac{1}{2}\left(S^{1,2}-S^{2,1}\right) = 2.
\end{equation}

Using the intersection point $(3.4, 4.6)$ between the chord and the linear segment between $t_3$ and $t_4$, one can easily compute the areas $A_{+}=6.8$ and $A_{-}=-4.8$ and confirm that $A_{\text{L{\'e}vy}} = A_{+}+A_{-}$ is equivalent to \eqref{eq:signature_levy_x1x2}.

\subsubsection{The log-signature of a path}

The shuffle product identity (Theorem~\ref{thm:shuffle}) implies that any polynomial function on signatures can be written as a linear combination of signature elements.
However, these relations imply that the signature contains `repeated information', e.g. $S(X)^{i,i}_{a,b}$ can always be determined from the level 1 signature since $S(X)^{i,i}_{a,b}=\frac12 (S(X)^i_{a,b})^2$; for some tasks (e.g. linear regression) this repetition can be useful, as it increases the number of potentially useful features, but in some cases, one may wish to remove it.
The log-signature (also known as a logarithmic signature), as defined in Section \ref{subsubsec:logSig}, is a condensed form of the signature that removes this repeated information: it contains the same information as the signature but with fewer terms. 

For example, consider a two-dimensional data stream $\{X\}$ and the vector space of formal series in two non-commuting indeterminates $\{e_1, e_2\}$ (see Definition~\ref{def:formal_power_series}). Recalling the log-signature from Definition~\ref{def:logsig} and the result of Theorem~\ref{thm:log_sig_series}, the log-signature truncated at level $L=4$ can be written as
\begin{equation}\label{eq:log_sig}
\begin{aligned}
    \log S(X) &= \lambda_{1}e_1 + \lambda_{2}e_2+ \lambda_{12}[e_1, e_2]  \\
    &\quad+ \lambda_{112}[e_1,[e_1,e_2]] + \lambda_{212}[e_2,[e_1,e_2]] \\
    &\quad +\lambda_{1112}[e_1,[e_1,[e_1,e_2]]] + \lambda_{2112}[e_2,[e_1,[e_1,e_2]]]
    \\
    &\quad + \lambda_{2212}[e_2,[e_2,[e_1,e_2]]],
\end{aligned}
\end{equation}
where $\lambda_{i_1,...i_k}\in \R$ are coefficients and $[x,y] = x\otimes y - y \otimes x$ is the Lie bracket as defined in \eqref{eq:lie_brackets}.
There are \textit{a priori} no relations between the coefficients $\lambda_1,\ldots, \lambda_{2212}$ (in fact, for any choice of $\lambda_1,\ldots, \lambda_{2212}$, one can find a piecewise linear path $X$ so that $\log S(X)$ is equal to the right-hand side of~\eqref{eq:log_sig}, see \cite[Theorem~7.28]{FrizVictoir10}).
This is in contrast to $S(X)$ because the shuffle identities imply non-trivial constraints on its coefficients (e.g. $S(X)^1S(X)^2 = S(X)^{1,2}+S(X)^{2,1}$).
Since $S(X)$ and $\log S(X)$ determine each other, this shows that $\log S(X)$ is a compressed version of $S(X)$.

The correspondence between the signature and log-signature terms can be computed from the definition~\eqref{eq:log_def}. For example, building on~\eqref{eq:direct_computation_logsig}, the correspondence between the coefficients of the signature and log-signature up to the level $L=3$ is
\begin{equation}\label{eq:sig_logsig_correspondence_l_3}
\begin{aligned}
    \lambda_1 &= S^{1}, \;\;\;\lambda_2 = S^{2}, \;\;\;\lambda_{12} = \frac{1}{2!}\left(S^{1,2} - S^{2,1}\right),\\
    \lambda_{112} &= \frac{1}{3!}\left(S^{1,1,2} + S^{2,1,1} - 2 S^{1,2,1}\right),
    \\
    \lambda_{212} &= -\frac{1}{3!}\left(S^{1,2,2} + S^{2,2,1} - 2 S^{2,1,2}\right),
\end{aligned}
\end{equation}
and it is possible to extend this correspondence to any truncation level $L$.

Let us illustrate the geometric intuition for the low-order log-signature terms. Using the relationships in \eqref{eq:sig_logsig_correspondence_l_3}, the log-signature coefficients up to level $L=3$ of a two-dimensional path $X = \{(X^1, X^2)\}$ from \eqref{eq:sig_x1x2} are
\begin{center}
        \begin{tabular}{lrrrrrrr}
        $\log S(X)$ = & ($\lambda_{1}$, & $\lambda_{2}$, & $\lambda_{12}$, & $\lambda_{112}$, & $\lambda_{212}$ ) \\
                \hfill =  & ($5$,       & $-5$,      & $2$,      & $-12$,     & $-35/6$).
    \end{tabular}
\end{center}
One can immediately see that $\lambda_1$ and $\lambda_2$ are the total increments along each of the axes and $\lambda_{12}$ corresponds to the L{\'e}vy area \eqref{eq:LevyArea} as depicted in Fig.~\ref{fig:signature_levy_area}. The higher order log-signature terms $L\ge3$ correspond to ``areas of areas'' \cite{reizenstein2019iterated} and can be represented as iterated Lie brackets of non-commuting indeterminates.
The log-signature, containing the same information as the signature, can likewise be used as a feature set in statistical and machine learning applications, at times with more efficiency, see e.g. \cite{2019arXiv190808286L,Liao22,Reizenstein20}.

\subsubsection{Numerical computation of the signature}
To streamline the development and application of the signature method, several numerical packages with Python interfaces have been developed for the computation of the signature. The `ESig' package is implemented in C++ with an intuitive Python interface \cite{esiglib}. This package computes the signature and log-signature coefficients, provides additional auxiliary tools, and can be run on CPU and, if available, on GPUs to accelerate computations. Another package with Python interface `iisignature' was proposed and its implementation details are described in \cite{reizenstein2017calculation, Reizenstein20}. In addition, the `signatory' Python package was developed to enable the integration of the signature transformation layer in deep learning applications \cite{kidger2021signatory}. More information about the usage of these numerical packages can be found in their respective documentations.

\subsubsection{Further relationship between the signature terms and data points}
\label{subsec:further_relationship}

\begin{figure}[ht]
\centering
\includegraphics[scale = 0.99]{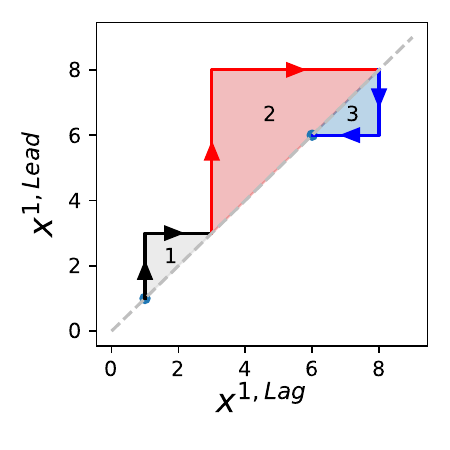}
\caption{Decomposition of the full lead-lag path into three individual parts. The direction of movement along the path is indicated by arrows.}
\label{fig:threeTriangls}
\end{figure}

\begin{figure}[ht]
\centering
\includegraphics[width=\linewidth]{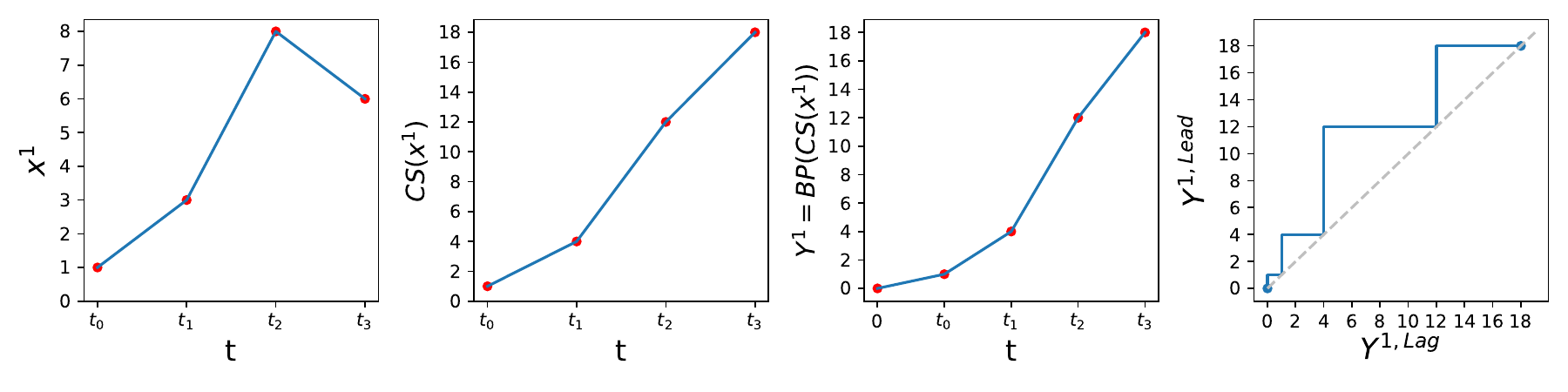}
\caption{The subsequent transformations of the original path $\{X^1\}$ into a resulting path $\{\tilde{X}^1\} = \mathrm{LL}\big(\mathrm{BP}\big(\mathrm{CS}(\{X^1\})\big)\big)$ as shown in \eqref{eq:composition_transformations_example}.}
\label{fig:ll_bp_cs}
\end{figure}

We now illustrate how the signature allows us to compute statistical moments between data streams.
Consider the lead-lag path (Fig.~\ref{fig:threeTriangls}) constructed from \eqref{eq:twoOneDimSeq_2} and its lead-lag transformation \eqref{eq:leadLagAlgo_X1}. We can decompose this figure into three right-angled isosceles triangles Fig.~\ref{fig:threeTriangls}. Note that the direction of movement along this path, starting from the point $X^1_{t_0}$ and moving towards the endpoint $X^1_{t_3}$ corresponds to a negative sign, thus the total area will be negative. All three triangles have the same direction of movement, resulting in the sum of their individual contributions. 
The absolute value of the total area is given by the sum of the half areas of squares
\begin{align*}
|A| &= \frac{1}{2}\left[\left(X^{1}_{t_1} - X^{1}_{t_0}\right)\left(X^{1}_{t_1} - X^{1}_{t_0}\right) +
\left(X^{1}_{t_2} - X^{1}_{t_1}\right)\left(X^{1}_{t_2} - X^{1}_{t_1}\right)\right.
\\
&\qquad + \left.\left(X^{1}_{t_3} - X^{1}_{t_2}\right)\left(X^{1}_{t_3} - X^{1}_{t_2}\right)\right]\\
&= \frac{1}{2}\left[(3-1)^2 + (8-3)^2 + (6-8)^2\right] = 16.5,
\end{align*}
which exactly equals to the L{\'e}vy area enclosed by the lead-lag path $\{X^1\}$ and the straight line connecting the endpoints, and can be computed via the signature terms \eqref{eq:LevyArea}.
Let us write
\begin{equation}\label{eq:quadraticVar}
[X]_t = \sum_{i=1}^{N}\left(X_{t_{i}}- X_{t_{i-1}}\right)^2,
\end{equation}
which has the simple meaning of the \emph{quadratic variation} of the path constructed from $\{X_{t_i}\}_{i=0}^N$ and is related to the \emph{variance} of the path. Thus one can generally write for any sequence $\{X_{t_i}\}_{i=0}^N$
\begin{equation*}
A_{\text{L{\'e}vy}}^{\{(X^{\mathrm{Lead}}, X^{\mathrm{Lag}})\}} = \frac{1}{2}[X].
\end{equation*}
For example, for the lead-lag path $\{(X^{1, \mathrm{Lead}}, X^{1, \mathrm{Lag}})\}$ from~\eqref{eq:leadLagAlgo_X1}, originated from a one-dimensional sequence \eqref{eq:twoOneDimSeq_2}, the first order log-signature terms correspond to the total increments along each dimension, and the second order term is the L{\'e}vy area which equals to the quadratic variation of the one-dimensional sequence:
\begin{eqnarray*}
\log S\left(\{(X^{1, \mathrm{Lead}}, X^{1, \mathrm{Lag}})\}\right) &=& \left(\Delta X^{1, \mathrm{Lead}}, \Delta X^{1, \mathrm{Lag}}, \frac{1}{2}[X^1], ...\right) \\\nonumber
&=& (5, 5, 16.5, ...).
\end{eqnarray*}
Equivalently, one can write the above result in terms of the signature with all the terms included as
\begin{equation*}
S\left(\{(X^{1, \mathrm{Lead}}, X^{1, \mathrm{Lag}})\}\right) = (1, 5, 5, 12.5, 29, -4, 12.5, ...)
\end{equation*}
and confirm that
\begin{equation*}
[\{X^1\}] = S^{1,2}\left(\{(X^{1, \mathrm{Lead}}, X^{1, \mathrm{Lag}})\}\right) - S^{2,1}\left(\{(X^{1, \mathrm{Lead}}, X^{1, \mathrm{Lag}})\}\right).
\end{equation*}
Next we demonstrate certain properties of paths which originate from composite transformations. Consider the one-dimensional sequence $\{X^1 \}$  as in~\eqref{eq:twoOneDimSeq_2}
and consider applying to $\{X^1 \}$ the series of transformations given by the cumulative sum $\mathrm{CS}(.)$ defined by~\eqref{eq:cumSumDef},
followed by base-point augmentation $\mathrm{BP}(.)$ 
defined by~\eqref{eq:bp_augmentation},
followed by lead-lag transformation $\mathrm{LL}(.)$ 
defined by\eqref{eq:lead_stream}-\eqref{eq:lag_stream}. The resulting sequence can be written as the composition of transformations
\begin{equation}\label{eq:composition_transformations}
\{\tilde{X}^1\} = \mathrm{LL} \circ \mathrm{BP} \circ \mathrm{CS}(\{X^1\}) = \mathrm{LL}\big(\mathrm{BP}\big(\mathrm{CS}(\{X^1\})\big)\big).
\end{equation}
It follows that, defining $\{Y^1\} = \mathrm{BP}\big(\mathrm{CS}(\{X^1\})$,
\begin{eqnarray}
\{X^1_{t_i}\}_{i=0}^{N=3} &=& \{1,3,8,6\},\nonumber
\\
\mathrm{CS}(\{X^1\}) &=& \{1,4,12,18\},\nonumber
\\
\{Y^1\} &=& \{0,1,4,12,18\},
\nonumber\\
\{\tilde{X}^1\} &=& \mathrm{LL}\big(\mathrm{BP}\big(\mathrm{CS}(\{X^1\})\big)\big) \label{eq:composition_transformations_example}
\\
&=& \{(Y^{1,\mathrm{Lead}}, Y^{1,\mathrm{Lag}})\}\nonumber
\\
&=& \{(0,0),(1,0),(1,1),(4,1),
(4,4),(12,4),(12,12),
(18,12),(18,18)\}. \nonumber
\end{eqnarray}
The resulting $\{\tilde{X}^1\}$ path is shown in Fig.~\ref{fig:ll_bp_cs}.

Explicitly \eqref{eq:composition_transformations_example} can be written as
\begin{eqnarray*}
    Y^{1,\mathrm{Lead}} &=& \{0, S_0, S_0, S_1, S_1, S_2, S_2, S_3, S_3\},\\
    Y^{1,\mathrm{Lag}} &=& \{0,0,S_0, S_0, S_1, S_1, S_2, S_2, S_3\},
\end{eqnarray*}
where $S_k$ is a partial sum defined in \eqref{eq:cumSumDef}. We denote the signature of the composite transformation $\{\tilde{X}^1\}$ from~\eqref{eq:composition_transformations} as
\begin{equation}\label{eq:xtilde_sig}
    \tilde{S} = S(\{\tilde{X}^1\}) = (1, \tilde{S}^{1}, \tilde{S}^{2}, \tilde{S}^{1,1}, \tilde{S}^{1,2}, \tilde{S}^{2,1}, \tilde{S}^{2,2}, ...).
\end{equation}
Numerical evaluation of the expression in \eqref{eq:xtilde_sig} yields
\begin{equation}\label{eq:xtilde_sig_num}
    \tilde{S} = (1, 18,  18, 162, 217, 107, 162, ...).
\end{equation}
From the definition of the signature terms and the cumulative sum, it follows that
\begin{eqnarray*}
    \tilde{S}^{1} &=& \int d {Y}^{1,\mathrm{Lag}} = {Y}^{1,\mathrm{Lag}}_{t_{N}} - {Y}^{1,\mathrm{Lag}}_{t_0} = \Delta{X^{1,\mathrm{Lag}}}\\ &=& \mathrm{BP}\big(\mathrm{CS}(\{X^1\})\big)_{t_4} - \mathrm{BP}\big(\mathrm{CS}(\{X^1\})\big)_{t_0}\nonumber\\ &=& S_{k=3} - 0\\ &=& X^1_{t_0}+X^1_{t_1}+X^1_{t_2}+X^1_{t_3} - 0\\ &=& 1+3+8+6 = 18.
\end{eqnarray*}
%
Similarly, one can show that $\tilde{S}^{2} = \tilde{S}^{1}$. Therefore, the first order terms of the signature of the transformed sequence $\{\tilde{X}^1\}$ equal to the sum of the original sequence $\{{X}^1\}$:
\begin{equation*}
    \tilde{S}^{1} = \tilde{S}^{2} = \Delta{Y^{1,\mathrm{Lead}}} = \Delta{Y^{1,\mathrm{Lag}}}.
\end{equation*}
This result can be easily confirmed numerically by comparing the corresponding terms in \eqref{eq:xtilde_sig} and \eqref{eq:xtilde_sig_num}. In a similar fashion, one can demonstrate that quadratic variation \eqref{eq:quadraticVar} of the transformed sequence $[\{\tilde{X^1}\}]$ is proportional to \textit{variance} of the original sequence $\{X^1\}$ from~\eqref{eq:twoOneDimSeq_2}.


\subsection{Machine learning with signature features}

The signature of a path is a collection of iterated integrals that succinctly capture the path's geometric properties and encode its essential information. This makes it particularly well-suited for machine learning applications involving sequential data and time series analysis. The workflow is simple and summarized in the following algorithm:
$$ \text{{\it data}} \rightarrow \text{{\it path}} \rightarrow \text{{\it signature of path}} \rightarrow \text{{\it features of data}}\rightarrow \text{{\it learning algorithms}} $$
This algorithm is general and works for any type of sequential data which can be embedded into a continuous path. The extracted features can be used for various types of machine learning applications, including both supervised and unsupervised learning. For example, one can classify time series or distinguish clusters of data. One of the advantages of feature extraction with the signature method is that the signature is sensitive to the geometric shape of a path.
Sensitivity to the geometric shape of the input data has led to a successful application of the signature method to Chinese character recognition problem \cite{graham2013sparse}. One of the most well-known and natural applications of the signature method is in quantitative finance, namely analysis of time series data \cite{gyurko2014extracting,levin2015learning}. Time series represent ordered sequential data, which is a natural candidate for creating a path, followed by computing the signature and applying machine learning algorithms for further analysis.
Moreover, if the input data come from several parallel sources, this will result in a multi-dimensional path. An example of such data is {\it panel} data (in Econometrics) or {\it longitudinal} data (in Medicine, Psychology, Biostatistics etc.) which involve repeated observations of the same quantity over periods of time.

Concluding this brief introduction to the applications of the signature method, we emphasize that this method introduces a new concept of dealing with data: thinking of data as geometric paths and using the signature method to analyze this data.
In the rest of this subsection, we demonstrate how to apply the signature transformation to problems in statistical machine learning with streamed data, focusing on the common analytical task of classification.  However, other machine learning and inferential statistical tasks with streamed data can also be performed using the signature formalism, including regression, clustering, and kernel learning, some of which we review in Section \ref{subsec:overview_applications}.

\subsubsection{Classification of handwritten digits: an example}

\begin{figure}[ht]
    \centering
    \includegraphics[width=\linewidth]{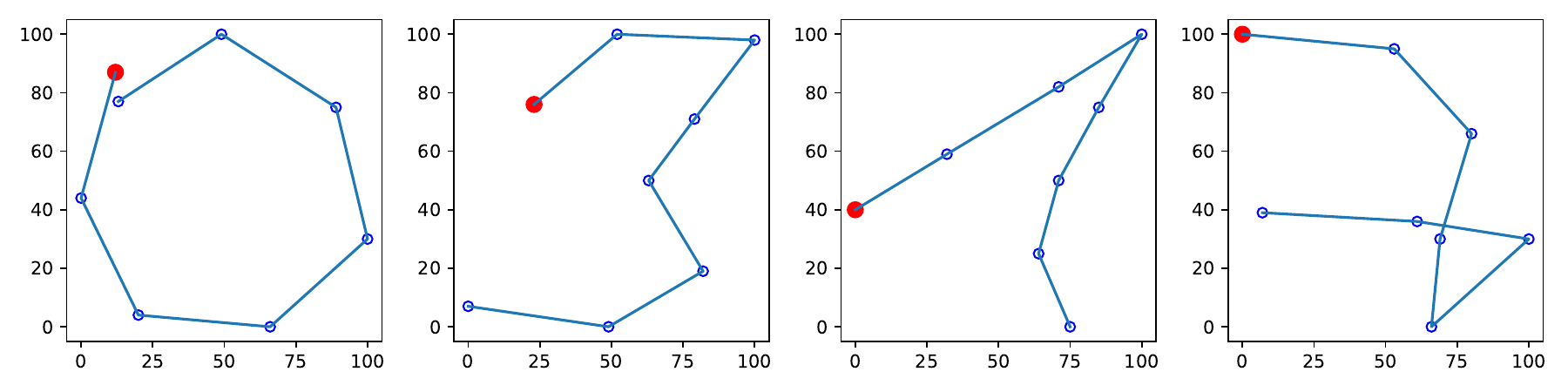}
    \caption{Pictorial representation of the first four rows from the Table~\ref{tab:digits}. From left to right: `0', `3', `1' and `7'. Solid red dots represent the starting point $(x_1, y_1)$.}
    \label{fig:digits}
\end{figure}

In this section, we demonstrate the application of the signature method to the machine learning task of pen-based classification of handwritten digits. The dataset is available from the University of California, Irvine, Machine Learning Repository \cite{misc_pen-based_recognition_of_handwritten_digits_81}. The data set comprises 250 handwritten integer numbers in the range of 0 to 9, generated by 44 writers using a tablet with a stylus. The data represent a collection of consecutive data points represented by $(x, y)$ coordinates. The raw data that were captured from the tablet consisted of integer values and were normalized to lie in the range between 0 and 100. The consecutive data points were resampled and represented as a sequence of coordinate pairs $(x_i, y_i), i=1,\dots,8$, regularly spaced on the trajectory. An example of ten random digits in the tabular format is shown in Tab. \ref{tab:digits} and the first four rows are displayed in Fig.~\ref{fig:digits}.
\begin{table}[ht]
    \centering
    \normalsize
    \setlength{\tabcolsep}{3pt}
\begin{tabular}{rrrrrrrrrrrrrrrrr}
 $x_1$ &  $y_1$ &  $x_2$ &  $y_2$ &  $x_3$ &  $y_3$ &  $x_4$ &  $y_4$ &  $x_5$ &  $y_5$ &  $x_6$ &  $y_6$ &  $x_7$ &  $y_7$ &  $x_8$ &  $y_8$ &  Label \\
\hline
 $12$ &   $87$ &    $0$ &   $44$ &   $20$ &    $4$ &   $66$ &    $0$ &  $100$ &   $30$ &   $89$ &   $75$ &   $49$ &  $100$ &   $13$ &   $77$ &      $0$ \\
 $23$ &   $76$ &   $52$ &  $100$ &  $100$ &   $98$ &   $79$ &   $71$ &   $63$ &   $50$ &   $82$ &   $19$ &   $49$ &    $0$ &    $0$ &    $7$ &      $3$ \\
 $0$ &   $40$ &   $32$ &   $59$ &   $71$ &   $82$ &  $100$ &  $100$ &   $85$ &   $75$ &   $71$ &   $50$ &   $64$ &   $25$ &   $75$ &    $0$ &      $1$ \\
 $0$ &  $100$ &   $53$ &   $95$ &   $80$ &   $66$ &   $69$ &   $30$ &   $66$ &    $0$ &  $100$ &   $30$ &   $61$ &   $36$ &    $7$ &   $39$ &      $7$ \\
 $0$ &   $67$ &   $43$ &   $74$ &   $78$ &   $91$ &  $100$ &  $100$ &   $95$ &   $74$ &   $87$ &   $49$ &   $84$ &   $23$ &   $96$ &    $0$ &      $1$ \\
 $10$ &   $97$ &    $2$ &   $47$ &   $24$ &    $1$ &   $74$ &    $0$ &  $100$ &   $43$ &   $82$ &   $90$ &   $33$ &  $100$ &    $0$ &   $65$ &      $0$ \\
 $0$ &   $48$ &   $30$ &   $67$ &   $65$ &   $85$ &  $100$ &  $100$ &   $84$ &   $75$ &   $65$ &   $51$ &   $47$ &   $25$ &   $34$ &    $0$ &      $1$ \\
 $55$ &   $70$ &   $73$ &  $100$ &   $62$ &   $66$ &   $49$ &   $26$ &   $30$ &    $0$ &    $0$ &    $2$ &   $50$ &    $3$ &  $100$ &    $6$ &      $1$ \\
$100$ &  $100$ &   $66$ &   $88$ &   $32$ &   $66$ &    $7$ &   $39$ &    $0$ &   $10$ &   $44$ &    $0$ &   $62$ &   $18$ &   $14$ &   $22$ &      $6$ \\
 $85$ &  $100$ &   $43$ &   $76$ &   $11$ &   $48$ &    $0$ &   $17$ &   $44$ &    $0$ &  $100$ &   $10$ &   $59$ &   $20$ &    $1$ &   $10$ &      $6$ \\
 \hline
\end{tabular}
    \caption{An example of ten random samples of handwritten digits in a tabular format from the Pen-based Recognition of Handwritten Digits data set \cite{misc_pen-based_recognition_of_handwritten_digits_81}. Columns denoted by $x_i$, and $y_i$ $(i=1,\dots,8)$ correspond to coordinates (features) and the last column designates the class label of digits.}
    \label{tab:digits}
\end{table}

Each data sample is represented by a two-dimensional coordinate sequence $(x_i, y_i) \in \mcX$, that can be seen as a path in $\mathbb{R}^2$.
For simplicity, we apply the signature transformation without composite augmentations. Following the steps outlined in Section \ref{sec:signature_of_a_path}, the signature is the feature map
\begin{equation*}
    S:\mcX \rightarrow \Sigma,
\end{equation*}
where $S^{I}\in\Sigma$ are the elements of the infinite signature series \eqref{eq:definSig} and $I$ is a multi-index defined in \eqref{eq:words_def}. For example, applying the level-2 truncated signature transformation on the first four rows
\begin{eqnarray*}
    X^{\mathrm{zero}} = \{(x_i, y_i)\}_{i=1}^{N=8} &=& \{(12, 87), (0, 44), (20, 4), (66, 0),\\
     && (100, 30), (89, 75), (49, 100), (13, 77)\},\\
    X^{\mathrm{three}} &=& \{(23, 76), (52, 100), (100, 98), (79, 71),\\
     && (63, 50), (82, 19), (49, 0), (0, 7)\},\\
    X^{\mathrm{one}} &=& \{(0, 40), (32, 59), (71, 82), (100, 100),\\
     && (85, 75), (71, 50), (64, 25), (75, 0)\},\\
    X^{\mathrm{seven}} &=& \{(0, 100), (53, 95), (80, 66), (69, 30),\\
     && (66, 0), (100, 30), (61, 36), (7, 39)\}     
\end{eqnarray*}
yields
\begin{eqnarray*}
    S(X^{\mathrm{zero}}) &=& (1, 1, -10, 0.5, 7044.5, -7054.5, 50),\\
    S(X^{\mathrm{three}}) &=& (1, -23, -69, 264.5, -4893, 6480, 2380.5),\\
    S(X^{\mathrm{one}}) &=& (1, 75, -40, 2812.5, -4660, 1660, 800),\\
    S(X^{\mathrm{seven}}) &=& (1, 7, -61, 24.5, -3693, 3266, 1860.5).
\end{eqnarray*}
Applying the signature transformation on each row of the data matrix in Tab.~\ref{tab:digits} will generate the \textit{signature feature matrix} given in Tab.~\ref{tab:sig_features}.

\begin{table}[ht]
    \centering
    \normalsize
    \setlength{\tabcolsep}{4pt}
    \begin{tabular}{rrrrrrrr}
     $1$ &   $S^{1}$ &   $S^{2}$ &     $S^{1,1}$ &       $S^{1,2}$ &      $S^{2,1}$ &      $S^{2,2}$ &  Label \\
    \hline
     $1$ &   $1$ & $-10$ &    $0.5$ &  $7044.5$ & $-7054.5$ &   $50$ &      $0$ \\
     $1$ & $-23$ & $-69$ &  $264.5$ & $-4893$ &  $6480$ & $2380.5$ &      $3$ \\
     $1$ &  $75$ & $-40$ & $2812.5$ & $-4660$ &  $1660$ &  $800$ &      $1$ \\
     $1$ &   $7$ & $-61$ &   $24.5$ & $-3693$ &  $3266$ & $1860.5$ &      $7$ \\
     $1$ &  $96$ & $-67$ & $4608$ & $-7123$ &   $691$ & $2244.5$ &      $1$ \\
     $1$ & $-10$ & $-32$ &   $50$ &  $7388.5$ & $-7068.5$ &  $512$ &      $0$ \\
     $1$ &  $34$ & $-48$ &  $578$ & $-4179$ &  $2547$ & $1152$ &      $1$ \\
     $1$ &  $45$ & $-64$ & $1012.5$ &   $178$ & $-3058$ & $2048$ &      $1$ \\
     $1$ & $-86$ & $-78$ & $3698$ &  $5984$ &   $724$ & $3042$ &      $6$ \\
     $1$ & $-84$ & $-90$ & $3528$ &  $6028.5$ &  $1531.5$ & $4050$ &      $6$ \\
    \hline
    \end{tabular}
    \caption{The signature feature matrix from the original data presented in Tab. \ref{tab:digits} and the corresponding labels.}
    \label{tab:sig_features}
\end{table}
Now, one can use the signature feature matrix as input to statistical hypothesis tests for downstream analytical tasks or machine learning algorithms to predict the class labels. 

\begin{figure}[ht]
    \centering
    \includegraphics[width=\linewidth]{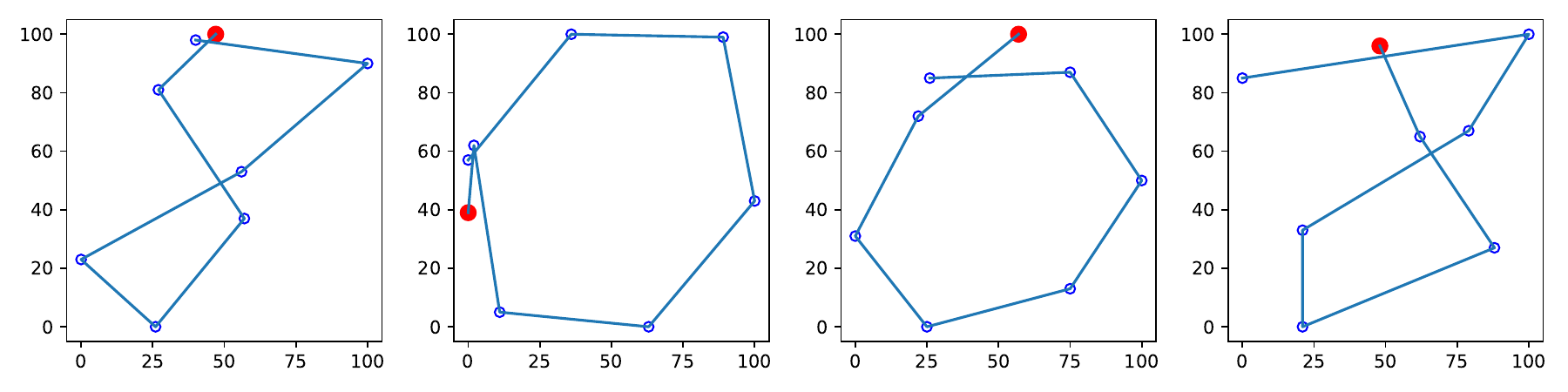}
    \caption{Examples of handwritten zeros and eights. From left to right: `8', `0', `0', `8'. Solid red dot represents the starting point $(x_1, y_1)$.}
    \label{fig:digits_8_0}
\end{figure}

\begin{figure}
    \centering
    \includegraphics[width=\linewidth]{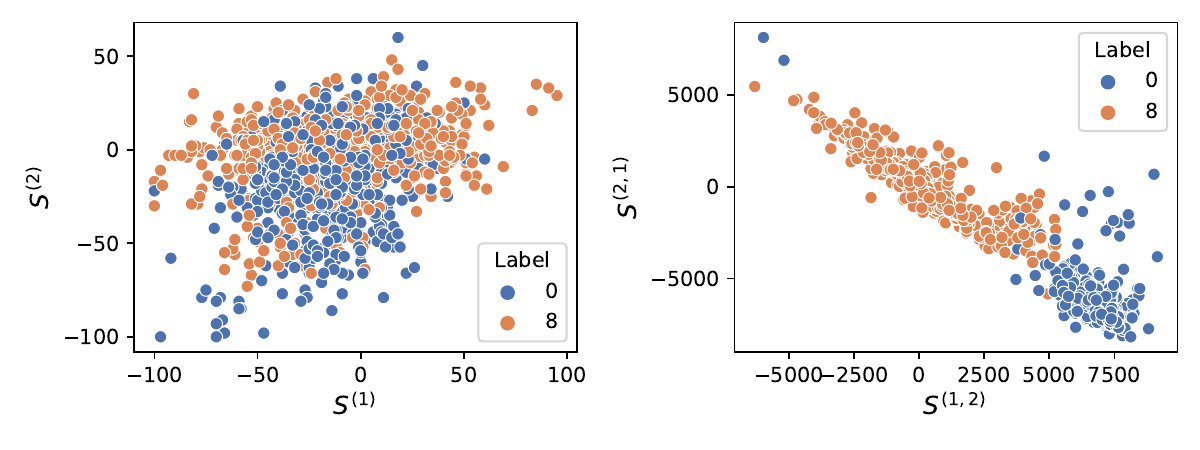}
    \caption{The distribution of the first (left panel) and second (right panel) signature terms describing `zero' and `eight' digits. The second order signature terms, that are proportional to a signed enclosed area, demonstrate good separability between the two classes.}
    \label{fig:sig_s1_s2_vs_s12_s21}
\end{figure}

Another interesting observation is that the higher order signature terms can capture more nuanced patterns of paths. Consider only samples of handwritten `zero' and `eight' digits, such as those shown in Fig.~\ref{fig:digits_8_0}.
Both `zero' and `eight' shapes share a similar geometric pattern: the starting and end points usually coincide. However, the \textit{areas} enclosed by both shapes will be different due to their orientations as discussed in Section \ref{sec:signed_area}. The clear differences between the distributions of the first $S^{1}, S^{2}$ and second order $S^{1,2}, S^{2,1}$ signature terms are shown in Fig.~\ref{fig:sig_s1_s2_vs_s12_s21} and can improve the performance of machine learning or statistical approaches.

\begin{figure}
    \centering
    \includegraphics[width=0.7\linewidth]{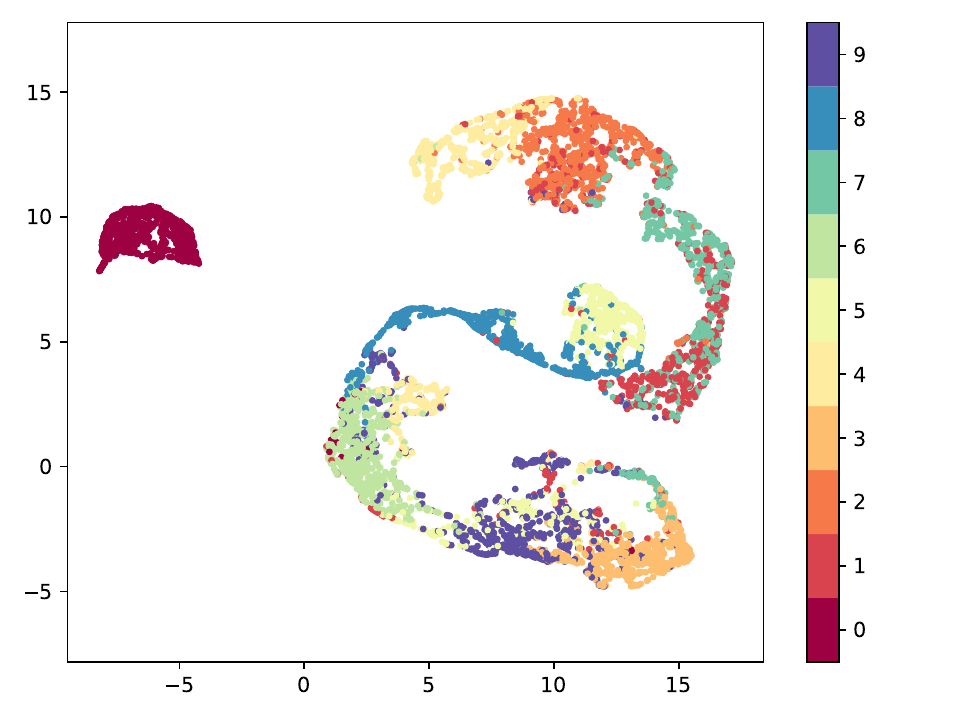}
    \caption{Projection of the signature feature space from $\mathbb{R}^7$ onto $\mathbb{R}^2$ using UMAP. The right column designates class labels $0, \dots, 9$.}
    \label{fig:UMAP}
\end{figure}

Furthermore, the signature feature matrix can also be used for unsupervised learning tasks, including dimensionality reduction for clustering approaches. For example, applying the Uniform Manifold Approximation and Projection (UMAP) for Dimension Reduction method \cite{mcinnes2018umap} on the signature feature matrix, one can find clusters of points belonging to the same class label, see Fig.~\ref{fig:UMAP}.

\subsection{Overview of signature-based machine learning applications}
\label{subsec:overview_applications}

Over the past two decades, the signature method has been the subject of extensive research and application across various scientific domains.
The method has found success in diverse areas such as finance, healthcare, natural language processing, wearable devices, and computer vision, demonstrating its versatility and the powerful expressiveness inherent in its approach. This adaptability has established the signature method as a valuable asset in the ongoing research on modern machine learning techniques for streamed data.

The signature method's application to sequential data representation has evolved substantially, extending far beyond its original scope and undergoing considerable refinement. To conclude, we aim to provide a concise overview of some notable examples where the signature method has been employed to tackle diverse challenges in machine learning and data analysis. Through these examples, we hope to showcase the wide-ranging implications and potential of this innovative approach in further advancing the field of machine learning. 

Already in 2005, Lyons and Sidorova worked on the application of rough path theory to the problem of sound compression \cite{lyons2005sound}. They presented a new approach based on the signature that provides an alternative to traditional Fourier and wavelet transforms. The main difference between these methods is in the linearity of the Fourier approach, while the signature method accounts for non-linear dependencies. 

The signature transformation has been extensively used in financial applications. Field {\it et al.}~\cite{gyurko2014extracting} have applied the signature transformation to multidimensional financial data streams to identify atypical market behavior and recognize patterns of trading algorithms. In~\cite{levin2015learning}, Levin, Lyons, and Ni proposed a new non-parametric regression method based on the expected signature of paths. Truncating the signature results in an efficient local description and offers a provable efficiency gain over linear path segment descriptions with the same number of features. This approach can potentially achieve significant dimension reduction in highly oscillatory data streams. In numerical examples, their method achieves similar accuracy to that of the Gaussian process approach but with lower computational cost, especially for large sample sizes. The truncated signature-based regression and classification models have further led to numerous applications in finance, such as the obtaining approximate solutions to the problem of optimal execution \cite{kalsi2020optimal}, non-parametric pricing and hedging of exotic derivatives \cite{lyons2020non} and many others \cite{perez2020signatures}.

The ability of the signature transformation to handle complex data streams, reduce dimensionality of the feature space, and enhance prediction accuracy makes it a valuable asset in healthcare and medicine. As wearable technology becomes more common, the amount of data generated by these devices is enormous. By analyzing data from connected devices, e.g., heart rate monitors, fitness trackers, sleep monitors, gait and other modalities, researchers can gain insights into patients' health conditions, such as identifying early patients with neurological disorders that have a higher risk of falls \cite{rehman2020gait}. Electronic health records (EHRs) contain vast amounts of data, including patient demographics, medical history, medications, and lab results. Applying the signature method to EHRs can help identify patterns and trends, enable better disease management, and support early detection of potential health issues, such as an early onset of sepsis \cite{morrill2020utilization} and diagnosis of Alzheimer's disease using brain image-derived biomarkers (e.g., the hippocampus, ventricles and whole brain volumes) \cite{moore2019using}. The signature method was applied to analyze data from self-reporting tools and mobile applications, to detect longitudinal patterns \cite{kormilitzin2017detecting} in clinical trials \cite{geddes2015comparative}, predict mood changes and episodes \cite{kormilitzin2016application}, derive clinically meaningful information from daily self-reported mood ratings to classify patients’ diagnosis, predict mood scores \cite{perez2018signature}, and efficiently capture the chronological information in medical texts \cite{kormilitzin2020efficient}.

The signature approach has been applied to computer vision tasks such as human-pose estimation \cite{yang2022developing} and recognition of handwriting in various languages \cite{wilson2018path,xie2017learning}. The signature approach has been applied in deep learning \cite{kidger2019deep}, neural controlled differential equations \cite{kidger2020neural}, and topological data analysis \cite{CNO20}. Signatures were furthermore combined with kernel methods applicable to problems in classification and hypothesis testing \cite{CO18,KO19}.

Further examples of applications of rough paths, signatures, and neural controlled differential equations can be found at \url{https://www.datasig.ac.uk/papers}.

\bibliographystyle{abbrv}
\bibliography{Refs}

\end{document}